%% file: main.tex
\documentclass[10pt]{article} 
\usepackage[accepted]{tmlr}

\input{math_commands.tex}

\usepackage{hyperref}
\usepackage{url}
\usepackage{booktabs}       
\usepackage{amsfonts}       
\usepackage{nicefrac}       
\usepackage{microtype}      
\usepackage{xcolor}         

\usepackage{microtype}
\usepackage{graphicx}
\usepackage{subfigure}

\usepackage{url}
\usepackage{subfigure}
\usepackage{wrapfig}
\usepackage{multirow}

\usepackage{mathtools}
\usepackage{amsmath}
\usepackage{amssymb}
\usepackage{mathtools}
\usepackage{amsthm}

\usepackage[ruled,vlined]{algorithm2e} 
\usepackage{colortbl}
\usepackage{array}
\usepackage{colortbl}

\definecolor{mygray}{gray}{.88}
\newcolumntype{g}{>{\columncolor[gray]{0.88}}c}

\usepackage[capitalize,noabbrev]{cleveref}

\theoremstyle{plain}
\newtheorem{theorem}{Theorem}[section]

\theoremstyle{definition}
\newtheorem{definition}[theorem]{Definition}
\newtheorem{assumption}[theorem]{Assumption}
\newtheorem{insight}[theorem]{Insight}
\theoremstyle{remark}
\newtheorem{remark}[theorem]{Remark}

\newcommand{\sysname}{\texttt{MOON} }

\title{Embracing Unknown Step by Step: \\Towards Reliable Sparse Training in Real World}

\author{\name Bowen Lei \email bowenlei@stat.tamu.edu \\
      \addr Department of Statistics\\
      Texas A\&M University
      \AND
      \name Dongkuan Xu \email dxu27@ncsu.edu \\
      \addr Department of Computer Science\\
      North Carolina State University
      \AND
      \name Ruqi Zhang \email ruqiz@purdue.edu\\
      \addr Department of Computer Science\\
      Purdue University
      \AND
      \name Bani Mallick \email bmallick@stat.tamu.edu \\
      \addr Department of Statistics\\
      Texas A\&M University}

\def\month{03}  
\def\year{2024} 
\def\openreview{\url{https://openreview.net/forum?id=Db5c3Wxj9E}} 

\begin{document}

\maketitle

\begin{abstract}
  Sparse training has emerged as a promising method for resource-efficient deep neural networks (DNNs) in real-world applications. However, the reliability of sparse models remains a crucial concern, particularly in detecting unknown out-of-distribution (OOD) data. This study addresses the knowledge gap by investigating the reliability of sparse training from an OOD perspective and reveals that sparse training exacerbates OOD unreliability. The lack of unknown information and the sparse constraints hinder the effective exploration of weight space and accurate differentiation between known and unknown knowledge. To tackle these challenges, we propose a new unknown-aware sparse training method, which incorporates a loss modification, auto-tuning strategy, and a voting scheme to guide weight space exploration and mitigate confusion between known and unknown information without incurring significant additional costs or requiring access to additional OOD data. Theoretical insights demonstrate how our method reduces model confidence when faced with OOD samples. Empirical experiments across multiple datasets, model architectures, and sparsity levels validate the effectiveness of our method, with improvements of up to \textbf{8.4\%} in AUROC while maintaining comparable or higher accuracy and calibration. This research enhances the understanding and readiness of sparse DNNs for deployment in resource-limited applications. Our code is available on: \url{https://github.com/StevenBoys/MOON}.
\end{abstract}

\input{sections/_s1_introduction}

\input{sections/_s2_related_work}

\input{sections/_s3_method}

\input{sections/_s4_theory.tex}

\input{sections/_s5_experiments}

\input{sections/_s6_conclusion}

\bibliography{tmlr}
\bibliographystyle{tmlr}

\newpage
\appendix

\input{sections/_a_Appendix.tex}

\input{sections/_b_Exp_del.tex}

\input{sections/_c_Theoty.tex}

\input{sections/_d_limit.tex}

\input{sections/_e_flops.tex}

\end{document}

%% file: math_commands.tex
\usepackage{amsmath,amsfonts,bm}

\def\eqref#1{equation~\ref{#1}}

\def\1{\bm{1}}

\DeclareMathAlphabet{\mathsfit}{\encodingdefault}{\sfdefault}{m}{sl}
\SetMathAlphabet{\mathsfit}{bold}{\encodingdefault}{\sfdefault}{bx}{n}



%% file: sections/_s1_introduction.tex
\section{Introduction}

Sparse training is one popular method for achieving resource efficiency in deep neural networks (DNNs), and it is receiving growing attention when considering the resource-limited real-world applications of DNNs~\citep{bellec2017deep, evci2020rigging, yuan2021mest}. By maintaining sparse weights throughout the training process, sparse training accelerates the training of DNNs, saves training memory, and produces sparse models with dense performance levels~\citep{mocanu2018scalable, dettmers2019sparse, evci2020rigging, liu2022comprehensive}, which has been applied to an increasing number of tasks~\citep{yuan2021mest, sokar2021dynamic, bibikar2022federated}. There remains, however, a critical question that needs to be answered before sparse training can be safely introduced in real-world applications: how reliable the sparse model is in the real world?

Reliable models in real-world scenarios possess the ability to recognize situations in which they are likely to be wrong, such as identifying unknown OOD data, and refrain from providing incorrect responses~\citep{wang2021energy, vaze2021open, du2022vos}. However, the investigation of OOD detection within the context of sparse training remains unexplored, creating an important knowledge gap that hinders a comprehensive assessment of the readiness of DNNs for deployment. Models with weak OOD detection are prone to feign understanding and provide arbitrary guesses on unrecognizable OOD data, leading to safety concerns~\citep{liu2020energy, hsu2020generalized, du2022vos, wang2021energy} in scenarios, e.g., automated healthcare and self-driving cars~\citep{rasheed2022explainable, de2022toward}.

\begin{figure}[!t]
\setlength{\abovecaptionskip}{0.0cm}
\vspace{-1.4cm}
\begin{center}
\subfigure[CIFAR-10, RigL]{\includegraphics[scale=0.14]{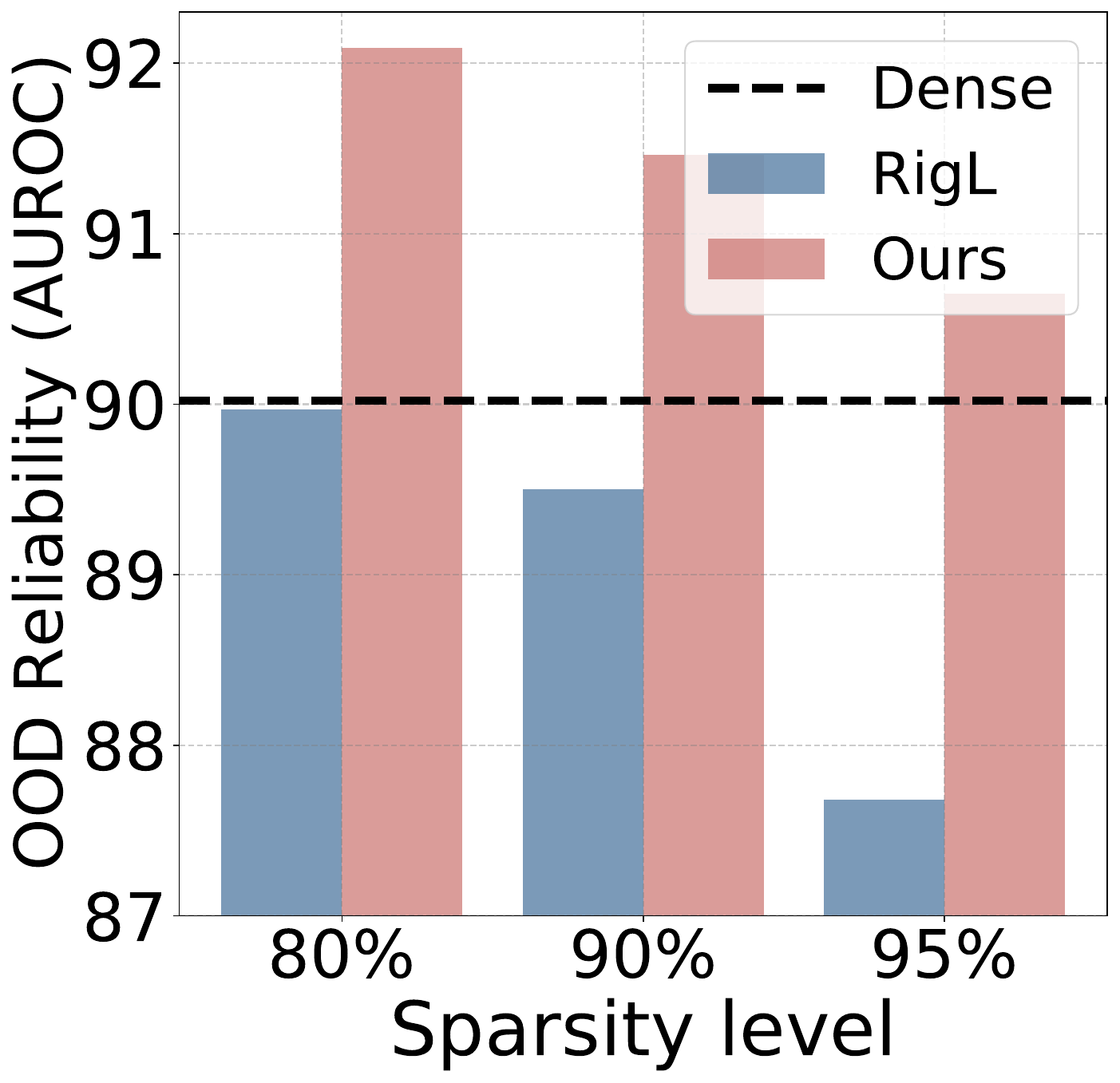}}
\subfigure[CIFAR-10, SET]{\includegraphics[scale=0.14]{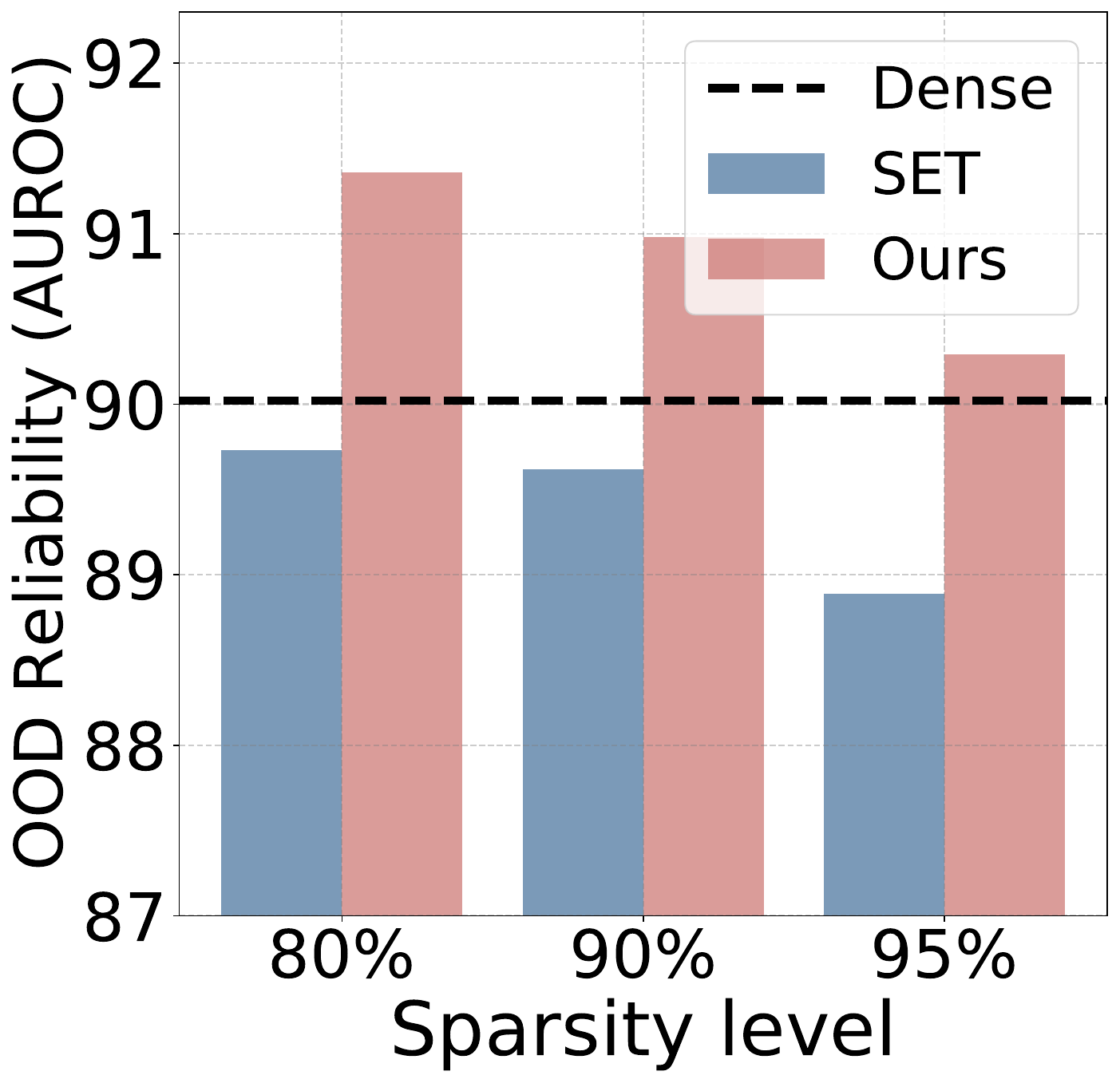}}
\subfigure[CIFAR-100, RigL]{\includegraphics[scale=0.14]{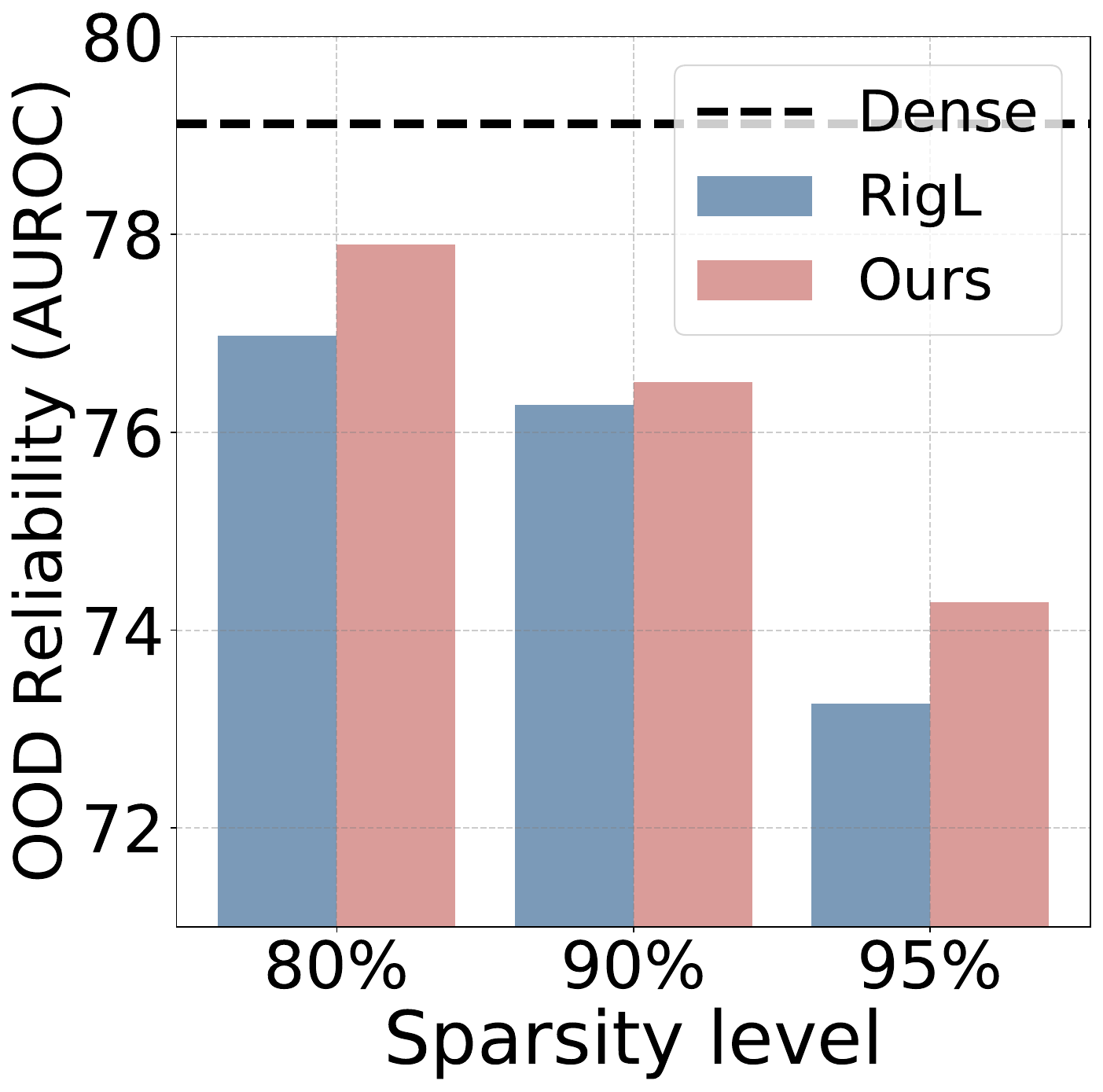}}
\subfigure[CIFAR-100, SET]{\includegraphics[scale=0.14]{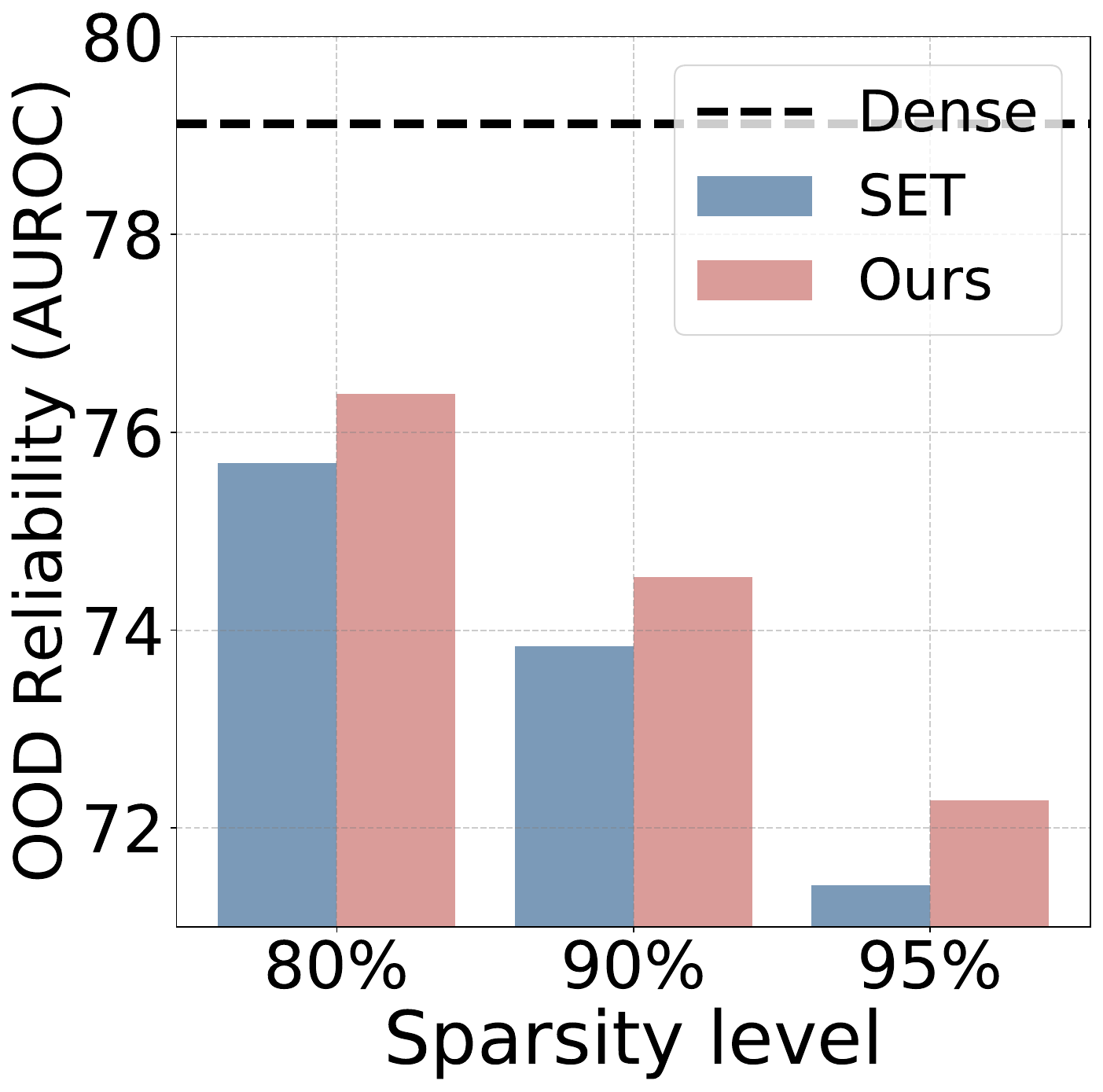}}
\caption{OOD reliability (measured by AUROC (\%), the higher the better) of the ResNet-18 produced by dense and sparse training (RigL \& SET) on CIFAR-10/100. Compared to dense training (black line), sparse training (blue bar) has a smaller AUROC, indicating sparse training exacerbates the unreliability on OOD data. Our \sysname (red bar) improves AUROC and OOD detection.}
\label{fig: intro}
\end{center}
\vspace{-0.8cm}
\end{figure}

In this work, we investigate for the \emph{first} time the reliability of sparse training from the OOD perspective, and find that sparse training exacerbates the OOD unreliability. Figure~\ref{fig: intro} compares AUROC (i.e., a metric for OOD detection, the higher the better)~\citep{yang2022openood} between dense and sparse training, e.g., RigL~\citep{evci2020rigging} and SET~\citep{mocanu2018scalable}. The AUROC is smaller for sparse training (blue bar) compared to dense training (black line), indicating that sparse training weakens the OOD detection capacity. In order to address the OOD unreliability of sparse training, we raise the following questions:

\textbf{\textit{Question 1.}} \textit{Why sparse training exacerbates the OOD unreliability in DNNs?}

The underlying determinant of weak OOD detection is the lack of unknown information, and sparse constraints hinder the effective communication of unknown concepts with DNNs. In order to recognize OOD samples, the model needs to distinguish between the known and the unknown. However, it has been shown that DNNs tend to focus on the known due to the close nature of the training process and ignore the rest~\citep{bendale2016towards, padhy2020revisiting, wang2021energy}. There are studies trying to encourage DNNs to learn more about the unknown. But they usually focus on dense training. When moving to sparse training, the sparsity cuts off update routes and produces spurious local minima~\citep{evci2019difficulty, sun2021effectiveness, he2022sparse}, which makes the exploration of the weight space more challenging, and thus more difficult to search for and find reliable models~\citep{lei2023calibrating}. Thus, the sparse training exacerbates the OOD unreliability in DNNs.

\textbf{\textit{Question 2.}} \textit{How can we leverage unknown information under sparse constraints to effectively guide weight space exploration to OOD-reliable state?}

To improve OOD reliability, it is crucial to utilize unknown information to guide model training. The sparsity constraint increases the difficulty of exploring the weight space and prevents the unknown information from guiding the model training to an OOD-reliable state. Efforts have been made to exploit unknown information in dense DNNs~\citep{thulasidasan2019combating, wang2021energy, roy2022does}. However, DAC~\citep{thulasidasan2019combating} is limited to the treatment to label noise and lacks a comprehensive guide to OOD detection. EOW-Softmax~\citep{wang2021energy} only indirectly and inadequately exploits the unknown and incurs considerable additional costs. Moreover, HOD~\citep{roy2022does} require access to OOD data, which is often a non-trivial task. To address these issues and adapt to the challenges in sparse training, we propose a new outlier-exposure-free training loss that extracts unknown information from hard ID samples and then directly tells the DNNs this unknown information, which gives an effective guide for weight space exploration under sparse constraints with only a slight increase in computational and storage burden.

\textbf{\textit{Question 3.}} \textit{How can we address the confusion between known and unknown information in sparse training to avoid providing misleading guidance for weight space exploration?}

Despite the possibility of communicating unknown information, sparse DNNs are more likely to confuse known and unknown information, resulting in poor performance and reliability. DNNs show performance degradation in sparse training with high sparsity~\citep{mocanu2018scalable, evci2020rigging, lei2023calibrating}, especially in the early training stage, which indicates the challenge of understanding knowable content under sparsity constraints. Therefore, to avoid confusion and provide proper guidance for training, we also design an auto-tuning strategy that allows the DNN to mainly discover the known in the early stage and embrace the unknown in the later stage step by step. In addition, we propose a voting scheme that combines information from multiple models for a more comprehensive understanding of the known and unknown.

In summary, our contributions are summarized as follows:

\begin{itemize}
    \item We for the first time investigate how sparse training influences OOD reliability and analyze the hidden reasons. We find that sparse training exacerbates the OOD unreliability of DNNs.
    \item We propose a new \underline{mo}del-agnostic unkn\underline{o}w\underline{n}-aware sparse training method (\texttt{MOON}) to improve real-world reliability, which allows the model to be aware of what it does not know via a loss modification, auto-tuning strategy, and a voting scheme at little additional cost.
    \item We provide theoretical insights on how our \sysname method reduces the confidence of the model when faced with OOD samples and improves OOD reliability.
    \item Empirically, we conduct extensive experiments on multiple benchmark datasets, model architectures, and sparsity. Our \sysname improves the AUROC by up to \textbf{8.4\%} and successfully upholds the accuracy, calibration, and other pertinent metrics associated with sparse training, thereby complementing existing research and further enhancing the efficacy and applicability of the sparse training paradigm.
\end{itemize}

%% file: sections/_s2_related_work.tex
\section{Related Work}

\textbf{Sparse Training.} As models continue to grow in size, there is an increasing focus on sparse training, where sparse weights are maintained during training to accommodate resource-limited real-world deployments~\citep{mocanu2018scalable, bellec2018deep, huang2023neurogenesis}. To find a sparse model with good performance, the sparse topology is updated by pruning and growth steps after every fixed number of iterations~\citep{evci2020rigging, huang2022dynamic}. A variety of sparse training methods have been investigated and various pruning and growth criteria, such as weights and gradient magnitude, have been developed~\citep{mostafa2019parameter, dettmers2019sparse, evci2020rigging, jayakumar2020top, liu2021we, ozdenizci2021training, zhou2021effective, schwarz2021powerpropagation, yin2022superposing, lei2023balance}. Existing work mainly targets saving more resources and at the same time maintaining accuracy. However, despite saving resources, sparse training can result in more difficult training due to the cut-off update routes and the generation of spurious local minima, raising safety concerns on both ID and OOD data.

\textbf{Confidence Calibration.} DNNs' reliability on ID data is increasingly important, which usually refers to whether the confidence of DNNs on ID data is well calibrated~\citep{guo2017calibration, nixon2019measuring, zhang2020mix, wu2023towards}. Prior work has shown that DNNs tend to be over-confident~\citep{guo2017calibration, rahaman2021uncertainty, patel2022improving, zhu2023rethinking}, misleading humans and raising safety concerns. Research has been conducted to improve confidence calibration, where widely-used methods include temperature scaling~\citep{guo2017calibration}, Mixup~\citep{zhang2017mixup}, label smoothing~\citep{szegedy2016rethinking}, and Bayesian methods~\citep{gal2016dropout, ashukha2020pitfalls}. Recently, ID reliability in sparse training has also been studied in Sup-tickets~\citep{yin2022superposing} and CigL~\citep{lei2023calibrating}.

\textbf{OOD Detection \& Open-set Classification.} For real-world tasks, reliability on OOD data, i.e., OOD detection capability, is also critical due to the presence of unknown categories. It is also referred to as open-set classification when ID classification is required as well~\citep{yang2021generalized, vaze2021open}. When these unknown categories are given, a reliable model should detect them and refuse to answer, rather than giving random guesses. Research has been devoted to OOD detection, including post-processing methods~\citep{hendrycks2016baseline, liang2017enhancing, liu2020energy}, training time regularization~\citep{devries2018learning, hsu2020generalized, du2022vos}, and training with outlier exposure~\citep{hendrycks2018deep, yu2019unsupervised, yang2021semantically}. However, existing OOD reliability studies have mainly focused on dense training and have not yet fully explored sparse training settings.

\textbf{OOD reliability of sparse DNNs.} Despite the lack of exploration of sparse training, there are studies on the OOD reliability of sparse DNNs. The effect of random pruning on the sparse DNNs with fixed sparsity patterns is studied~\citep{liu2022unreasonable}, while sparse training has dynamic sparsity patterns that provide better performance compared to fixed patterns. It is also investigated how to obtain deep ensembles that outperform dense models through sparse training, which requires a non-trivial additional cost~\citep{liu2021deep}. Thus, generating a single sparse network through sparse training remains irreplaceable when resources are limited. In addition, efforts have been devoted to OOD reliability of pruning methods, which start with a dense model and gradually prune the weights, reducing the difficulty of exploring the weight space~\citep{ayle2022robustness, cheng2023average}. In contrast, sparse training maintains a high level of sparsity throughout the training process, which cuts off a large portion of the route and produces more spurious local minima, thus making weight space exploration more challenging and having different properties from pruning methods~\citep{chen2022can}.

\textbf{Extra Dimension.} Adding an extra dimension to the output of the K-way classification model has been studied and used to relax closed-world setting to open-world setting~\citep{hsu2020generalized, mozannar2020consistent, wang2021energy, verma2022calibrated, liu2023good}. Specifically, the close-world setting assumes the test data has the same distribution as the training data, which is not always true in the real world. The open-world setting, on the other hand, allows for unknown classes in the test data. Previous work has used extra dimension to abstain from making decisions on noisy labels~\citep{thulasidasan2019combating} or to improve OOD detection in dense DNNs~\citep{wang2021energy, roy2022does}. However, they do not directly exploit the unknown information and incur extra costs~\citep{wang2021energy}, or require exposure to OOD data~\citep{roy2022does}. In addition, they all ignore how to add extra dimensions in sparse training.

%% file: sections/_s3_method.tex
\section{Method}

We propose a new sparse training method \sysname to improve the real-word reliability of DNNs, i.e., to provide effective detection for OOD data under sparsity constraints with comparable or higher ID accuracy and reliability. Importantly, \sysname can be seamlessly integrated as a plug-in method and combined with existing OOD detection methods. Specifically, we add an extra dimension in the output probability and propose a new unknown-aware loss with an auto-tuning strategy to gradually encourage DNNs to consider more unknown information. Then, we combine information from multiple models into the output model via a voting scheme so that the output model has a comprehensive view of what is known and what is unknown, providing reliable predictions.

\subsection{Tell Model What It Doesn't Know}\label{sec:3-1}

In general, given the data $\{x_i, y_i\}_{i=1}^N$ where $y \in \{1,\cdots, K\}$, the model is forced to classify new input data into K predefined classes by the widely used cross-entropy loss, even if the model is uncertain about the class of the samples or has never seen this class before~\citep{wang2021energy}. We argue that ignorance of unknown information during weight space exploration, i.e., the sample for which the model gives incorrect predictions ($\widehat{y}_i \neq y_i$), is the cause of this unreliability. Sparse constraints further make the exploration more challenging and thus require better guidance from unknown information. To solve these problems, we utilize K+1-way formulation and propose a new unknown-aware loss to tell the model what is unknown to it during sparse training. Unlike existing methods, we directly tell DNNs the unknown, providing an effective guide to cope with the challenges posed by sparsity.

\textbf{K+1-way formulation:} To store the unknown information, inspired by previous work on dense DNNs with extra dimensions ~\citep{thulasidasan2019combating, wang2021energy, roy2022does}, we add an extra dimension in the softmax formulation, allowing the model to provide a $K+1$-dimensional output probability $f(x_i)=(p_1,\cdots, p_{K}, p_{K+1})$ for each $x_i$, unlike the previous $K$-way formulation. The first $K$ dimensions $(p_1,\cdots, p_{K})$ represent the model's beliefs about class attribution, while the $K+1$-th dimension $p_{K+1}$ is designed to store unknown information. Ideally, the model gives larger probabilities for the true class dimension $p_{y_i}$ when the model knows the input samples very well. Otherwise, the model provides larger $p_{K+1}$ and smaller $(p_1,\cdots, p_{K})$ to avoid unreliable random guesses in the $K$ classes.

\textbf{Unknown-aware Loss Modification:} To incorporate unknown information into the model, we design an unknown-aware loss shown in Eq.~(\ref{eq:loss-moon}). On the one hand, when the model knows the sample $x_i$ well and makes correct predictions (i.e., $\widehat{y}_i=y_i$), we choose the broadly-used cross-entropy loss for $x_i$ to encourage larger $p_{y_i}$. On the other hand, if the model does not have enough understanding of harder sample $x_i$ and makes wrong predictions (i.e., $\widehat{y}_i \neq y_i$), we use a new-designed loss to increase reliability. Specifically, in the new loss, we add a larger weight $(1 + \frac{w}{1 + w p_{K+1}})$ to encourage the model to focus more on the samples it predicts incorrectly. To reduce the loss, the model can determine the correct class (i.e., give a larger $p_{y_i}$) or provide a larger $p_{K+1}$ such that $(p_1, \cdots, p_{K})$ becomes smaller, indicating that the model is uncertain about its prediction.
\begin{equation}\label{eq:loss-moon}
    L(x_i) = 
    \begin{cases}
    - \log p_{y_i}& \widehat{y}_i=y_i\\
    - (1 + \frac{w}{1+w p_{K+1}}) \log p_{y_i}& \widehat{y}_i \neq y_i
    \end{cases}
\end{equation}

\subsection{Do not Frustrate Model in Early Stage}

We find that telling the model too much about the unknowns at the beginning can lead to more severe unreliability (see Section~\ref{subsec:abla}), especially in cases where the weight space is insufficiently explored~(e.g., highly sparse models). It is as if too much emphasis on unknowns frustrates the model and prevents it from finding out what it can know. Thus, inspired by~\citep{thulasidasan2019combating}, we design an auto-tuning strategy for $w$ in the unknown-aware loss (i.e., Eq.~(\ref{eq:loss-moon})) to step-by-step inform the model about the unknown information, which is outlined in Algorithm~\ref{alg:w-tuning}. In this way, $p_{y_i}$ and $p_{K+1}$ can better help the model learn and store both the known and unknown information.

\begin{wrapfigure}{R}[0pt]{0.5\textwidth}
\vspace{-0.5cm}
\begin{minipage}{0.5\textwidth}
\begin{algorithm}[H]
\DontPrintSemicolon
  \SetAlgoLined
  \KwIn {Total epoch $T$, current epoch $t$, unknown-free epoch $T_e$, probability $\{p_{i}\}_{i=1}^{K+1}$, unknown-free loss $L$, init factor $r$, final weight $w_f$, smoothing factor $\alpha$.}
  \KwOut {The weight $w$ for the unknown-aware loss.}
  Initialization:\;
  $w=0$, $\widetilde{\beta}=0$, and $\delta=-1$\; 
  
  \uIf{$t <  T_e$}{
    Prepare information for the initial weight: \;
    
    $\beta = (1-p_{K+1}) L(x_i)$ \; 
    
    Use moving average $\widetilde{\beta} =(1-\alpha)\widetilde{\beta} + \alpha\beta $ \;
  }
  \uElseIf{$\delta=-1$}{
    Calculate the initial weight $w_i=\frac{\widetilde{\beta}}{r}$ and $w = w_i$\;
    
    Calculate the step size $\delta = \frac{w_f - w_i}{T - T_e}$ of the scheduler \;
  }
  \Else{
    Calculate weight at epoch $t$: $w=w_i + (t-T_e) \cdot \delta$ \;
  }
  \caption{$w$ Auto-tuning Scheduler}\label{alg:w-tuning}
\end{algorithm}
\end{minipage}
\vspace{-0.9cm}
\end{wrapfigure}

\textbf{In the early stage} (i.e., epoch $t <= T_e$), we fix $w$ to zero so that the new loss is equivalent to the cross-entropy loss and no unknown information is emphasized. DNNs usually learn easy samples in the early stage of training and tend to fit the hard ones later~\citep{wang2021rethinking}. 
Thus, in the beginning, the model does not learn the data well and usually makes many wrong predictions. With insufficient exploration of the weight space, a large $w$ will unnecessarily put more emphasis on harder samples and prevent the model from learning easy samples, leading to poor reliability. The model is like a new learner, and it is advisable not to point out its mistakes too much, in case it gets frustrated.

\textbf{After the early stage} (i.e., epoch $t > T_e$), we start telling the model what is unknown. We keep $w$ as a constant at each epoch and increase it from the initial value $w_i$ to the final value $w_f$ after each epoch via a linear scheduler.
\textbf{The initial $w_i$} is based on the $\beta=(1-p_{K+1}) L(x_i)$ and calculated in the early stage. On the one hand, if we get smaller $p_{K+1}$ and larger $L(x_i)$, we can know that the model hasn't learned the data well while it is over-confident about its wrong predictions. Thus, we will set a larger $\beta$ and $w_i$, in this case, to emphasize more on the unknown for reliability. On the other hand, if we get larger $p_{K+1}$ and smaller $L(x_i)$, we can know that the model learns the data well and is cautious about its output. Therefore, we can encourage the model to find out more about what it can know with smaller $\beta$ and $w_i$. For the initial factor $r$, it is used to control the initial value $w_i$ to avoid $w$ being too large.
\textbf{For the final $w_f$}, it is a pre-determined hyper-parameter, which is usually larger than $w_i$. At each epoch after $T_e$, we set a linear increase scheduler $w=w_i + (t-T_e) \cdot \delta$ where the step size $\delta=(w_f - w_i)/(T - T_e)$, allowing more attention on the unknown at the later stage of the training. 

\subsection{Vote for Known and Unknown: Weight Averaging}

We further design a voting strategy that combines information from multiple models to give the output model a more comprehensive sense of what it knows and what it does not know. Specifically, at the end of the proposed unknown-aware training (e.g., after 80\% training epochs), we collect models at each epoch and use the weight averaging method~\citep{izmailov2018averaging, wortsman2022model} to uniformly average them into one output model. In this way, we let multiple models decide what is known and unknown together, allowing a fuller understanding of the learned knowledge and more reliable predictions.

%% file: sections/_s4_theory.tex
\section{Theoretical Insight}

We provide a couple of theoretical insights to demonstrate that our model-agnostic unknown-aware sparse training method \sysname can produce more reliable predictions in the face of out-of-distribution (OOD) data. The unknown information is extracted from hard in-distribution (ID) data, which explains the outlier-exposure-free property of \texttt{MOON}. Detailed proofs are presented in the Appendix. Without loss of generality, suppose the goal is a two-way classification based on $\{x_i, y_i\}_{i=1}^N$ where $y_i \in \{1, 2\}$.  We have a deep neural network $f=g\circ h$, which can be seen as a composition of feature mapping function $h$ and the softmax classification function $g$. 

We first show how we can extract the unknown from the hard ID data. The following analysis is based on Assumption~\ref{assup:gp}, which has been widely used in several studies of OOD detection~\citep{lee2018simple, ahuja2019probabilistic, morteza2022provable}.

\begin{assumption}
(Gaussian Mixture Feature Space): The feature mapping function $h$ maps the input ID data to a Gaussian mixture $v_1\mathcal{N}(\mu_1, \Sigma_1) + v_2\mathcal{N}(\mu_2, \Sigma_2)$. Specifically, when $y=1$, we have $h(x)\sim \mathcal{N}(\mu_1, \Sigma_1)$. And when $y=2$, we have $h(x)\sim \mathcal{N}(\mu_2, \Sigma_2)$.
\label{assup:gp}
\end{assumption}

We define the definition of unreliability for a set of ID data.

\begin{definition} (Unreliability)\label{def:unre}
For samples whose features are around $h(x_0)$, i.e., $D(\epsilon_0)=\{(x, y); ||h(x)-h(x_0)||<\epsilon_0$\}, the predictions are unreliable around $x_0$ if for any $0<\epsilon < \epsilon_0$, we can find $\eta>0$ such that
\begin{equation}
    \mathbf{E}_{D(\epsilon)} [\max_{c\in \{1,2\}} \mathcal{N}(h(x) ; \mu_{c}, \Sigma_{c})] - \mathbf{E}_{D(\epsilon)} [1\{ \hat{y}=y \}] > \eta.    
\end{equation}
\end{definition}

\begin{remark}
Definition~\ref{def:unre} gives a mathematical expression for the ID unreliability in feature space (i.e., the discrepancy between confidence and accuracy) around $h(x_0)$. The first term $\mathbf{E}_{D(\epsilon)} [\max_{c\in \{1,2\}} \mathcal{N}(x ; \mu_{c}, \Sigma_{c})]$ denotes the expected confidence around $h(x_0)$, and the second term $\mathbf{E}_{D(\epsilon)} [1\{ \hat{y}=y \}]$ refers to the expected accuracy. For a threshold (e.g., $\eta$), we consider predictions to be unreliable if the discrepancy between confidence and accuracy cannot be reduced below $\eta$ no matter how we alter $\epsilon$. This extends the definition of unreliability from the global aspect to the local region around $h(x_0)$ in feature space and helps to describe unreliability within hard samples.
\end{remark}

\begin{insight}\label{pro:unrelia}
(Unreliability) Suppose we have $(x_1, y_1)$ from class 1 and $D_2 = \{(x, y); ||h(x)-h(x_1)||<\epsilon, y=2 \}$ from class 2. Then, unreliability can occur around  $h(x_1)$.
\end{insight}

\begin{remark}
The setting in Insight~\ref{pro:unrelia} is common for the hard ID samples, resulting in unreliability. Samples with large feature differences within two classes are easy to classify, with both high accuracy and confidence. While samples with minor differences can puzzle the model, with low accuracy and high confidence, resulting in over-confidence. 
\end{remark}

\begin{insight}\label{pro:reli}
(Hard-ID Reliability) Suppose we have the same $(x_1, y_1)$ and $D_2$ as in Insight~\ref{pro:unrelia}. If the model is trained with our \sysname method and the extra dimension successfully stores the unknown information, the unreliability can be solved, i.e., for any $\eta>0$, we can find $0<\epsilon < \epsilon_0$ such that
\begin{equation}
    \mathbf{E}_{D(\epsilon)} [\max_{c\in \{1,2\}} \mathcal{N}(h(x) ; \mu_{c}, \Sigma_{c})] - \mathbf{E}_{D(\epsilon)} [1\{ \hat{y}=y \}] < \eta.    
\end{equation}
\end{insight}

\begin{remark}
Insight~\ref{pro:reli} shows that our \sysname method reduces the confidence level and decreases the discrepancy, solving the unreliability issue when the model is faced with some hard ID samples. 
\end{remark}

We further show how OOD detection can benefit from the unknown information from hard ID data.

\begin{insight}\label{pro:ood-reli}
(OOD Reliability) Suppose we achieve Hard-ID Reliability in Insight~\ref{pro:reli} with our \sysname method, we can have lower confidence on OOD data, implying stronger OOD detection.
\end{insight}

\begin{remark}
Insignt~\ref{pro:ood-reli} show that OOD detection can be enhanced by the unknown information extracted from the hard ID data. Existing work has found that ID data contain information for the OOD samples~\citep{du2022vos}. It is found that ID samples from the low-likelihood region of the class-conditional distribution, i.e., hard-to-detect ID data, are similar to OOD samples in the feature space, which is similar to the scenario described in Insight~\ref{pro:unrelia}. Although we do not use OOD data in the training, these hard-to-detect ID samples can be viewed as pseudo-OOD data. With our \sysname method, the model knows which hard ID samples it does not know and provides lower confidence for these samples (i.e., pseudo-OOD data). Since the real OOD data are neighbors of the pseudo-OOD data in the feature space, the model also gives a low confidence for the real OOD data.
\end{remark}

%% file: sections/_s5_experiments.tex
\section{Experiments}

We perform a comprehensive empirical evaluation of our \sysname, where we add our \sysname to the training of multiple benchmark datasets, model architectures, and sparsities, and compare it with the original sparse training results.

\textbf{Datasets \& Model Architectures:} 
For in-distribution (ID) data, we include four benchmark datasets: MNIST~\citep{deng2012mnist}, CIFAR-10 and CIFAR-100~\citep{krizhevsky2009learning} and ImageNet-2012~\citep{russakovsky2015imagenet}. For out-of-distribution (OOD) data, we follow the setups of OpenOOD~\citep{yang2022openood} (more details in Appendix B.1).
For model architectures, we choose ResNet-18 to train MNIST, CIFAR-10, and CIFAR-100, and choose ResNet-50 to train ImageNet-2012~\citep{he2016deep}. We repeat all experiments 3 times and calculate the mean and standard deviation.

\textbf{Training Settings:} For sparse training, we study two widely-used methods, e.g., RigL~\citep{evci2020rigging} and SET~\citep{mocanu2018scalable}, and a variety of sparsities, including 80\%, 90\%, 95\%, and 99\%, which sufficiently reduces the cost and is of great interest in real-world deployments. We also examine the performance of dense training to show the broad applicability of our method (see Appendix A.3). 

\textbf{Implementations:} We follow the settings of OpenOOD~\citep{yang2022openood}. We choose SGD with momentum and use the cosine annealing learning rate scheduler. We train all the models for 100 epochs. For the batch size, we set it to 128 in MNIST, CIFAR-10, and CIFAR-100, and 32 in ImageNet.

\textbf{Comparison Metrics:} 
To measure OOD detection capability, we chose AUROC and FPR-95~\citep{yang2022openood}. For AUROC, it calculates the area under the receiver operating characteristic (ROC) curve, where a larger AUROC suggests better OOD detection. For FPR-95, it measures the false positive rate (FPR) at a true positive rate (TPR) of 95\%, where a lower FPR-95 implies better OOD detection. To compare ID reliability, we choose ECE~\citep{guo2017calibration} and test accuracy. ECE measures the discrepancy between confidence and true accuracy, with lower ECE indicating higher ID reliability.

\subsection{OOD Detection in Sparse Training}

In this section, we show that our \sysname can improve OOD detection in sparse training, including 80\%, 90\%, 95\%, and 99\% sparsities using RigL. Since our \sysname requires no additional OOD data and training, to fairly compare the OOD reliability, we incorporate our \sysname into the widely-used post-process OOD detection methods MSP~\citep{hendrycks2016baseline} and compare the results MSP.

\linespread{1.2}
\begin{table*}[!t]
\vspace{-0.0cm}
\setlength\tabcolsep{4.5pt} 
\caption{Comparison of OOD detection by AUROC (\%) ($\uparrow$) between \sysname+MSP (\textit{M-MSP}) and MSP for CIFAR-10 in sparse training. Our \sysname leads to larger AUROC on each OOD data of CIFAR-10, showing its ability to improve reliability on OOD data.}
\label{tb: ood-sparse}
\begin{center}
\begin{scriptsize}
\begin{sc}
\begin{tabular}{cgggggggggg}
\toprule \toprule
\rowcolor{white} &  &  CIFAR-100 & TIN   & NearOOD & & MNIST & SVHN  & Texture & Places365 & FarOOD \\
\hline
\rowcolor{white} \multirow{2}{*}{80\%} & MSP        &   87.43 (0.2) & 88.92 (0.2) & 88.17 (0.2) & & 92.92 (0.8) & 91.81 (0.7) & 89.35 (0.3) & 89.44 (0.2) & 90.88 (0.5) \\
~ & \textit{M-MSP} &   \textbf{89.93} (0.1) & \textbf{91.14} (0.1)& \textbf{90.53} (0.1)& & \textbf{95.00} (0.7)   & \textbf{93.82} (0.5) & \textbf{91.80} (0.2)  & \textbf{90.82} (0.3) & \textbf{92.86} (0.4) \\
\rowcolor{white} \multirow{2}{*}{90\%} & MSP        &   87.78 (0.2) & 89.12 (0.1) & 88.45 (0.1) & & 91.82 (0.9) & 91.48 (0.5) & 87.90 (0.3)  & 88.93 (0.2) & 90.03 (0.4)\\
~ & \textit{M-MSP}   & \textbf{88.92} (0.2) & \textbf{90.34} (0.2) & \textbf{89.63} (0.2) & & \textbf{95.39} (0.8) & \textbf{93.09} (0.4)& \textbf{91.09} (0.4) & \textbf{89.92}(0.2) & \textbf{92.37} (0.4) \\ 
\rowcolor{white} \multirow{2}{*}{95\%} &MSP          & 86.04 (0.1) & 87.30 (0.1)  & 86.67 (0.1) & & 93.74 (0.6) & 86.09 (0.8) & 85.26 (0.3) & 87.68 (0.2) & 88.19 (0.5)\\
~ & \textit{M-MSP}   & \textbf{88.04} (0.1) & \textbf{89.39} (0.2)& \textbf{88.71} (0.1) & & \textbf{94.76} (0.7) & \textbf{93.35} (0.8) & \textbf{89.86} (0.3) & \textbf{88.51} (0.3) & \textbf{91.62} (0.5) \\
\rowcolor{white} \multirow{2}{*}{99\%} &MSP         & 80.94 (0.2) & 82.75 (0.2)& 81.84 (0.2) & & 88.34 (0.8) & 85.75 (0.7) & 83.43 (0.2) & 80.45 (0.2) & 84.49 (0.5)\\
~ & \textit{M-MSP} &   \textbf{83.75} (0.1) & \textbf{85.11} (0.1) & \textbf{84.43} (0.1) & & \textbf{91.96} (0.8) & \textbf{89.01} (0.8) & \textbf{86.44} (0.3) & \textbf{83.38} (0.2) & \textbf{87.70} (0.5)\\ \bottomrule
\bottomrule
\end{tabular}
\end{sc}
\end{scriptsize}
\end{center}
\vspace{-0.5cm}
\end{table*}

\textbf{For CIFAR-10 based models}, we examine its OOD detection capacity on two near OOD data, i.e., CIFAR-100~\citep{chandola2009anomaly} and TIN~\citep{han2022adbench}), and four far OOD data, i.e., MNIST~\citep{deng2012mnist}, SVHN~\citep{netzer2011reading}, Texture~\citep{ahmed2020detecting}, Places365~\citep{guo2017calibration}. As shown in Table~\ref{tb: ood-sparse}, the columns named ``NearOOD" and ``FarOOD" represent the average detection scores of the near OOD data and the far OOD data, respectively. As shown in Table~\ref{tb: ood-sparse}, the AUROC value drops with increasing sparsity, which implies the unreliability issues in sparse training. Our \sysname obtains a larger AUROC on each OOD data of CIFAR-10, where the improvement of AUROC can be up to 8.4\%. This indicates \sysname's ability to improve the reliability on OOD data in sparse training.

\begin{wrapfigure}{r}[0pt]{10cm}
\setlength{\abovecaptionskip}{-0.0cm}
\vspace{-0.8cm}
\begin{center}
\subfigure[AUROC (\%) ($\uparrow$)]{\includegraphics[width=0.245\textwidth,height=0.22\textwidth]{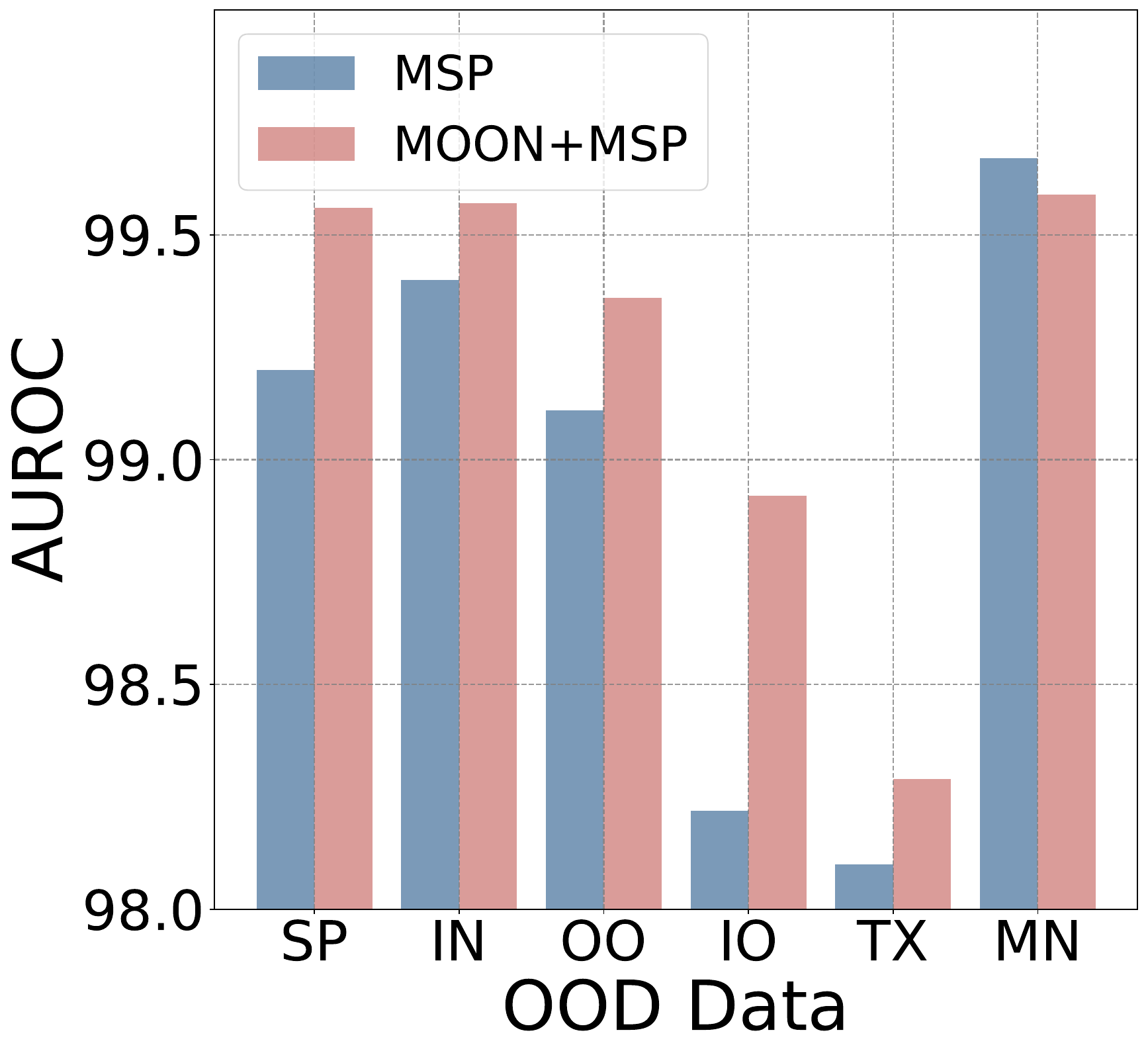}}
\subfigure[FPR-95 (\%) ($\downarrow$)]{\includegraphics[width=0.245\textwidth,height=0.22\textwidth]{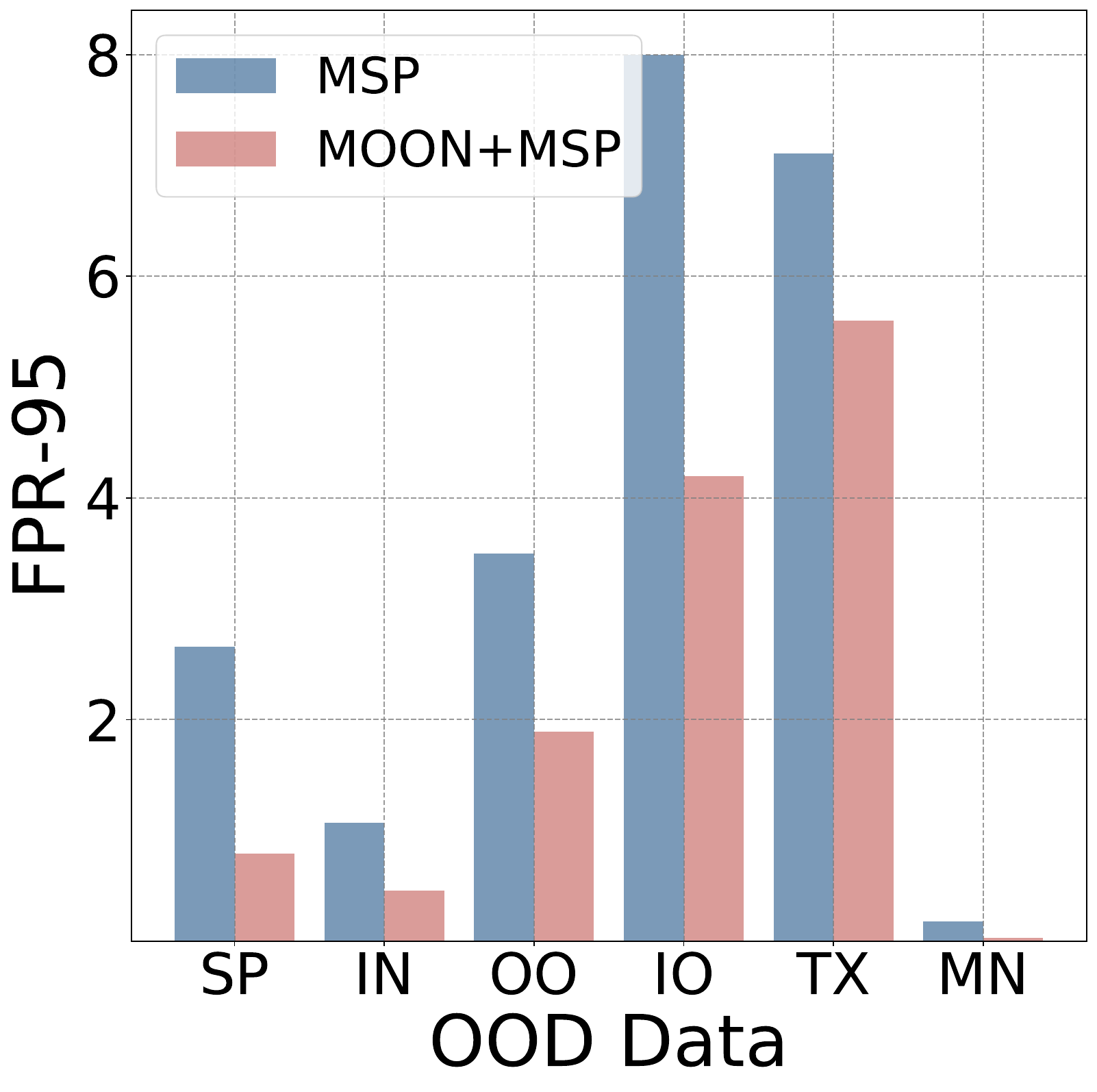}}
\caption{Comparison of OOD detection by AUROC (\%) ($\uparrow$) and FPR-95 (\%) ($\downarrow$) between \sysname+MSP and MSP for ImageNet-2012 using RigL (90\%). \sysname leads to larger AUROC and smaller FPR-95 on OOD data of ImageNet-2012, showing improved OOD reliability in sparse training.}
\label{fig: ood-img-rigl}
\end{center}
\vspace{-0.5cm}
\end{wrapfigure}

\textbf{For the model trained on ImageNet-2012}, we examine its OOD detection ability on four near OOD data, i.e., Species (SP)~\citep{torralba200880}, iNaturalist (IN)~\citep{shorten2019survey}, OpenImage-O (OO)~\citep{li2021cutpaste}, ImageNet-O (IO)~\citep{sun2022out}, and two far OOD data, i.e., Texture (TX)~\citep{ahmed2020detecting}, MNIST (MN)~\citep{deng2012mnist}. As shown in Figure~\ref{fig: ood-img-rigl}, the red and blue bars represent our \sysname and MSP, respectively. We can see that our \sysname provides larger AUROC and smaller FPR-95 almost on all of the OOD data of ImageNet-2012, showing its effective OOD detection in sparse training.

\textbf{To show the general applicability of our method}, we also incorporate our \sysname into other broadly-used post-processing OOD detection methods, e.g., ODIN~\citep{liang2017enhancing}, EBO~\citep{liu2020energy}, KNN~\citep{sun2022out}, and KLM~\citep{hendrycks2019scaling}, and compare the results with those using only ODIN and EBO. We take sparse models on CIFAR-10 using RigL as an example. First, we consider 99\% sparsity while considering ODIN and EBO as baselines. As shown in Table~\ref{tb: ood-other-c10}, our \sysname leads to larger AUROC and smaller FPR-95 on each OOD data compared to ODIN and EBO. Then, for KNN and KLM, we also experiment on different sparsity levels including 80\%, 90\%, 95\%, and 99\%. As shown in Table~\ref{tb:ood-new-b}, our \sysname leads to a larger AUROC on each OOD data compared to KNN and KLM. This shows that the ability of our \sysname to improve OOD reliability is consistent under sparsity constraints for different OOD detection methods.

\linespread{1.2}
\begin{table*}[!t]
\vspace{-0.0cm}
\setlength\tabcolsep{4.5pt} 
\caption{Comparison of OOD detection by AUROC (\%) ($\uparrow$) \& FPR-95 (\%) ($\downarrow$) between \sysname+ODIN (\textit{M-ODIN}), \sysname+EBO (\textit{M-EBO}) and baseline post process methods (i.e., ODIN and EBO) for CIFAR-10 using RigL (99\% sparsity). Our \sysname leads to larger AUROC and smaller FPR-95 on each OOD data, showing its ability to improve reliability on OOD data in sparse training.}
\label{tb: ood-other-c10}
\begin{center}
\begin{scriptsize}
\begin{sc}
\begin{tabular}{lgggggggggg}
\toprule \toprule
\rowcolor{white} & &  CIFAR-100 & TIN   & NearOOD & & MNIST & SVHN  & Texture & Places365 & FarOOD \\
\hline
\rowcolor{white} \multirow{4}{*}{\rotatebox{90}{AUROC}} & ODIN   & 84.09 (0.2) & 86.98 (0.3) & 85.54 (0.2) & & 96.82 (0.5) & 89.11 (0.7) & 87.47 (0.3) & 85.49 (0.2) & 89.72 (0.4) \\
~ &  \textit{M-ODIN}  & \textbf{90.59} (0.3) & \textbf{92.64} (0.2) & \textbf{91.61} (0.2) & & \textbf{99.21} (0.7) & \textbf{93.91} (0.8) & \textbf{93.26} (0.2) & \textbf{93.35} (0.2)& \textbf{94.93} (0.5)\\
~ &  \cellcolor{white} EBO   & \cellcolor{white} 83.73 (0.3) & \cellcolor{white} 86.93 (0.3) & \cellcolor{white} 85.33 (0.3) & \cellcolor{white}& \cellcolor{white} 94.68 (0.6) & \cellcolor{white} 85.71 (0.5) & \cellcolor{white} 85.09 (0.2) & \cellcolor{white} 85.44 (0.2) & \cellcolor{white} 87.73 (0.4) \\
~ & \textit{M-EBO}   & \textbf{85.64} (0.2) & \textbf{88.55} (0.3) & \textbf{87.09} (0.2) & & \textbf{97.10} (0.7)  & \textbf{87.35} (0.4) & \textbf{87.11} (0.3) & \textbf{88.55} (0.3) & \textbf{90.03} (0.4)\\ \hline
\rowcolor{white} \multirow{4}{*}{\rotatebox{90}{FPR-95}} & ODIN  &  67.99 (0.8) & 61.54 (0.5)& 64.76 (0.6) & & 19.40 (1.6)  & 65.32 (1.7) & 56.58 (0.6) & 62.56 (0.7) & 50.97 (1.1)\\
~ & \textit{M-ODIN}  & \textbf{61.23} (0.8) & \textbf{56.00} (0.6) & \textbf{58.62} (0.7) & & \textbf{11.58} (1.8)  & \textbf{57.67} (2.1) & \textbf{51.81} (0.5)  & \textbf{54.09} (0.7) & \textbf{43.79} (1.3)\\
~ & \cellcolor{white} EBO  &  \cellcolor{white} 69.42 (0.7) & \cellcolor{white} 61.15 (0.6) & \cellcolor{white} 65.29 (0.6) & \cellcolor{white}& \cellcolor{white} 34.49 (1.4) & \cellcolor{white} 78.09 (1.2) & \cellcolor{white} 67.64 (0.6) & \cellcolor{white} 61.23 (0.7) & \cellcolor{white} 60.36 (1.0) \\
~ & \textit{M-EBO}   & \textbf{63.30} (0.7)  & \textbf{55.85} (0.5) & \textbf{59.58} (0.6) & & \textbf{16.37} (1.6) & \textbf{77.70} (0.7)  & \textbf{60.83} (0.7) & \textbf{51.48} (0.8) & \textbf{51.60} (0.9) \\ 
\bottomrule
\bottomrule
\end{tabular}
\end{sc}
\end{scriptsize}
\end{center}
\end{table*}

\linespread{1.2}
\begin{table*}[!t]
\setlength\tabcolsep{7pt} 
\caption{Comparison of OOD detection by AUROC (\%) ($\uparrow$) between \sysname+KNN, \sysname+KLM and baseline post-process methods (i.e., KNN and KLM) for CIFAR-10 using RigL (80\%, 90\%, 95\%, 99\% sparsity). Our \sysname leads to larger AUROC on each OOD data, showing its ability to improve reliability on OOD data in sparse training.}
\label{tb:ood-new-b}
\begin{center}
\begin{scriptsize}
\begin{tabular}{ccccccccccc}
\toprule \toprule
 &  &  CIFAR-100 & TIN   & NearOOD & & MNIST & SVHN  & Texture & Places365 & FarOOD \\
 \cline{1-2} \cline{3-5}   \cline{7-11}
 \multirow{4}{*}{80\%} & KNN        &   89.50  & 91.12  & 90.31  & & 94.11  & 93.76 & 92.69 & 90.83  & 92.85 \\
~ & \cellcolor{mygray} \sysname+KNN &  \cellcolor{mygray}  \textbf{90.16}  & \cellcolor{mygray} \textbf{92.11} & \cellcolor{mygray} \textbf{91.14} & \cellcolor{mygray}  & \cellcolor{mygray} \textbf{95.58}    & \cellcolor{mygray} \textbf{93.77}  & \cellcolor{mygray} \textbf{94.02}   & \cellcolor{mygray} \textbf{92.47}  & \cellcolor{mygray} \textbf{93.96} \\
~ & KLM        &   77.29  & 78.98  & 78.14  & & 83.65  & 86.22 & 80.29 & 77.89  & 82.01 \\
~ & \cellcolor{mygray} \sysname+KLM &   \cellcolor{mygray} \textbf{78.92}  & \cellcolor{mygray} \textbf{81.92} & \cellcolor{mygray} \textbf{80.42} & \cellcolor{mygray}  & \cellcolor{mygray} \textbf{89.29}    & \cellcolor{mygray} \textbf{86.44}  & \cellcolor{mygray} \textbf{84.78}   & \cellcolor{mygray} \textbf{80.23}  & \cellcolor{mygray} \textbf{85.18} \\
\cline{2-5}   \cline{7-11}
 \multirow{4}{*}{90\%} & KNN        &   88.49  & 90.54  & 89.51 & & 92.93  & 92.12  & 92.06   & 89.98 & 91.77 \\
~ & \cellcolor{mygray} \sysname+KNN   & \cellcolor{mygray} \textbf{89.40}  & \cellcolor{mygray} \textbf{91.32}  & \cellcolor{mygray} \textbf{90.36} & \cellcolor{mygray} & \cellcolor{mygray} \textbf{95.04}  & \cellcolor{mygray} \textbf{93.97}  & \cellcolor{mygray} \textbf{92.99}   & \cellcolor{mygray} \textbf{91.73} & \cellcolor{mygray} \textbf{93.43}  \\ 
~ & KLM        &   77.60& 79.66& 78.63& & 87.9& 85.58& 80.93& 76.64& 82.76 \\
~ & \cellcolor{mygray} \sysname+KLM   & \cellcolor{mygray} \textbf{79.21}  & \cellcolor{mygray} \textbf{81.29}  & \cellcolor{mygray} \textbf{80.25}  & \cellcolor{mygray} & \cellcolor{mygray} \textbf{89.38}  & \cellcolor{mygray} \textbf{89.94}& \cellcolor{mygray} \textbf{83.33} & \cellcolor{mygray} \textbf{80.04} & \cellcolor{mygray} \textbf{85.67}  \\ 
\cline{2-5}   \cline{7-11}
 \multirow{4}{*}{95\%} & KNN          & 88.10 & 89.87 & 88.98 &  & 94.85 & 93.10 & 92.60 &  89.50 &  92.51 \\
~ & \cellcolor{mygray} \sysname+KNN   & \cellcolor{mygray} \textbf{88.36}  & \cellcolor{mygray} \textbf{90.34} & \cellcolor{mygray} \textbf{89.35}  & \cellcolor{mygray} & \cellcolor{mygray} \textbf{94.98}  & \cellcolor{mygray} \textbf{93.62}  & \cellcolor{mygray} \textbf{92.91}  & \cellcolor{mygray} \textbf{90.51}  & \cellcolor{mygray} \textbf{93.01}  \\
~ & KLM          & 76.56 & 78.34 & 77.45 &  & 87.84 & 78.86 & 77.67 & 76.15 &  80.13 \\
~ & \cellcolor{mygray} \sysname+KLM   & \cellcolor{mygray} \textbf{79.06}  & \cellcolor{mygray} \textbf{81.26} & \cellcolor{mygray} \textbf{80.16}  & \cellcolor{mygray} & \cellcolor{mygray} \textbf{89.64}  & \cellcolor{mygray} \textbf{84.04}  & \cellcolor{mygray} \textbf{83.59}  & \cellcolor{mygray} \textbf{79.35}  & \cellcolor{mygray} \textbf{84.16}  \\
\cline{2-5}   \cline{7-11}
 \multirow{2}{*}{99\%} & KNN        & 81.87 & 84.18 & 83.03 &  & 92.69 & 86.00 &  86.15 &  82.96 &  86.95 \\
~ & \cellcolor{mygray} \sysname+KNN &   \cellcolor{mygray} \textbf{82.97}  & \cellcolor{mygray} \textbf{84.78}  & \cellcolor{mygray} \textbf{83.88}  & \cellcolor{mygray} & \cellcolor{mygray} \textbf{96.62} & \cellcolor{mygray} \textbf{95.64} & \cellcolor{mygray} \textbf{89.15}  & \cellcolor{mygray} \textbf{83.03}  & \cellcolor{mygray} \textbf{91.11} \\
~ & KLM         & 72.27 &  74.47 &  73.37 &  &  78.44 &  65.96 &  74.22 &  73.64 &  73.06 \\
~ & \cellcolor{mygray} \sysname+KLM &   \cellcolor{mygray} \textbf{75.40}  & \cellcolor{mygray} \textbf{76.63}  & \cellcolor{mygray} \textbf{76.01}  & \cellcolor{mygray} & \cellcolor{mygray} \textbf{85.03} & \cellcolor{mygray} \textbf{82.32} & \cellcolor{mygray} \textbf{78.51}  & \cellcolor{mygray} \textbf{74.90}  & \cellcolor{mygray} \textbf{80.19} \\ \bottomrule
\bottomrule
\end{tabular}
\end{scriptsize}
\end{center}
\end{table*}

\subsection{Comparison with Calibration Method}

\begin{figure*}[!t]
\vspace{-0.0cm}
\begin{center}
\subfigure[Temperature Scaling (90\%)]{\includegraphics[scale=0.42]{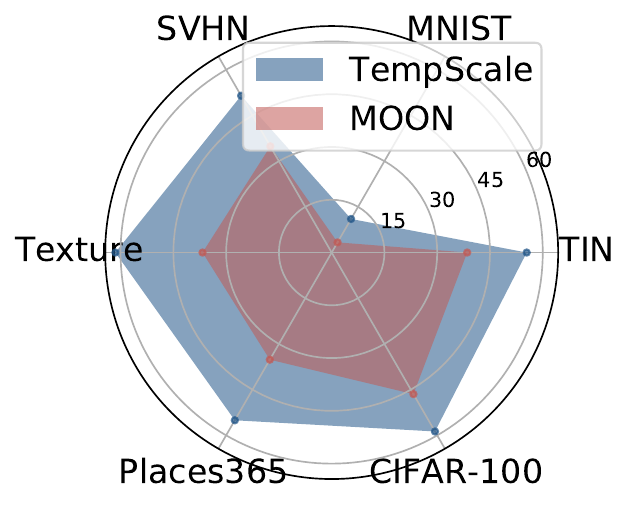}}
\subfigure[Mixup (90\%)]{\includegraphics[scale=0.42]{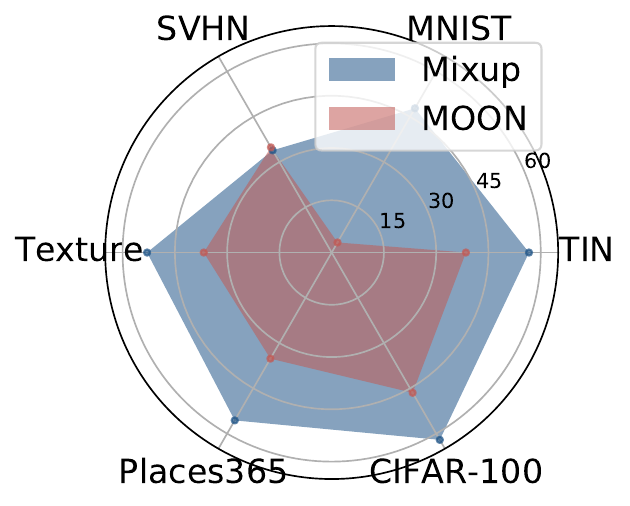}}
\subfigure[CigL (90\%)]{\includegraphics[scale=0.42]{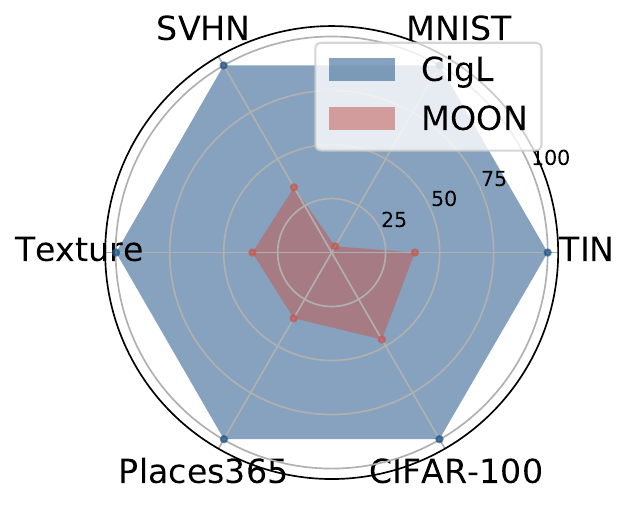}}
\caption{Comparison of OOD detection by FPR-95 (\%) ($\downarrow$) on CIFAR-10 between \sysname and other calibration methods using RigL (90\%). The red hexagons (\sysname) are smaller than the blue hexagons (other calibration methods), indicating a better OOD detection using \sysname compared to (a) Temperature Scaling, (b) Mixup, and (c) CigL.}
\label{fig: ood-calib-sp}
\end{center}
\end{figure*}

Prior work shows that calibration methods can help with OOD detection in dense DNNs~\citep{wang2021energy, yang2022openood}. To show the ineffectiveness of existing calibration methods for OOD detection in sparse training, we also compare our \sysname to calibration methods including temperature scaling~\citep{lee2018simple}, Mixup~\citep{zhang2017mixup}, and CigL~\citep{lei2023calibrating} using RigL (90\%) for OOD detection capability. For our \sysname, we report the results of ``\sysname + ODIN".  As shown in Figure~\ref{fig: ood-calib-sp}, the red and blue hexagons represent the FPR-95 of \sysname and calibration methods, respectively. We can see that the red hexagons are smaller than the blue hexagons, indicating better OOD detection from \sysname compared to calibration methods.

\subsection{ID Test Accuracy and Reliability in Sparse Training}

In this section, we show that, in addition to improving OOD detection, our \sysname can also maintain or improve the accuracy and reliability on ID data. We incorporate our \sysname into sparse training (RigL) and compare it with the original sparse training results without \texttt{MOON}. 

\begin{figure*}[!t]
\setlength{\abovecaptionskip}{0.0cm}
\vspace{-0.2cm}
\begin{center}
\subfigure[Test ACC (CIFAR-10)]{\includegraphics[width=0.245\textwidth,height=0.22\textwidth]{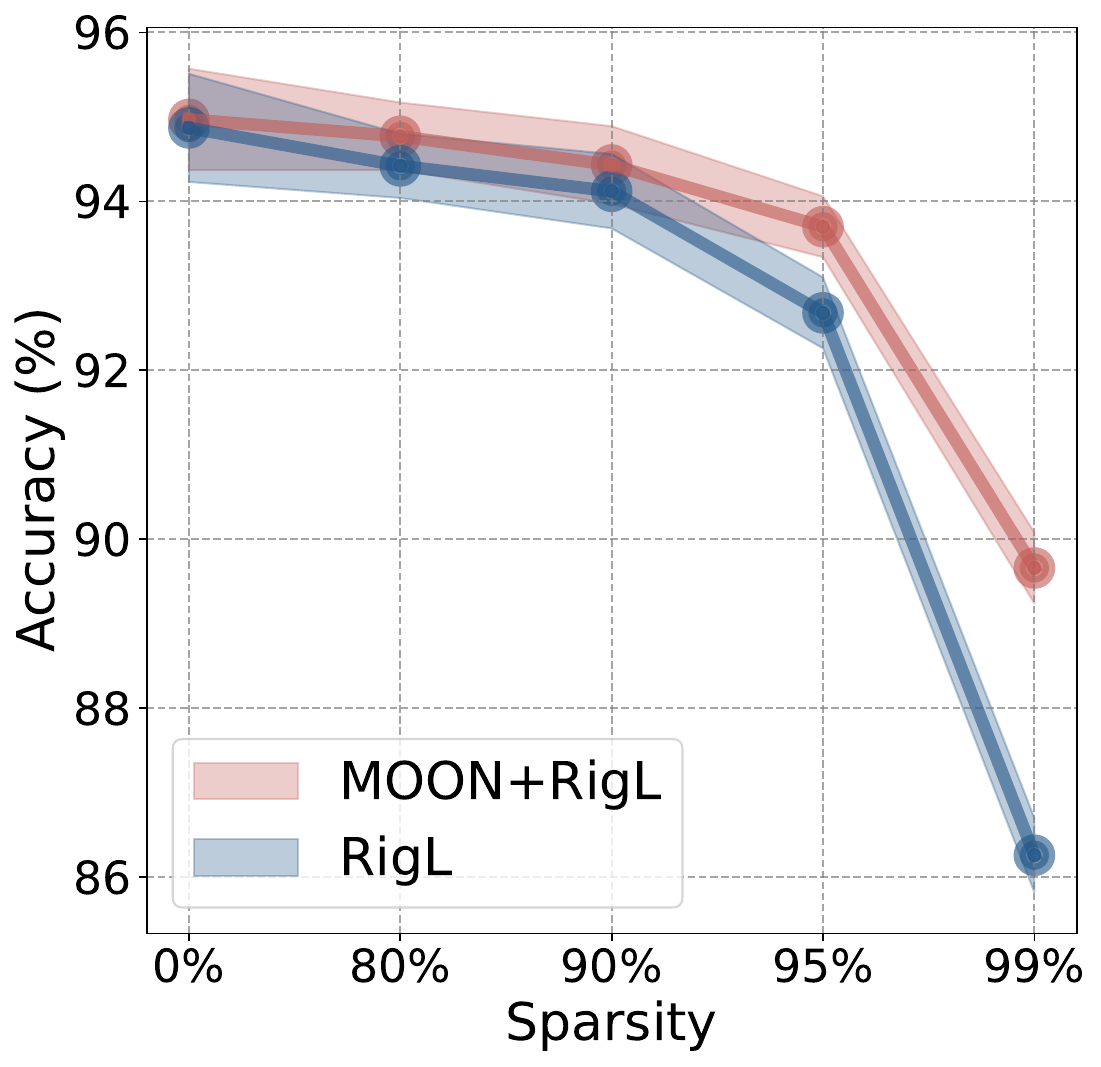}}
\subfigure[Test ACC (CIFAR-100)]{\includegraphics[width=0.245\textwidth,height=0.22\textwidth]{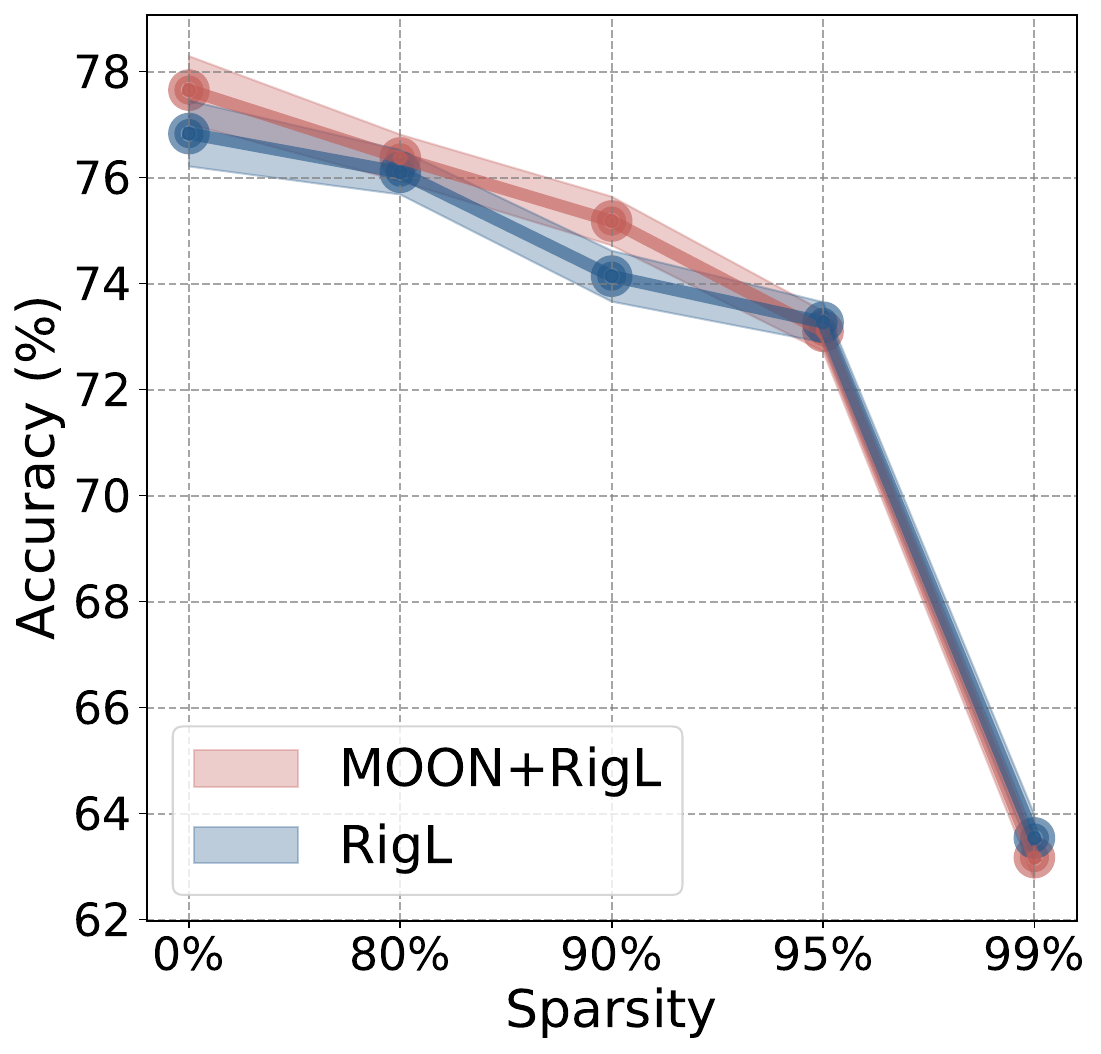}}
\subfigure[ECE (CIFAR-10)]{\includegraphics[width=0.245\textwidth,height=0.22\textwidth]{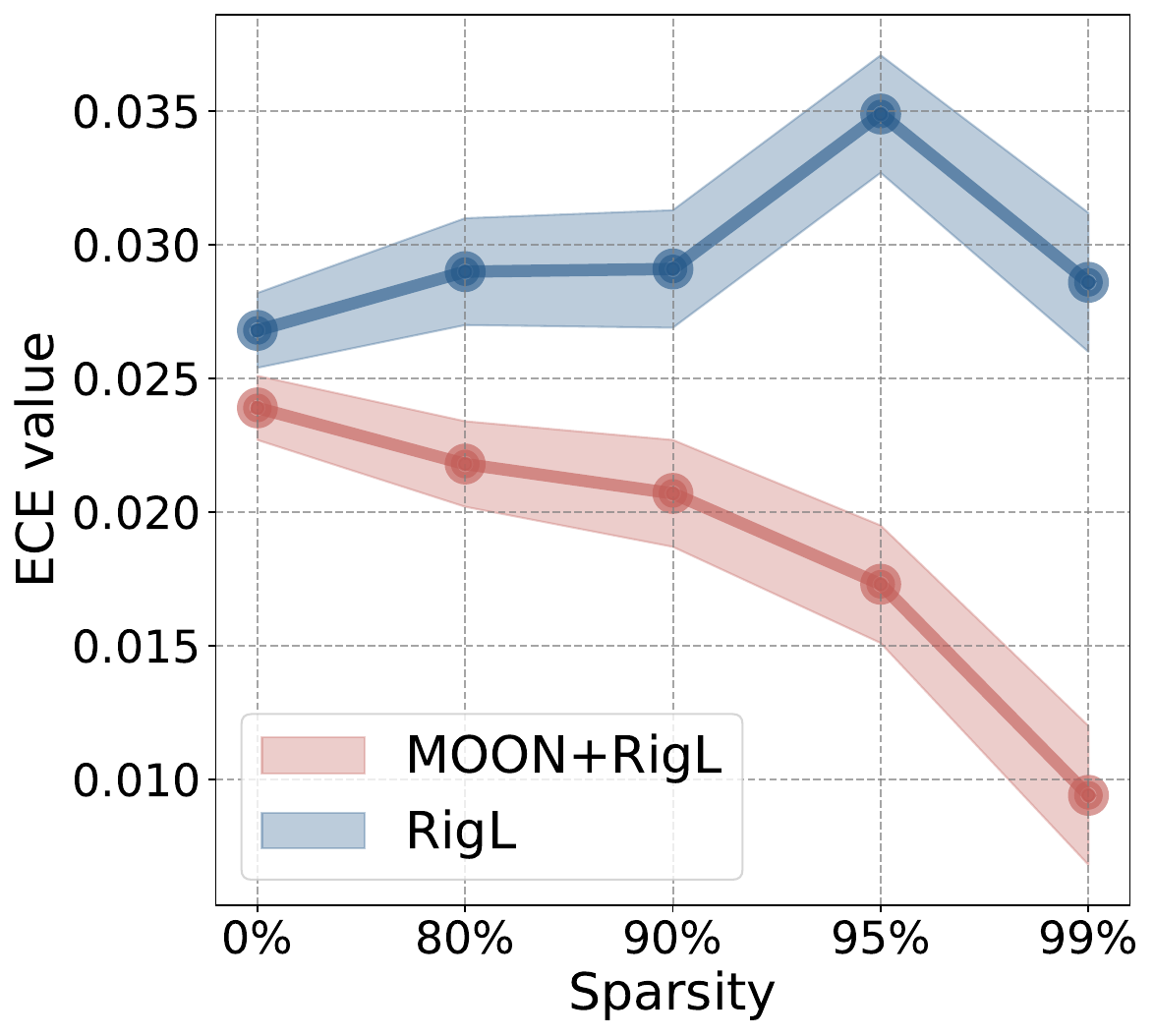}}
\subfigure[ECE (CIFAR-100)]{\includegraphics[width=0.245\textwidth,height=0.22\textwidth]{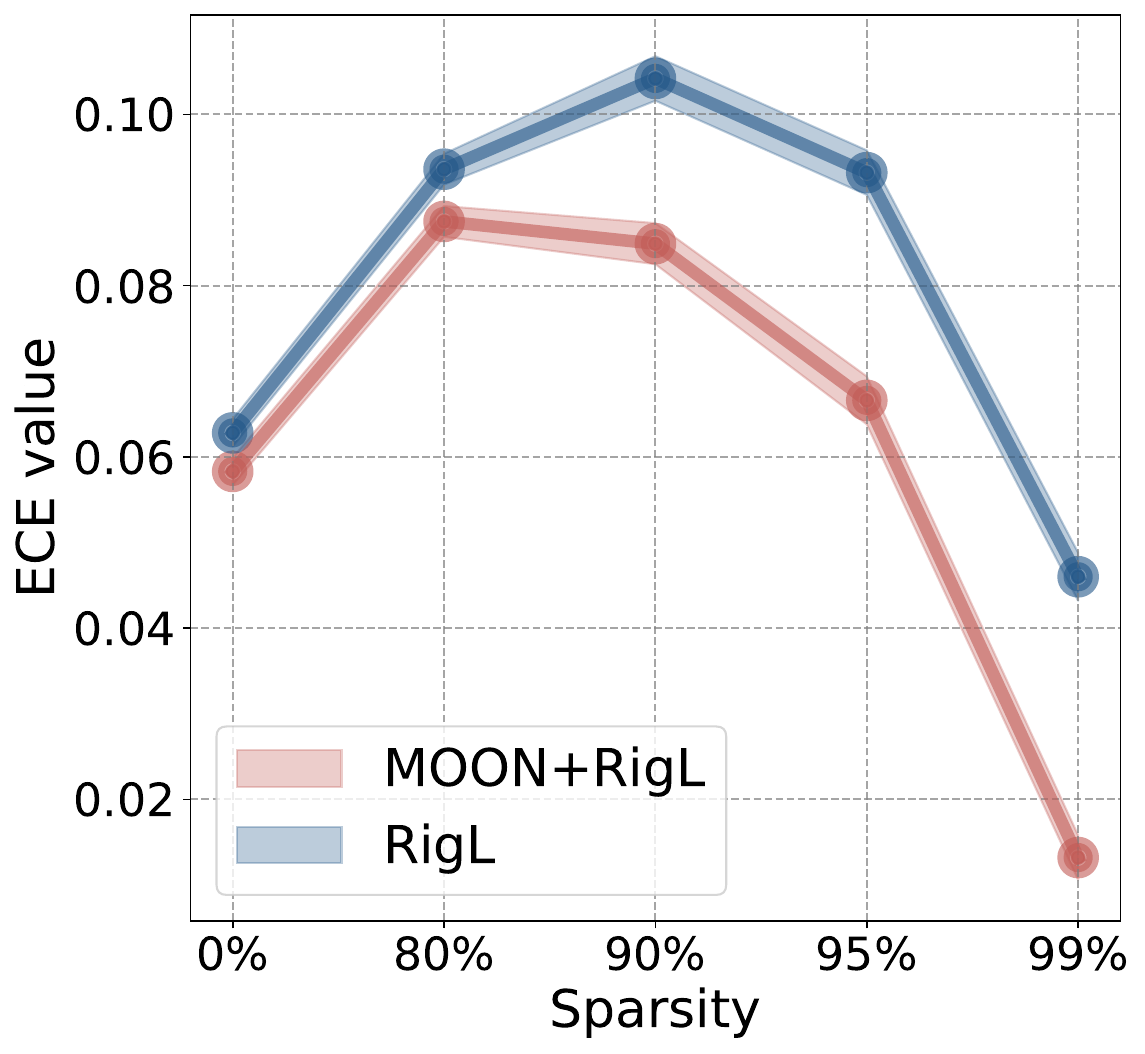}}
\caption{Comparison of performance on ID data by ECE ($\downarrow$) and test accuracy (ACC) (\%) ($\uparrow$) between \sysname and RigL. Our \sysname leads to smaller ECE and maintains comparable or higher test accuracy on ID data, showing its ability to improve reliability on ID data.}
\label{fig: id-comp}
\end{center}
\vspace{-0.4cm}
\end{figure*}

\textbf{For ID Test Accuracy}, as shown in Figure~\ref{fig: id-comp} (a)-(b), the red and blue curves represent our \sysname+RigL and RigL, respectively. The red curve is usually higher or equal to the blue curve, implying that our \sysname can maintain comparable or higher test accuracy on the ID data.

\textbf{For ID Reliability}, we choose expected calibration error (ECE) \citep{guo2017calibration}, a widely-used measure of the discrepancy between a model's confidence and true accuracy, where a lower ECE indicates better confidence calibration and higher reliability. 
As shown in Figure~\ref{fig: id-comp} (c)-(d), the red and blue curves represent our \texttt{MOON}+RigL and RigL, respectively, where the colored ares represent the 95\% confidence intervals. The red curve is usually lower than the blue curve. This shows that our \sysname reduces the ECE, implying an improvement in the ID reliability of our \texttt{MOON}.

\linespread{1.2}
\begin{table*}[ht]
\setlength\tabcolsep{3.6pt} 
\setlength{\belowcaptionskip}{-0.0cm}
\caption{Comparison of reliability of ID and OOD data between \sysname+MSP (\textit{M-MSP}) and MSP for CIFAR-10 in sparse training using SET. Our \sysname leads to smaller ECE and comparable or higher accuracy (ACC) (\%) on ID data, and smaller FPR-95 (\%) on each OOD data compared to baseline method MSP, showing its ability to improve reliability on ID and OOD data.}
\label{tb: id-ood-sparse-set}
\begin{center}
\begin{scriptsize}
\begin{sc}
\begin{tabular}{cccccccccccc}
\toprule \toprule
 &  & & \multicolumn{2}{c}{ID Data} & & \multicolumn{6}{c}{OOD Data (FPR-95 (\%) ($\downarrow$))}\\ 
   \cline{4-5}  \cline{7-12}
  &  & &ECE($\downarrow$)& ACC ($\uparrow$) & & CIFAR-100 & TIN   & MNIST & SVHN  & Texture & Places365  \\
 \cline{1-2} \cline{4-5}   \cline{7-12}
 \multirow{2}{*}{80\%} & MSP       &  & 0.031 (0.002)& 94.03 (0.2)&  &   64.01 (0.5) & 60.64 (0.4) & \textbf{46.62} (1.2) & 66.43 (1.4) & 64.50 (0.7)  & 60.63 (0.4)  \\
~ & \cellcolor{mygray} \textit{M-MSP}       & \cellcolor{mygray}  &\cellcolor{mygray} \textbf{0.020} (0.001)& \cellcolor{mygray} \textbf{94.80} (0.2)&  \cellcolor{mygray} &   \cellcolor{mygray} \textbf{62.72} (0.4) & \cellcolor{mygray} \textbf{58.63} (0.3) & \cellcolor{mygray} 47.63 (1.5) & \cellcolor{mygray} \textbf{60.68} (1.2)  & \cellcolor{mygray} \textbf{59.10} (0.8)  & \cellcolor{mygray} \textbf{59.39} (0.3) \\
 \multirow{2}{*}{90\%} & MSP     &  & 0.032 (0.001)& 93.94 (0.2)&  &   64.70 (0.3)  & 62.52 (0.4) & 50.45 (1.6) & 67.50 (1.7)  & 64.13 (0.6) & 63.33 (0.4) \\
~ & \cellcolor{mygray} \textit{M-MSP}        &  \cellcolor{mygray} & \cellcolor{mygray} \textbf{0.019} (0.001)& \cellcolor{mygray} \textbf{94.44} (0.3)&  \cellcolor{mygray} & \cellcolor{mygray} \textbf{63.33} (0.3) & \cellcolor{mygray} \textbf{60.15}  (0.5)& \cellcolor{mygray} \textbf{43.02} (1.2) & \cellcolor{mygray} \textbf{58.81} (2.2) & \cellcolor{mygray} \textbf{61.26} (0.6) & \cellcolor{mygray} \textbf{60.79} (0.4) \\
 \multirow{2}{*}{95\%} &MSP          &  &0.035 (0.002)& 92.56 (0.3)&  & 67.62 (0.5) & 65.18 (0.4) & 51.79 (1.7) & 67.95 (1.5) & 67.46 (0.7) & 65.44 (0.5) \\
~ & \cellcolor{mygray}  \textit{M-MSP}  &  \cellcolor{mygray} & \cellcolor{mygray} \textbf{0.020} (0.003)& \cellcolor{mygray} \textbf{93.59} (0.3)& \cellcolor{mygray}  & \cellcolor{mygray} \textbf{63.69} (0.5) & \cellcolor{mygray} \textbf{60.20} (0.5) & \cellcolor{mygray} \textbf{41.82} (1.9) & \cellcolor{mygray} \textbf{62.02} (1.1) & \cellcolor{mygray} \textbf{57.29} (0.8) & \cellcolor{mygray} \textbf{60.85} (0.4) \\
 \multirow{2}{*}{99\%} &MSP   &  &0.025 (0.001)& 87.58 (0.2)&  & 74.98 (0.4)& 73.27 (0.5) & 54.12 (1.5) & 74.44 (1.2) & 74.36 (0.7) & 74.03 (0.3) \\
~ & \cellcolor{mygray} \textit{M-MSP}  &  \cellcolor{mygray} & \cellcolor{mygray} \textbf{0.008} (0.002)& \cellcolor{mygray} \textbf{88.67} (0.2)&  \cellcolor{mygray} &   \cellcolor{mygray} \textbf{71.88} (0.5) & \cellcolor{mygray} \textbf{69.59} (0.5) & \cellcolor{mygray} \textbf{45.14} (1.4) & \cellcolor{mygray} \textbf{71.10} (1.1)  & \cellcolor{mygray} \textbf{65.73} (0.9) & \cellcolor{mygray} \textbf{71.61} (0.3) \\ \hline
\bottomrule
\end{tabular}
\end{sc}
\end{scriptsize}
\end{center}
\end{table*}

\subsection{Comparison with Other Sparse Training Method}

Apart from the comparison based on RigL, we also include sparse training results using SET~\citep{mocanu2018scalable} on CIFAR-10 at 80\%, 90\%, 95\%, 99\% sparsities. As shown in Table~\ref{tb: id-ood-sparse-set}, on OOD data, our \sysname also leads to smaller FPR-95, indicating improved OOD reliability from our \texttt{MOON}. On ID data, our \sysname brings smaller ECE and comparable or higher accuracy (ACC), implying comparable or improved ID reliability and performance. This shows our \sysname is effective in different sparse training methods.

\begin{figure*}[!t]
\setlength{\abovecaptionskip}{-0.0cm}
\begin{center}
\subfigure[\sysname w/o LM (80\%)]{\includegraphics[scale=0.42]{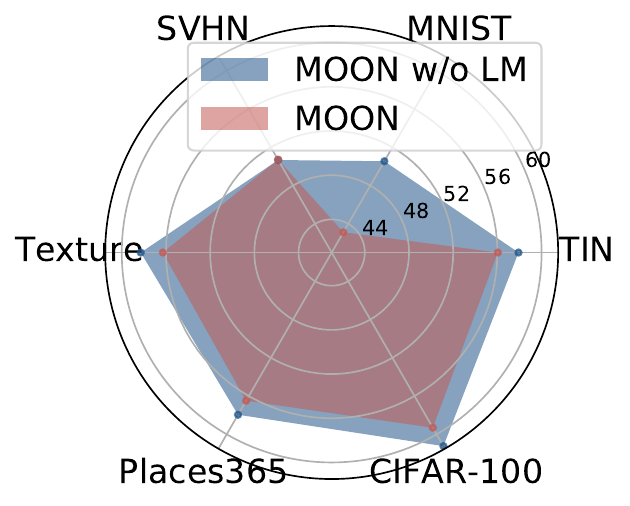}}
\subfigure[\sysname w/o AT (80\%)]{\includegraphics[scale=0.42]{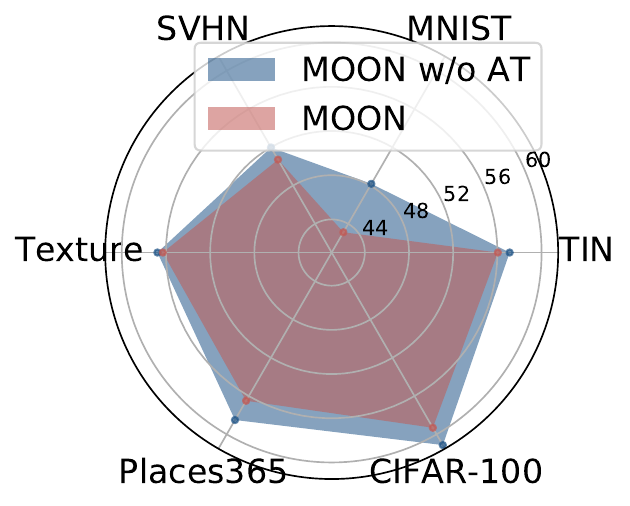}}
\subfigure[\sysname w/o VT (80\%)]{\includegraphics[scale=0.42]{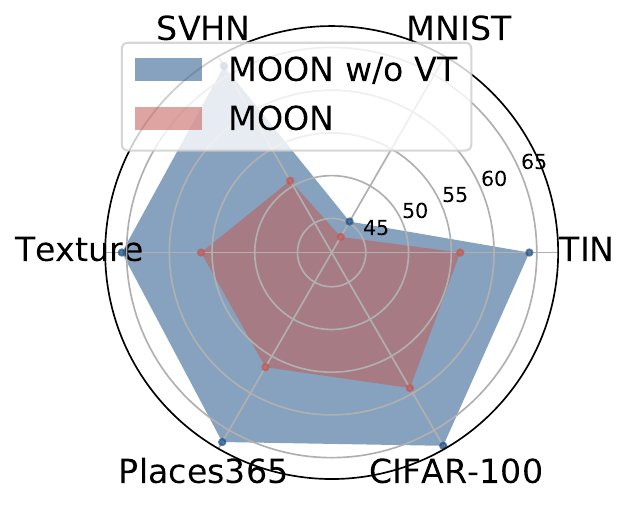}}
\caption{Ablation studies: comparison of FPR-95 (\%) ($\downarrow$) between \sysname, \sysname w/o LM, \sysname w/o AT, and \sysname w/o VT on OOD data of CIFAR-10. \sysname produces lower FPR-95 compared to the other three methods.}
\label{fig: abla-ood}
\end{center}
\end{figure*}

\subsection{Comparison with Other Training-time Regularization Method}

Since our loss modification can be viewed as a new type of training-time regularization, we also add additional experiments on other dense training-time regularization methods including LogitNorm~\citep{wei2022mitigating} and VOS~\citep{du2022vos}. As shown in Table~\ref{tb:ood-tr-reg}, our method MOON has larger AUROC compared to LogitNorm and VOS, indicating better OOD reliability from our method in sparse training.

\linespread{1.2}
\begin{table*}[!t]
\vspace{-0.2cm}
\setlength{\belowcaptionskip}{-0.1cm}
\setlength\tabcolsep{7pt} 
\caption{Comparison of OOD detection by AUROC (\%) ($\uparrow$) between \sysname+MSP (\textit{M-MSP}), LogitNorm+MSP (L-MSP), and VOS+MSP (V-MSP) for CIFAR-10 in sparse training. Our \sysname leads to larger AUROC on each OOD data of CIFAR-10, showing its ability to improve OOD reliability.}
\label{tb:ood-tr-reg}
\begin{center}
\begin{scriptsize}
\begin{sc}
\begin{tabular}{ccccccccccc}
\toprule \toprule
 &  &  CIFAR-100 & TIN   & NearOOD & & MNIST & SVHN  & Texture & Places365 & FarOOD \\
 \cline{1-2} \cline{3-5}   \cline{7-11}
 \multirow{3}{*}{80\%} & L-MSP        &   87.71  & 89.60  & 88.66  & & 94.74  & 93.29 & 89.36 & 90.79  & 92.05 \\
~ & V-MSP        &   87.57  & 89.04  & 88.31  & & 91.82  & 90.93  & 89.42  & 88.60  & 90.19  \\
~ & \cellcolor{mygray} \textit{M-MSP} &   \cellcolor{mygray} \textbf{89.93}  & \cellcolor{mygray} \textbf{91.14} & \cellcolor{mygray} \textbf{90.53} & \cellcolor{mygray} & \cellcolor{mygray} \textbf{95.00}    & \cellcolor{mygray} \textbf{93.82}  & \cellcolor{mygray} \textbf{91.80}   & \cellcolor{mygray} \textbf{90.82}  & \cellcolor{mygray} \textbf{92.86} \\
\cline{2-5}   \cline{7-11}
 \multirow{3}{*}{90\%} & L-MSP        &   87.81  & 89.41  & 88.61 & & 95.37  & 91.26  & 90.24   & 89.85 & 91.68 \\
 ~ & V-MSP        &   87.48  & 88.58  & 88.03 & & 92.97  & 92.17 & 88.93  & 88.81  & 90.53 \\
~ & \cellcolor{mygray} \textit{M-MSP}   & \cellcolor{mygray} \textbf{88.92}  & \cellcolor{mygray} \textbf{90.34}  & \cellcolor{mygray} \textbf{89.63}  & \cellcolor{mygray} & \cellcolor{mygray} \textbf{95.39}  & \cellcolor{mygray} \textbf{93.09}& \cellcolor{mygray} \textbf{91.09} & \cellcolor{mygray} \textbf{89.92} & \cellcolor{mygray} \textbf{92.37}  \\ 
\cline{2-5}   \cline{7-11}
 \multirow{3}{*}{95\%} &L-MSP          & 87.35  & 88.58  & 87.96  & & 93.36  & 86.36 & 87.79  & 88.49  & 89.03 \\
 ~ &V-MSP          &  87.12  & 87.98  & 87.55  & & 93.64  & 86.40 & 88.10 & 87.89  & 89.01 \\
~ & \cellcolor{mygray} \textit{M-MSP}   & \cellcolor{mygray} \textbf{88.04}  & \cellcolor{mygray} \textbf{89.39} & \cellcolor{mygray} \textbf{88.71}  & \cellcolor{mygray} & \cellcolor{mygray} \textbf{94.76}  & \cellcolor{mygray} \textbf{93.35}  & \cellcolor{mygray} \textbf{89.86}  & \cellcolor{mygray} \textbf{88.51}  & \cellcolor{mygray} \textbf{91.62}  \\
\cline{2-5}   \cline{7-11}
 \multirow{2}{*}{99\%} &L-MSP         & 82.60  & 84.52 & 83.56  & & 91.75  & 87.56  & 84.16  & 82.68  & 86.54 \\
 ~ &V-MSP        & 82.22  & 83.09 & 82.65  & & 84.75  & 89.99  & 82.75  & 82.26  & 84.69 \\
~ & \cellcolor{mygray} \textit{M-MSP} &   \cellcolor{mygray} \textbf{83.75}  & \cellcolor{mygray} \textbf{85.11}  & \cellcolor{mygray} \textbf{84.43}  & \cellcolor{mygray} & \cellcolor{mygray} \textbf{91.96} & \cellcolor{mygray} \textbf{89.01} & \cellcolor{mygray} \textbf{86.44}  & \cellcolor{mygray} \textbf{83.38}  & \cellcolor{mygray} \textbf{87.70} \\ \bottomrule
\bottomrule
\end{tabular}
\end{sc}
\end{scriptsize}
\end{center}
\vspace{-0.4cm}
\end{table*}

\subsection{Comparison with Other Extra-dimension Method}

Since our loss modification also belongs to the extra-dimension method, we further compare our \sysname with another extra-dimension method, i.e., DAC~\citep{thulasidasan2019combating} regarding the reliability of ID and OOD data. As shown in Table~\ref{tb: ood-dac}, our \sysname also provides larger AUROC when given OOD data. When faced with ID data, our \sysname leads to smaller ECE and comparable or higher accuracy (ACC). This suggests that our \sysname makes better use of the additional dimension to encourage the model to think about the unknown, thus improving real-world reliability while maintaining ID performance and reliability.

\linespread{1.2}
\setlength\tabcolsep{5.5pt}
\begin{table*}[!t]
\caption{Comparison of reliability of ID and OOD data between \texttt{MOON} and DAC for CIFAR-10 and CIFAR-100. Our \texttt{MOON} leads to smaller ECE and comparable or higher accuracy (ACC) on ID data, and larger AUROC on each OOD data, showing its ability to improve reliability on ID and OOD data.}
\label{tb: ood-dac}
\begin{center}
\begin{scriptsize}
\begin{sc}
\begin{tabular}{cccccccccccc}
\toprule \toprule
 &  & &\multicolumn{2}{c}{ID Data} & & \multicolumn{6}{c}{OOD Data (AUROC ($\uparrow$))}\\ 
  \cline{1-2} \cline{4-5}  \cline{7-12}
 Data & Method & &ECE($\downarrow$)& ACC($\uparrow$) & & CIFAR-100/10 & TIN   & MNIST & SVHN  & Texture & Places365  \\
 \cline{1-2} \cline{4-5}  \cline{7-12}
 \multirow{2}{*}{CIFAR-10} & DAC  &      & 0.0302&  94.43&  & 86.17 & 87.96 & 89.92 & 87.26 & 89.32 & 88.48  \\
~ & \cellcolor{mygray} \texttt{MOON} & \cellcolor{mygray} & \cellcolor{mygray} \textbf{0.0239}& \cellcolor{mygray} \textbf{94.97}&  \cellcolor{mygray} & \cellcolor{mygray} \textbf{89.93} & \cellcolor{mygray} \textbf{91.14} & \cellcolor{mygray} \textbf{95.00}    & \cellcolor{mygray} \textbf{93.82} & \cellcolor{mygray} \textbf{91.80}  & \cellcolor{mygray} \textbf{90.82}  \\
 \multirow{2}{*}{CIFAR-100} & DAC  & & 0.0873& 76.03&     &   76.99 & 80.54 & 74.43 & 77.70 & \textbf{77.89}  & 77.86\\
~ & \cellcolor{mygray} \texttt{MOON} & \cellcolor{mygray} & \cellcolor{mygray} \textbf{0.0583}& \cellcolor{mygray} \textbf{77.65}&   \cellcolor{mygray} & \cellcolor{mygray} \textbf{78.82}&  \cellcolor{mygray} \textbf{82.38} &  \cellcolor{mygray} \textbf{80.28}&  \cellcolor{mygray} \textbf{81.57}&  \cellcolor{mygray} 77.62&  \cellcolor{mygray} \textbf{79.65}\\  \hline
\bottomrule
\end{tabular}
\end{sc}
\end{scriptsize}
\end{center}
\end{table*}

\subsection{Ablation Studies}\label{subsec:abla}

We do ablation studies to demonstrate the importance of each component in our \sysname, where we train CIFAR-100 using our \sysname without the unknown-aware loss modification (\sysname w/o LM), \sysname without the auto-tuning strategy (\sysname w/o AT), and \sysname without the average-based voting scheme (\sysname w/o VT), respectively. The OOD detection is shown by FPR-95 for each OOD data of CIFAR-100 in Figure~\ref{fig: abla-ood}, and the ID results are summarized in Table~\ref{tb: abla-id}.

\textbf{\sysname w/o LM}: As shown in Figure~\ref{fig: abla-ood} (a), the blue hexagon (\sysname w/o LM) is larger than the red hexagon (\texttt{MOON}), indicating a decrease in OOD reliability without the unknown-aware loss modification. Table~\ref{tb: abla-id} also shows an increased ECE and decreased ID reliability in \sysname w/o LM.

\textbf{\sysname w/o AT}: Similarly, as depicted in Figure~\ref{fig: abla-ood} (b), the blue hexagon (\sysname w/o AT) is larger than the red hexagon (\texttt{MOON}), indicating a decrease in OOD reliability in the absence of the auto-tuning strategy. Table~\ref{tb: abla-id} also shows an increased ECE and decreased ID reliability in \sysname w/o AT.

\textbf{\sysname w/o VT}: As summarized in Figure~\ref{fig: abla-ood} (c), the blue hexagon (\sysname w/o VT) is larger than the red hexagon (\texttt{MOON}), indicating a decrease in OOD reliability when no average-based voting scheme is present. Table~\ref{tb: abla-id} further shows a decreased accuracy and increased ECE in \sysname w/o VT.

\linespread{1.2}
\begin{table*}[!t]
\setlength\tabcolsep{3.5pt} 
\caption{Ablation studies: comparison of ECE ($\downarrow$) and accuracy (ACC, $\uparrow$) between \sysname, \sysname w/o LM, \sysname w/o AT, and \sysname w/o VT on CIFAR-100. \sysname produces lower ECE values and comparable or higher accuracy.}
\label{tb: abla-id}
\begin{center}
\begin{scriptsize}
\begin{sc}
\begin{tabular}{ccccccccccccc}
\toprule \toprule
\multirow{2}{*}{Method} & & \multicolumn{2}{c}{80\%} &  & \multicolumn{2}{c}{90\%} &   & \multicolumn{2}{c}{95\%}  & &  \multicolumn{2}{c}{99\%}  \\
\cline{3-4} \cline{6-7} \cline{9-10} \cline{12-13}
~ & &ECE ($\downarrow$)& ACC (\%) ($\uparrow$)  & &ECE ($\downarrow$)& ACC (\%) ($\uparrow$) & &ECE ($\downarrow$)& ACC (\%) ($\uparrow$) & &ECE ($\downarrow$)& ACC (\%) ($\uparrow$)  \\
\cline{1-1} \cline{3-4} \cline{6-7} \cline{9-10} \cline{12-13}
\sysname w/o LM &      & 0.0272& 94.68      & & 0.0282& 94.40      & & 0.0307& 93.44      & & 0.0207& 89.63\\
\sysname w/o AT & & 0.0250& \textbf{94.81}     &   & 0.0258& 94.51    &   & 0.0308& 93.26    &   & 0.0200& 89.50  \\
\sysname w/o VT &  & 0.0296& 92.97     &   & 0.0256& 91.81     &  & 0.0330& 90.23    &   & 0.0096& 85.38\\
\rowcolor{mygray} \sysname & & \textbf{0.0218}& 94.77    &   & \textbf{0.0207}& \textbf{94.53}     &  & \textbf{0.0173}& \textbf{93.70}    &   & \textbf{0.0094}& \textbf{89.66}\\  \hline
\bottomrule
\end{tabular}
\end{sc}
\end{scriptsize}
\end{center}
\end{table*}

%% file: sections/_s6_conclusion.tex
\section{Conclusion and Limitations}

We investigate for the first time the reliability of sparse training from an OOD perspective, which jointly considers OOD reliability and efficiency and has important implications for real-world DNN applications. We observe that sparse training causes more severe unreliability, with useful unknown information being ignored in the training. Thus, we propose a new unknown-aware sparse training method, \texttt{MOON}, to enhance OOD detection under sparsity constraints, with little additional cost and no extra OOD data requirement.

We provide theoretical insights to demonstrate the lower confidence brought by \sysname on OOD samples. We also conduct extensive experiments on multiple benchmark datasets, architectures, and sparsities, showing that our \sysname can improve AUROC by up to 8.4\%. Our \sysname can be used as a drop-in replacement for existing sparse training methods, which fills an important gap in sparse training and improves the effectiveness and applicability of the sparse training paradigm.

Despite the enhanced OOD reliability achieved in sparse training, \sysname does not address this matter comprehensively from a broader reliability perspective, including robust generalization~\citep{silva2020opportunities, ozdenizci2021training, chen2022sparsity} and adaptation~\citep{emam2021active, alayrac2022flamingo, hu2022pushing}, which are equally vital aspects of reliability and deserve further investigation~\citep{tran2022plex}. Additionally, the effectiveness of \sysname remains unexplored in the context of widely employed large generative models~\citep{ouyang2022training, xu2023rewoo, croitoru2023diffusion}. On the one hand, as model size expands, novel changes in model characteristics may arise~\citep{wei2022emergent}. On the other hand, given the prevalence of non-classification tasks within LLMs, the exploration of how to evaluate their reliability remains an uncharted domain.

%% file: sections/_a_Appendix.tex
\section{Appendix: Additional Experimental Results}

\subsection{More Sparse Training Results}\label{subsec:sp-more}

To further demonstrate the effectiveness of our \sysname, apart from the comparison of AUROC ($\uparrow$) on CIFAR-10 in the main manuscript, we also include more comparison of FPR-95~($\downarrow$) between \texttt{MOON}+MSP and MSP for CIFAR-10 in sparse training using RigL. As shown in Table~\ref{tb: ood-sparse-appe}, our \sysname usually leads to smaller FPR-95, indicating more effective OOD detection from our \sysname compared to baseline methods.

\linespread{1.2}
\begin{table*}[ht]
\setlength\tabcolsep{5.5pt} 
\caption{Comparison of OOD detection by FPR-95 ($\downarrow$) between \texttt{MOON}+MSP and MSP for CIFAR-10 in sparse training using RigL. Our \texttt{MOON} leads to larger AUROC on each OOD data of CIFAR-10, showing its ability to improve reliability on OOD data.}
\label{tb: ood-sparse-appe}
\begin{center}
\begin{scriptsize}
\begin{sc}
\begin{tabular}{ccccccccccc}
\toprule \toprule
 &  &  CIFAR-100 & TIN   & NearOOD & & MNIST & SVHN  & Texture & Places365 & FarOOD \\
 \cline{1-2} \cline{3-5}   \cline{7-11}
 \multirow{2}{*}{80\%} & MSP        &   62.97 & 59.71 & 61.34 &  &48.16 & 53.70  & 60.35 & 58.59 & 55.20 \\
~ & \cellcolor{mygray} \texttt{MOON}+MSP &   \cellcolor{mygray} \textbf{59.31} & \cellcolor{mygray} \textbf{56.04} & \cellcolor{mygray} \textbf{57.68} &  \cellcolor{mygray} &\cellcolor{mygray} \textbf{39.12} & \cellcolor{mygray} \textbf{50.71} & \cellcolor{mygray} \textbf{56.28} & \cellcolor{mygray} \textbf{56.49} & \cellcolor{mygray} \textbf{50.65} \\
 \multirow{2}{*}{90\%} & MSP        &   63.71 & 61.54 & 62.62 &  &54.85 & 58.33 & 66.76 & 61.43 & 60.34 \\
~ & \cellcolor{mygray} \texttt{MOON}+MSP   & \cellcolor{mygray} \textbf{62.70}  & \cellcolor{mygray} \textbf{58.18} & \cellcolor{mygray} \textbf{60.44} &  \cellcolor{mygray} &\cellcolor{mygray} \textbf{33.67} & \cellcolor{mygray} \textbf{54.65} & \cellcolor{mygray} \textbf{57.32} & \cellcolor{mygray} \textbf{58.41} & \cellcolor{mygray} \textbf{51.01} \\
 \multirow{2}{*}{95\%} &MSP          & 70.38 & 66.96 & 68.67 &  &44.19 & 74.57 & 73.01 & 65.37 & 64.28 \\
~ & \cellcolor{mygray} \texttt{MOON}+MSP   & \cellcolor{mygray} \textbf{65.29} & \cellcolor{mygray} \textbf{61.11} & \cellcolor{mygray} \textbf{63.20}  &  \cellcolor{mygray} &\cellcolor{mygray} \textbf{39.26} & \cellcolor{mygray} \textbf{52.72} & \cellcolor{mygray} \textbf{63.44} & \cellcolor{mygray} \textbf{63.38} & \cellcolor{mygray} \textbf{54.70} \\
 \multirow{2}{*}{99\%} &MSP         & 77.34 & 73.71 & 75.53 &  &60.45 & 77.77 & 71.54 & 76.49 & 71.56 \\
~ & \cellcolor{mygray} \texttt{MOON}+MSP &   \cellcolor{mygray} \textbf{70.38} & \cellcolor{mygray} \textbf{67.05} & \cellcolor{mygray} \textbf{68.72} &  \cellcolor{mygray} &\cellcolor{mygray} \textbf{44.34} & \cellcolor{mygray} \textbf{72.08} & \cellcolor{mygray} \textbf{62.75} & \cellcolor{mygray} \textbf{68.54} & \cellcolor{mygray} \textbf{61.93}\\ \hline
\bottomrule
\end{tabular}
\end{sc}
\end{scriptsize}
\end{center}
\end{table*}

To show that the improvement from our \sysname is consistent across different sparse training methods, apart from the comparison based on RigL, we also include more sparse training results using SET~\citep{mocanu2018scalable}. As shown in Tables~\ref{tb: ood-sparse-set-auroc}, our \sysname usually leads to larger AUROC, indicating more effective OOD detection from our \sysname compared to baseline methods.

\linespread{1.2}
\begin{table*}[ht]
\setlength\tabcolsep{5.5pt} 
\caption{Comparison of OOD detection by AUROC ($\uparrow$) between \texttt{MOON}+MSP and MSP for CIFAR-10 in sparse training using SET. Our \texttt{MOON} leads to larger AUROC on each OOD data of CIFAR-10, showing its ability to improve reliability on OOD data.}
\label{tb: ood-sparse-set-auroc}
\begin{center}
\begin{scriptsize}
\begin{sc}
\begin{tabular}{ccccccccccc}
\toprule \toprule
 &  &  CIFAR-100 & TIN   & NearOOD & & MNIST & SVHN  & Texture & Places365 & FarOOD \\
 \cline{1-2} \cline{3-5}   \cline{7-11}
 \multirow{2}{*}{80\%} & MSP        &   87.36 & 88.81 & 88.09 & & 93.41 & 88.70  & 88.51 & 88.77 & 89.85 \\
~ & \cellcolor{mygray} \texttt{MOON}+MSP &   \cellcolor{mygray} \textbf{89.52} & \cellcolor{mygray} \textbf{90.81} & \cellcolor{mygray} \textbf{90.16} & \cellcolor{mygray} & \cellcolor{mygray} \textbf{93.64} & \cellcolor{mygray} \textbf{92.47} & \cellcolor{mygray} \textbf{91.25} & \cellcolor{mygray} \textbf{90.46} & \cellcolor{mygray} \textbf{91.96} \\
 \multirow{2}{*}{90\%} & MSP        &   87.93 & 89.06 & 88.49 & & 93.54 & 90.26 & 88.40  & 88.57 & 90.19 \\
~ & \cellcolor{mygray} \texttt{MOON}+MSP   & \cellcolor{mygray} \textbf{88.85} & \cellcolor{mygray} \textbf{90.06} & \cellcolor{mygray} \textbf{89.45} & \cellcolor{mygray} & \cellcolor{mygray} \textbf{94.29} & \cellcolor{mygray} \textbf{92.34} & \cellcolor{mygray} \textbf{90.70}  & \cellcolor{mygray} \textbf{89.61} & \cellcolor{mygray} \textbf{91.74} \\
 \multirow{2}{*}{95\%} &MSP          & 86.81 & 87.83 & 87.32 & & \textbf{94.95} & 89.13 & 87.07 & 87.62 & 89.69 \\
~ & \cellcolor{mygray} \texttt{MOON}+MSP   & \cellcolor{mygray} \textbf{88.02} & \cellcolor{mygray} \textbf{89.33} & \cellcolor{mygray} \textbf{88.68} & \cellcolor{mygray} & \cellcolor{mygray} 93.78 & \cellcolor{mygray} \textbf{91.96} & \cellcolor{mygray} \textbf{90.09} & \cellcolor{mygray} \textbf{88.56} & \cellcolor{mygray} \textbf{91.10}  \\
 \multirow{2}{*}{99\%} &MSP         & 81.79 & 82.41 & 82.10  & & 90.13 & 87.08 & 81.31 & 81.77 & 85.07 \\
~ & \cellcolor{mygray} \texttt{MOON}+MSP &   \cellcolor{mygray} \textbf{83.10}  & \cellcolor{mygray} \textbf{83.84} & \cellcolor{mygray} \textbf{83.47} & \cellcolor{mygray} & \cellcolor{mygray} \textbf{91.13} & \cellcolor{mygray} \textbf{89.14} & \cellcolor{mygray} \textbf{84.92} & \cellcolor{mygray} \textbf{81.90}  & \cellcolor{mygray} \textbf{86.77} \\ \hline
\bottomrule
\end{tabular}
\end{sc}
\end{scriptsize}
\end{center}
\end{table*}

\begin{figure*}[!t]
\begin{center}
\subfigure[Temperature Scaling (99\%)]{\includegraphics[scale=0.42]{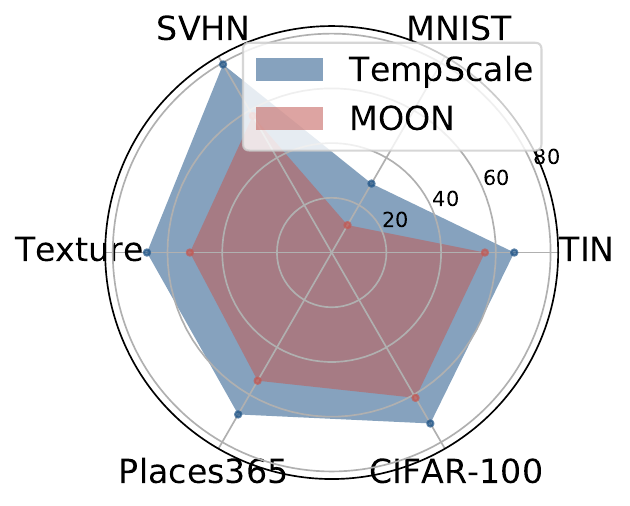}}
\subfigure[Mixup (99\%)]{\includegraphics[scale=0.42]{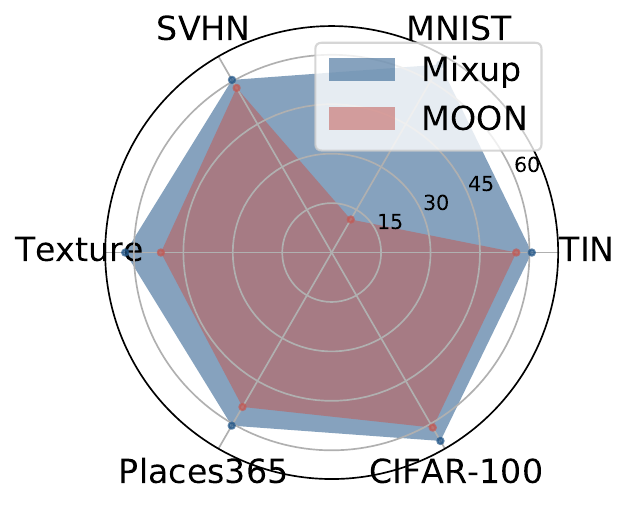}}
\subfigure[CigL (99\%)]{\includegraphics[scale=0.42]{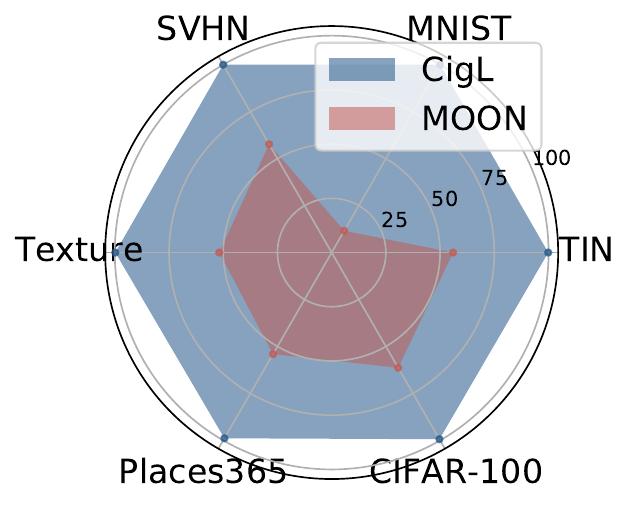}}
\caption{Comparison of OOD detection by FPR-95 ($\downarrow$) on CIFAR-10 between \texttt{MOON} and other calibration methods using RigL (90\%). The red hexagons (\texttt{MOON}) are smaller than the blue hexagons (other calibration methods), indicating a better OOD detection using \texttt{MOON} compared to (a) Temperature Scaling, (b) Mixup, and (c) CigL.}
\label{fig: ood-calib-sp-appen}
\end{center}
\end{figure*}

Prior work shows that calibration methods can help with OOD detection in dense DNNs~\citep{wang2021energy, yang2022openood}. To show the ineffectiveness of existing calibration methods for OOD detection in sparse training, we also compare our \texttt{MOON} to other confidence calibration methods including temperature scaling~\citep{lee2018simple}, Mixup~\citep{zhang2017mixup}, and CigL~\citep{lei2023calibrating} using RigL for OOD detection capability in 99\% sparsity. As shown in Figure~\ref{fig: ood-calib-sp-appen}, the red and blue hexagons represent the FPR-95 of \texttt{MOON} and other calibration methods, respectively. We can see that the red hexagons are smaller than the blue hexagons, indicating better OOD detection from \texttt{MOON} compared to other calibration methods.

\begin{figure}[!t]
\begin{center}
\subfigure[CIFAR-10, RigL]{\includegraphics[scale=0.17]{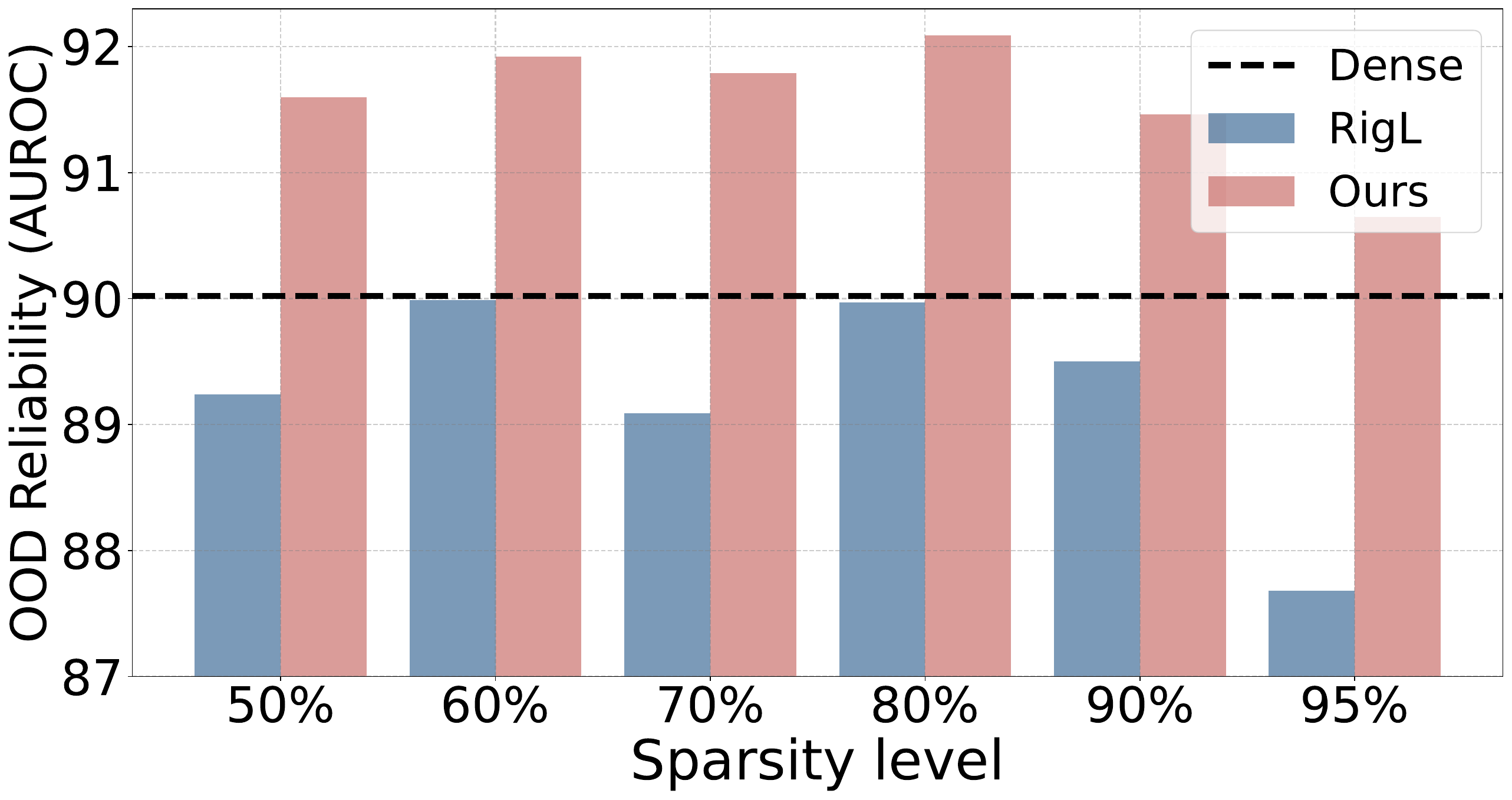}}
\subfigure[CIFAR-100, RigL]{\includegraphics[scale=0.17]{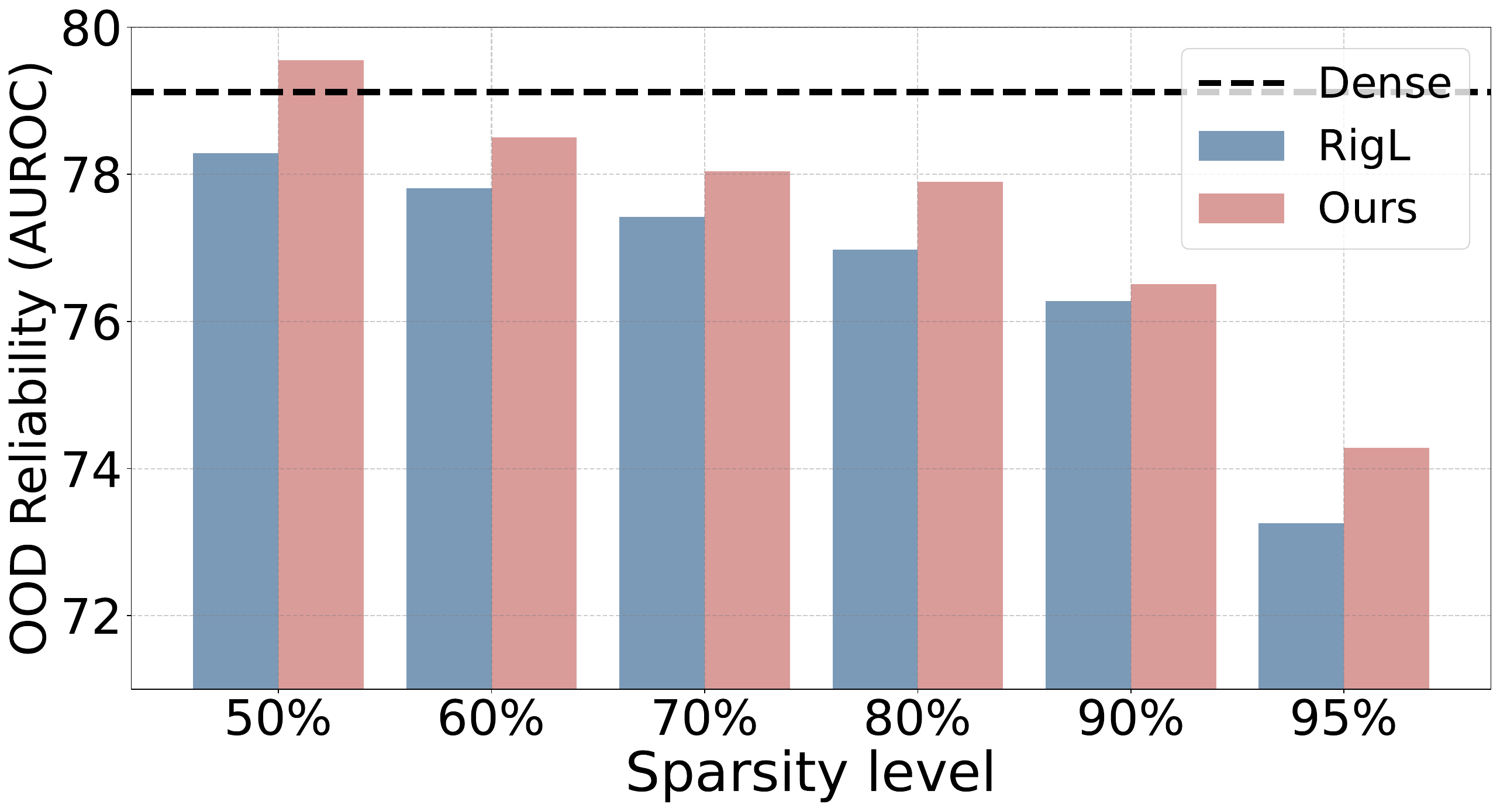}}
\caption{OOD reliability (measured by AUROC (\%), the higher the better) of the ResNet-18 produced by dense and sparse training (RigL \& SET) on CIFAR-10/100. Compared to dense training (black line), sparse training (blue bar) has a smaller AUROC, indicating sparse training exacerbates the unreliability on OOD data. Our \sysname (red bar) improves AUROC and OOD detection.}
\label{fig: intro-low-sp}
\end{center}
\end{figure}

To check the effectiveness of our \sysname at low sparsity level, we add additional experiments at 50\%, 60\%, and 70\% sparsity on CIFAR-10 and CIFAR-100. The detailed results are summarized in Figure~\ref{fig: intro-low-sp}.
Our \sysname can consistently improve the OOD reliability at different sparsity levels.

\subsection{More Results on MNIST}

To show that the improvement from our \sysname is consistent across different benchmark datasets, apart from the comparison on CIFAR-10, CIFAR-100, and ImageNet-2012, we also include more results on MNIST. As shown in Tables~\ref{tb: ood-mnist}, our \texttt{MOON} usually leads to larger AUROC and smaller FPR-95, indicating more effective OOD detection from our \texttt{MOON} compared to baseline methods.

\linespread{1.2}
\setlength\tabcolsep{4pt}
\begin{table*}[!t]
\caption{Comparison of OOD detection by FPR-95 (FPR, $\downarrow$) \& AUROC (AUC, $\uparrow$) between \texttt{MOON} and MSP for MNIST. Our \texttt{MOON} leads to smaller FPR-95 and larger AUROC on each OOD data, showing its ability to improve reliability on OOD data.}
\label{tb: ood-mnist}
\begin{center}
\begin{scriptsize}
\begin{sc}
\begin{tabular}{lcccccccccc}
\toprule \toprule
& &  NotMNIST & FashionMNIST   & NearOOD & & Texture & CIFAR10  & TIN & Places365 & FarOOD \\
 \cline{1-2}  \cline{3-5}   \cline{7-11}
\multirow{2}{*}{\rotatebox{90}{FPR}} & MSP   & 40.81 & 30.46 & 35.64 & & 7.43  & 7.11  & 6.11  & 9.28  & 7.48 \\
~ & \cellcolor{mygray} \texttt{MOON}+MSP & \cellcolor{mygray} \textbf{3.57}  & \cellcolor{mygray} \textbf{7.08}  & \cellcolor{mygray} \textbf{5.32}  & \cellcolor{mygray} & \cellcolor{mygray} \textbf{1.76}  & \cellcolor{mygray} \textbf{0.91}  & \cellcolor{mygray} \textbf{1.41}  & \cellcolor{mygray} \textbf{1.32}  & \cellcolor{mygray} \textbf{1.35}  \\
\hline
\multirow{2}{*}{\rotatebox{90}{AUC}} & MSP  &  89.20  & 93.69 & 91.45 & & 98.55 & 98.55 & 98.75 & 98.20  & 98.51 \\
~ & \cellcolor{mygray} \texttt{MOON}+MSP  &  \cellcolor{mygray} \textbf{98.07} & \cellcolor{mygray} \textbf{96.54} & \cellcolor{mygray} \textbf{97.31} & \cellcolor{mygray} & \cellcolor{mygray} \textbf{98.86} & \cellcolor{mygray} \textbf{99.18} & \cellcolor{mygray} \textbf{99.06} & \cellcolor{mygray} \textbf{99.12} & \cellcolor{mygray} \textbf{99.05} \\ \hline
\bottomrule
\end{tabular}
\end{sc}
\end{scriptsize}
\end{center}
\end{table*}

\subsection{OOD Detection in Dense Training}\label{subsec:ood-dense}

In this section, to show the broad applicability of our \texttt{MOON}, we show our \texttt{MOON} can improve OOD detection in dense training. The results of baseline methods are based on the scores reported in~\citep{yang2022openood}.

\textbf{For the model trained on CIFAR-10}, we examine its OOD detection ability on two near OOD data (i.e., CIFAR-100, TIN) and four far OOD data (i.e., MNIST, SVHN, Texture, Places365). As shown in Table~\ref{tb: ood-comp}, the columns named "NearOOD" and "FarOOD" represent the average detection scores of the near OOD data and the far OOD data, respectively. We can see that our \texttt{MOON} provides smaller FPR-95 and larger AUROC for different near and far OOD data compared to MSP, ODIN, and EBO, where the reduction of FPR-95 can be up to 77.6\% and the improvement of AUROC can be up to 27.3\%. This demonstrates that our \texttt{MOON} can perform OOD detection more effectively compared to the original methods.

\linespread{1.2}
\setlength\tabcolsep{5.5pt}
\begin{table*}[!t]
\caption{Comparison of OOD detection by FPR-95 ($\downarrow$) \& AUROC ($\uparrow$) between \sysname and baseline post process methods for CIFAR-10. Our \sysname leads to smaller FPR-95 and larger AUROC on each OOD data, showing its ability to improve reliability on OOD data.}
\label{tb: ood-comp}
\begin{center}
\begin{scriptsize}
\begin{sc}
\begin{tabular}{lcccccccccc}
\toprule \toprule
& &  CIFAR-100 & TIN   & NearOOD & & MNIST & SVHN  & Texture & Places365 & FarOOD \\
 \cline{1-2}  \cline{3-5}   \cline{7-11}
\multirow{6}{*}{\rotatebox{90}{FPR-95}} & MSP   & 62.01     & 60.69 & 61.35  & & 58.59 & 51.87 & 59.89   & 57.64     & 57.00     \\
~ & \cellcolor{mygray} \texttt{MOON}+MSP & \cellcolor{mygray} \textbf{61.01}     & \cellcolor{mygray} \textbf{56.87} & \cellcolor{mygray} \textbf{58.94}  & \cellcolor{mygray} & \cellcolor{mygray} \textbf{40.91} & \cellcolor{mygray} \textbf{48.25} & \cellcolor{mygray} \textbf{51.31}   & \cellcolor{mygray} \textbf{57.56}     & \cellcolor{mygray} \textbf{49.50} \\
~ & ODIN  &  59.09     & 59.06 & 59.07 &  & 36.23 & 67.92 & 51.10    & 50.51     & 51.44     \\
~ & \cellcolor{mygray} \texttt{MOON}+ODIN  & \cellcolor{mygray} \textbf{48.47}     & \cellcolor{mygray} \textbf{42.24} & \cellcolor{mygray} \textbf{45.36}  & \cellcolor{mygray} & \textbf{8.10}   & \cellcolor{mygray} \textbf{33.30}  & \textbf{32.58}   & \cellcolor{mygray} \textbf{40.64}     & \cellcolor{mygray} \textbf{28.65} \\
~ & EBO  &  51.46     & 45.02 & 48.24  & & 44.50  & 44.94 & 48.32   & 41.88     & 44.91     \\
~ & \cellcolor{mygray} \texttt{MOON}+EBO   & \cellcolor{mygray} \textbf{47.41}     & \cellcolor{mygray} \textbf{39.29} & \cellcolor{mygray} \textbf{43.35} &  \cellcolor{mygray} & \cellcolor{mygray} \textbf{14.98} &  \cellcolor{mygray} \textbf{21.34} & \cellcolor{mygray} \textbf{36.75}   & \cellcolor{mygray} \textbf{38.51}     & \cellcolor{mygray} \textbf{27.90} \\ 
\hline
\multirow{6}{*}{\rotatebox{90}{AUROC}} & MSP  &  87.11     & 86.62 & 86.87   & & 89.91 & 90.88 & 88.72   & 89.03     & 89.64  \\
~ & \cellcolor{mygray} \texttt{MOON}+MSP  &  \cellcolor{mygray} \textbf{89.06}     & \cellcolor{mygray} \textbf{90.61} & \cellcolor{mygray} \textbf{89.83}   & \cellcolor{mygray} & \cellcolor{mygray} \textbf{94.11} & \cellcolor{mygray} \textbf{93.83} & \cellcolor{mygray} \textbf{92.30}    & \cellcolor{mygray} \textbf{89.97}     & \cellcolor{mygray} \textbf{92.55}  \\
~ & ODIN   & 77.68     & 77.33 & 77.51  &  & 90.91 & 73.32 & 80.70    & 82.55     & 81.87  \\
~ & \cellcolor{mygray} \texttt{MOON}+ODIN  & \cellcolor{mygray} \textbf{87.91}     & \cellcolor{mygray} \textbf{90.26} & \cellcolor{mygray} \textbf{89.08}   & \cellcolor{mygray} & \cellcolor{mygray} \textbf{98.34} & \cellcolor{mygray} \textbf{93.37} & \cellcolor{mygray} \textbf{92.97}   & \cellcolor{mygray} \textbf{90.17}     & \cellcolor{mygray} \textbf{93.71}  \\
~ & EBO   & 86.15     & 88.58 & 87.36  &  & 90.59 & 88.39 & 86.85   & 89.60      & 88.86  \\
~ & \cellcolor{mygray} \texttt{MOON}+EBO   & \cellcolor{mygray} \textbf{90.05}     & \cellcolor{mygray} \textbf{92.36} & \cellcolor{mygray} \textbf{91.20}  &  \cellcolor{mygray}  & \cellcolor{mygray} \textbf{97.30}  & \cellcolor{mygray} \textbf{96.15} & \cellcolor{mygray} \textbf{93.34}   & \cellcolor{mygray} \textbf{92.04}     & \cellcolor{mygray} \textbf{94.71}\\ \hline
\bottomrule
\end{tabular}
\end{sc}
\end{scriptsize}
\end{center}
\end{table*}

\textbf{For the model trained on ImageNet-2012}, we examine its OOD detection ability on four near OOD data (i.e., Species, iNaturalist, OpenImage-O, ImageNet-O) and two far OOD data (i.e., Texture, MNIST). More details about the OOD data are in Appendix~\ref{subsec:ood-des}. As shown in Figure~\ref{fig: ood-comp}, the red and blue bars represent our \texttt{MOON} and MSP, respectively. We can see that our \texttt{MOON} provides larger AUROC and AUPR, and smaller FPR-95 almost on all of the OOD data of ImageNet-2012, showing its effective OOD detection.

\begin{figure*}[ht]
\begin{center}
\subfigure[AUROC ($\uparrow$)]{\includegraphics[scale=0.19]{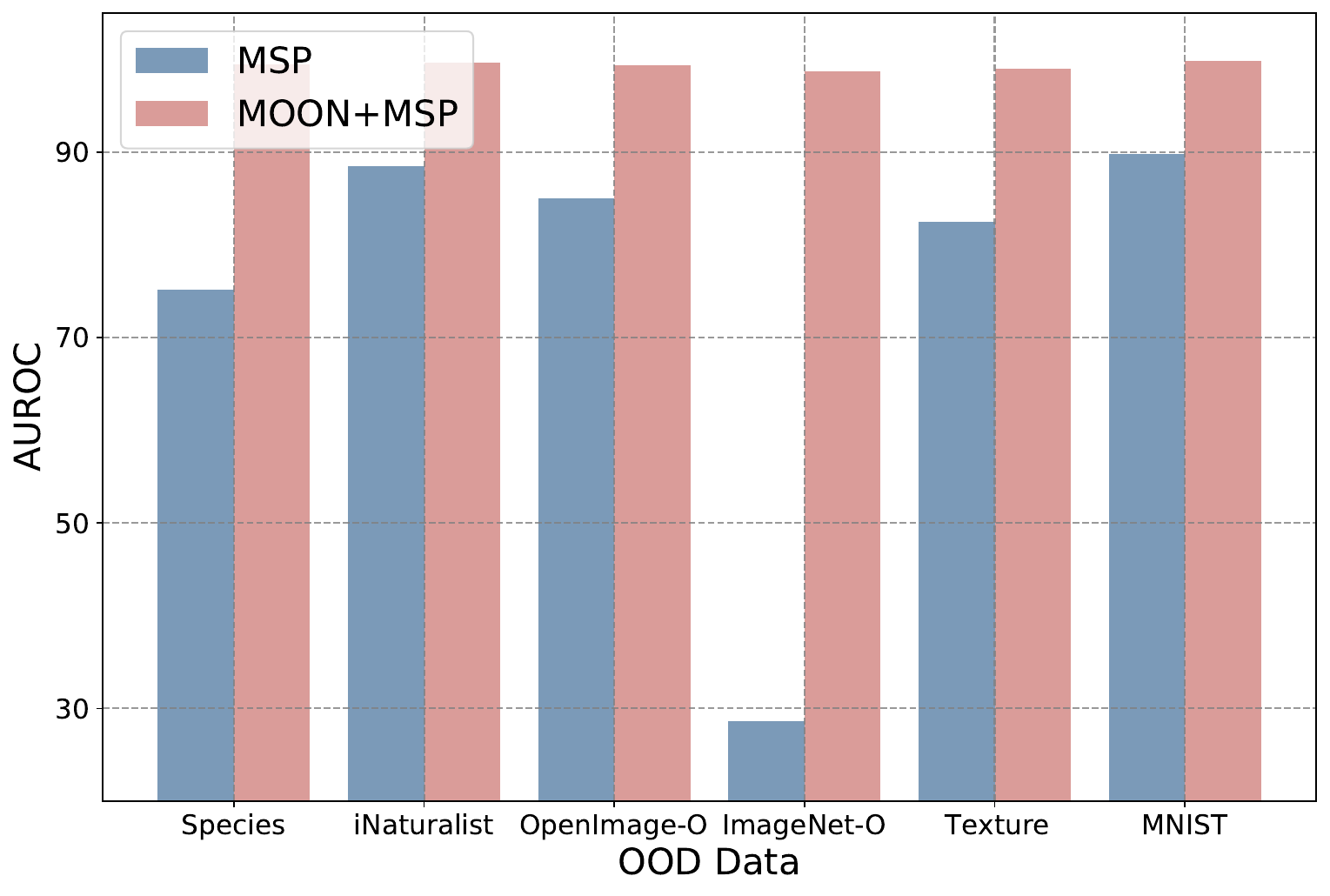}}
\subfigure[AUPR ($\uparrow$)]{\includegraphics[scale=0.19]{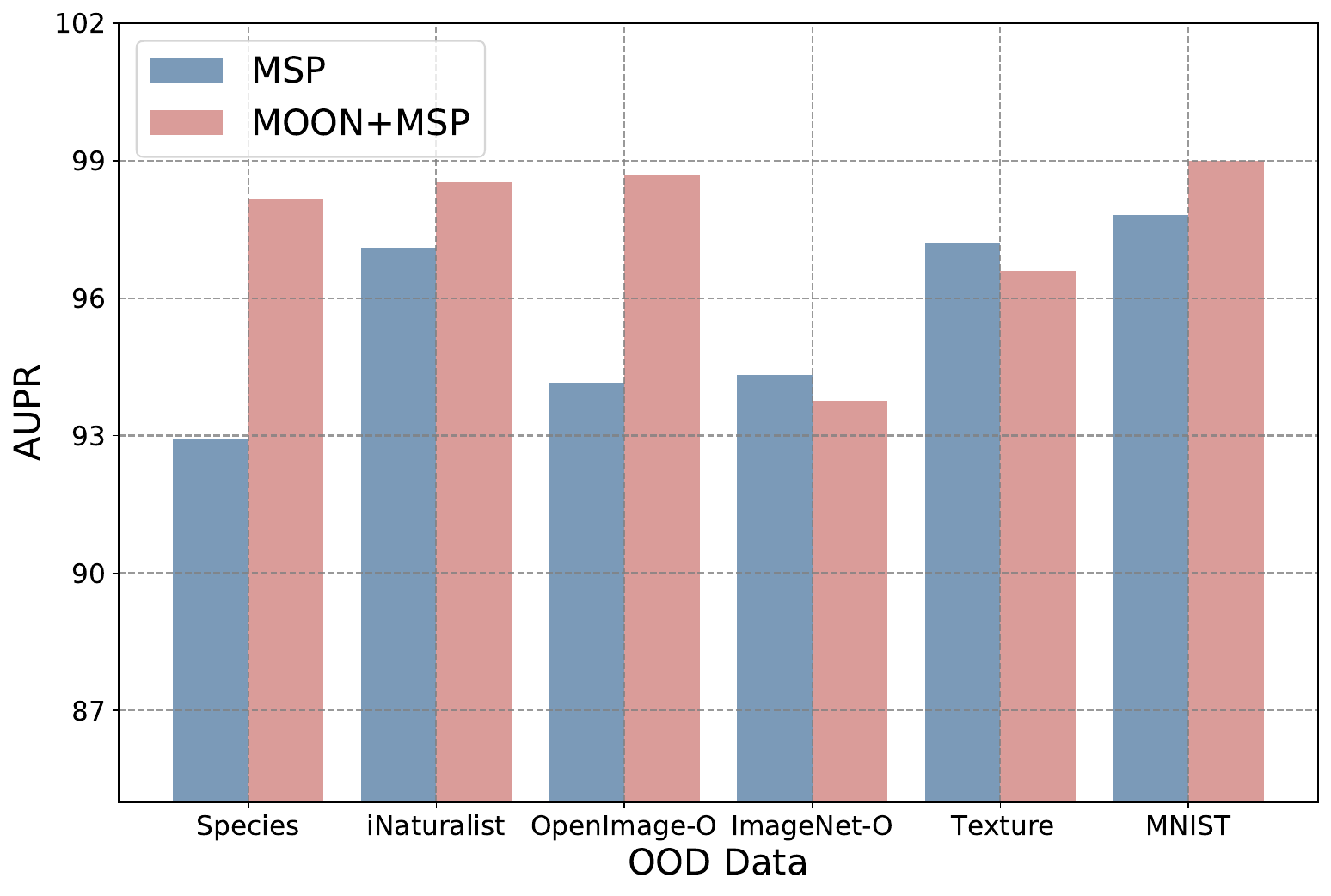}}
\subfigure[FPR-95 ($\downarrow$)]{\includegraphics[scale=0.19]{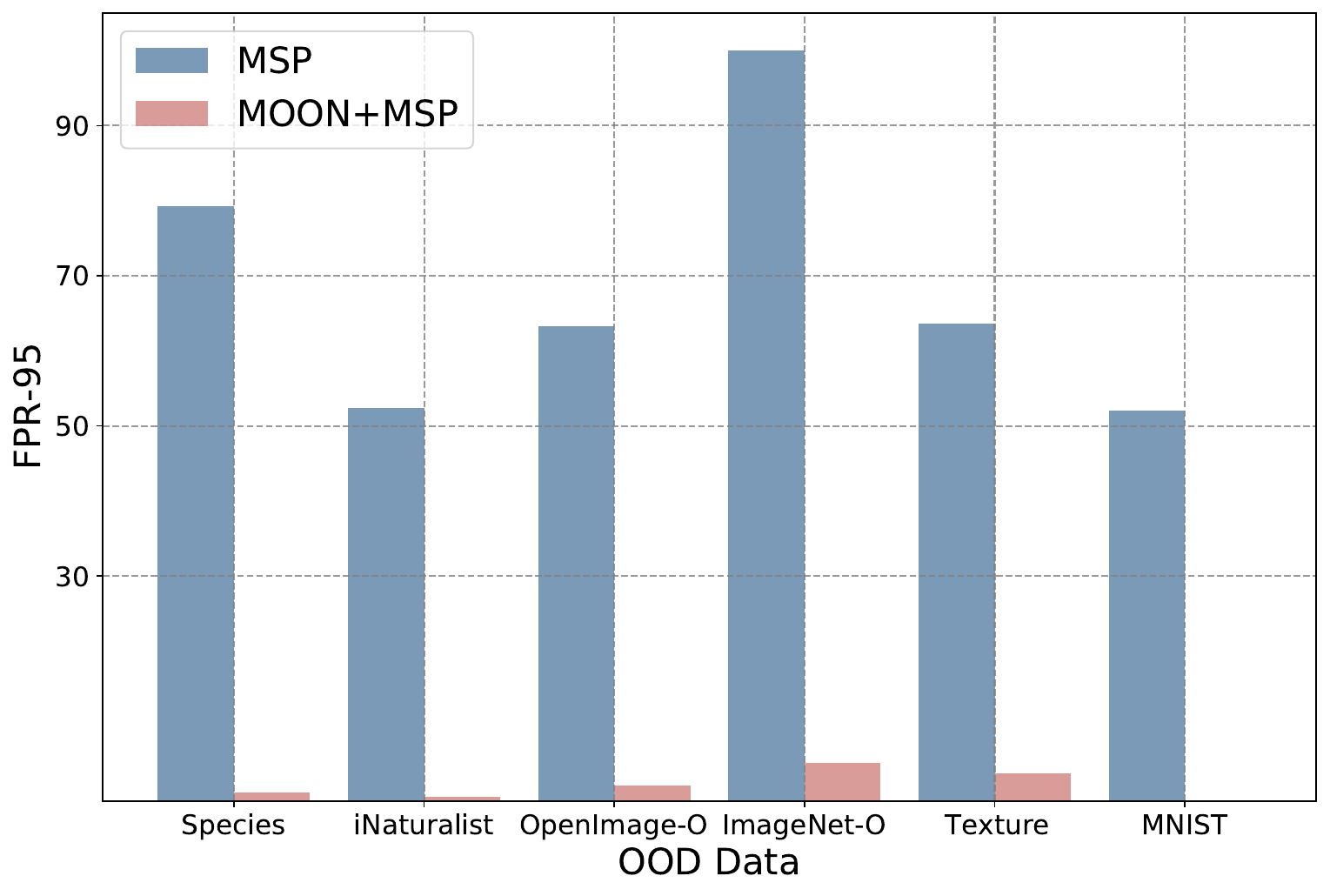}}
\caption{Comparison of OOD detection by AUROC ($\uparrow$), AUPR ($\uparrow$), and FPR-95 ($\downarrow$) between \texttt{MOON}+MSP and MSP for ImageNet-2012. Our \texttt{MOON} leads to larger AUROC and AUPR, and smaller FPR-95 on each OOD data of ImageNet-2012, showing its ability to improve reliability on OOD data.}
\label{fig: ood-comp}
\end{center}
\end{figure*}

\subsection{More Ablation Studies for $w_f$}

We add additional ablation studies for $w_f$ to test the sensitivity of $w_f$. As summarized in Appendix G, we set $w_f=1$ for CIFAR-10. We add experiments of 95\% sparse ResNet-18 with $w_f=0.1, 0.5, 2, 4$. The AUROC results are summarized in the table below. We also include them in Appendix A.4. As shown in the Table~\ref{tb: ood-wf-sen}, when we increase $w_f$ from 0.1 to 4, OOD reliability first increases and then decreases after 1. While OOD reliability may be degraded when $w_f$ is too small or too large, OOD reliability is not degraded much when $w_f$ is between 0.5 and 2. The proper $w_f$ value can be found with a simple hyperparameter tuning.

\linespread{1.2}
\setlength\tabcolsep{5.5pt}
\begin{table*}[!t]
\caption{More Ablation Studies for $w_f$ by AUROC ($\uparrow$) in \sysname with 95\% sparse ResNet-18 for CIFAR-10.}
\label{tb: ood-wf-sen}
\begin{center}
\begin{scriptsize}
\begin{sc}
\begin{tabular}{lcccccccccc}
\toprule \toprule
& &  CIFAR-100 & TIN   & NearOOD & & MNIST & SVHN  & Texture & Places365 & FarOOD \\
 \cline{1-2}  \cline{3-5}   \cline{7-11}
\multirow{5}{*}{\rotatebox{90}{AUROC}} & $w_f=0.1$   &  83.57  & 85.25 &  84.41  &  & 86.49  & 86.47 &  85.35  &  84.73  & 85.76    \\
~ & $w_f=0.5$  &  86.25&  87.07& 87.66&  &  93.46 &  91.02 & 87.80 &  85.11 &  89.35     \\
~ & \cellcolor{mygray} $w_f=1.0$ & \cellcolor{mygray} \textbf{88.04} & \cellcolor{mygray} \textbf{89.39} & \cellcolor{mygray} \textbf{88.71} & \cellcolor{mygray} & \cellcolor{mygray} \textbf{94.70} & \cellcolor{mygray} \textbf{93.35} & \cellcolor{mygray} \textbf{89.86} & \cellcolor{mygray} \textbf{88.51} & \cellcolor{mygray} \textbf{91.62} \\
~ & $w_f=2.0$  &  86.45 & 87.77 & 87.11 & & 89.96 & 91.44 & 89.24 & 86.05 & 89.17     \\
~ & $w_f=4.0$  &  84.78 & 85.84 & 85.31 & & 94.69 & 91.44 & 86.66 & 84.65 & 89.36  \\
\hline
\bottomrule
\end{tabular}
\end{sc}
\end{scriptsize}
\end{center}
\end{table*}

\subsection{Additional Incremental Ablation Study}

We add an incremental ablation study in which each component of the method is incremented to the previous component to better intuitively understand the utility of each component. We train 95\% sparse ResNet-18 on CIFAR-10 with MOON + RigL. We find that the OOD's reliability decreases as we gradually remove one of its components.

For MOON w/o LM, we set $w_f$ as 0.5, 0.001, and 0.00001, respectively, to incrementally remove the loss modification component. The AUROC (\%) results for MOON w/o LM are summarized in the Table~\ref{tb: inc-ab-lm}. $w_f$ = 1 refers to our MOON.

\linespread{1.2}
\setlength\tabcolsep{5.5pt}
\begin{table*}[ht]
\caption{Additional incremental ablation study for loss modification (LM) by AUROC ($\uparrow$) in \sysname with 95\% sparse ResNet-18 for CIFAR-10.}
\label{tb: inc-ab-lm}
\begin{center}
\begin{scriptsize}
\begin{sc}
\begin{tabular}{ccccc}
\toprule \toprule
& $w_f$ = 1 & $w_f$ = 0.5 & $w_f$ = 0.001 & $w_f$ = 0.00001 \\
\hline 
NearOOD & \textbf{88.71} & 86.97 & 86.52 & 85.76     \\
FarOOD & \textbf{91.62} & 90.18 & 89.41 & 87.76\\
\hline
\bottomrule
\end{tabular}
\end{sc}
\end{scriptsize}
\end{center}
\end{table*}

For MOON w/o AT, we shorten $T_e$ to 3, 2, and 1, respectively, to incrementally remove the auto tuning component. The AUROC (\%) results for MOON w/o AT are summarized in the Table~\ref{tb: inc-ab-at}. $T_e$ = 5 refers to our MOON.

\linespread{1.2}
\setlength\tabcolsep{5.5pt}
\begin{table*}[ht]
\caption{Additional incremental ablation study for auto tuning (AT) by AUROC ($\uparrow$) in \sysname with 95\% sparse ResNet-18 for CIFAR-10.}
\label{tb: inc-ab-at}
\begin{center}
\begin{scriptsize}
\begin{sc}
\begin{tabular}{ccccc}
\toprule \toprule
& $T_e$ = 5 &  $T_e$ = 3 &  $T_e$ = 2 &  $T_e$ = 1 \\
\hline 
NearOOD & \textbf{88.71} &  85.81 &  85.33 &  85.17     \\
FarOOD & \textbf{91.62} & 88.42 & 88.32 & 87.97\\
\hline
\bottomrule
\end{tabular}
\end{sc}
\end{scriptsize}
\end{center}
\end{table*}

For MOON w/o VT, we shorten the length of epochs using the voting scheme at the end of training to 10, 5, and 2 epochs, respectively, to incrementally remove the voting scheme component. The AUROC (\%) results for MOON w/o LM are summarized in the Table~\ref{tb: inc-ab-vt}. Vote time = 20 refers to our MOON.

\linespread{1.2}
\setlength\tabcolsep{5.5pt}
\begin{table*}[ht]
\caption{Additional incremental ablation study for voting scheme (VT) by AUROC ($\uparrow$) in \sysname with 95\% sparse ResNet-18 for CIFAR-10.}
\label{tb: inc-ab-vt}
\begin{center}
\begin{scriptsize}
\begin{sc}
\begin{tabular}{ccccc}
\toprule \toprule
& vote time = 20 & vote time = 10 & vote time = 5 & vote time = 2 \\
\hline 
NearOOD & \textbf{88.71} & 85.84 & 85.32 & 84.77    \\
FarOOD & \textbf{91.62} & 88.16 & 86.78 & 86.76\\
\hline
\bottomrule
\end{tabular}
\end{sc}
\end{scriptsize}
\end{center}
\end{table*}

%% file: sections/_b_Exp_del.tex
\section{Appendix: Additional Details about Experiment Settings}

\subsection{OOD Benchmark Description}\label{subsec:ood-des}

The choice of out-of-distribution (OOD) Benchmarks is according to OpenOOD~\citep{yang2022openood}. The distribution here refers to the~"label distribution". Following the common practice of constructing OOD detection benchmarks, we treat the entire dataset as in-distribution (ID) and then gather several datasets that are not correlated with any ID category as OOD datasets. We use four OOD benchmarks, which are named after ID datasets, including MNIST~\citep{deng2012mnist}, CIFAR-10~\citep{krizhevsky2009learning}, CIFAR-100~\citep{krizhevsky2009learning}, and ImageNet~\citep{russakovsky2015imagenet}. As mentioned by OpenOOD~\citep{yang2022openood}, compared to the ID dataset, the near-OOD dataset has only semantic shifts, while the far OOD further contains significant covariate (domain) shifts. The detailed description of each benchmark is as follows.

\textbf{For MNIST}~\citep{deng2012mnist}, it is a 10-class handwriting digit dataset with 60k images for training and 10k for test. Its near-OOD datasets include NOTMNIST~\citep{yang2021generalized} and FashionMNIST~\citep{hsu2020generalized}. The far-OOD datasets include Texture~\citep{ahmed2020detecting}, CIFAR-10~\citep{tax2002one} and TinyImageNet~\citep{han2022adbench}, and Places-365~\citep{guo2017calibration}. 

\textbf{For CIFAR-10}~\citep{krizhevsky2009learning}, it is a 10-class dataset for general object classification with 50k training images and 10k for test. Its near-OOD datasets include CIFAR-100~\citep{chandola2009anomaly} and TinyImageNet~\citep{han2022adbench}, where 1,207 images are removed from TinyImageNet since they belong to CIFAR-10 classes~\citep{yang2022openood}. The far-OOD datasets include MNIST~\citep{deng2012mnist}, SVHN~\citep{netzer2011reading}, Texture~\citep{ahmed2020detecting}, and Places365~\citep{guo2017calibration} with 1,305 images removed due to semantic overlaps.

\textbf{For CIFAR-100}~\citep{krizhevsky2009learning}, it is a 100-class dataset for general object classification with 50k training images and 10k for test. Its near-OOD datasets include CIFAR-10~\citep{tax2002one} and TinyImageNet~\citep{han2022adbench}, where 2,502 images are removed from TinyImageNet due to the overlapping semantics with CIFAR-100 classes~\citep{yang2022openood}. Its far-OOD include MNIST~\citep{deng2012mnist}, SVHN~\citep{netzer2011reading}, Texture~\citep{ahmed2020detecting}, and Places365~\citep{guo2017calibration} with 1,305 images removed due to semantic overlaps.

\textbf{For ImageNet}, it is a large-scale image classification dataset with 1000 classes. OpenOOD~\citep{yang2022openood} build the OOD dataset, where they use a 10k subset of Species~\citep{torralba200880} with 713k images, iNaturalist~\citep{shorten2019survey} with 10k images, ImageNet-O~\citep{li2021cutpaste} with 2k images, and OpenImage-O~\citep{sun2022out} with 17k images as near-OOD datasets, and use Texture~\citep{ahmed2020detecting}, MNIST~\citep{deng2012mnist} as far-OOD. All images belonging to the ID classes are deleted.

\subsection{Computing Resources}

The training and evaluation are mainly on NVIDIA Quadro RTX 6000. For CIFAR-10 and CIFAR-100, training ResNet-18 for 100 epochs with our \sysname using 1 GPU takes around 1.5 hours. For ImageNet-12,  training ResNet-50 for 100 epochs with our \sysname using 1 GPU takes around 6 days.

%% file: sections/_c_Theoty.tex
\section{Appendix: Theoretical Analysis}

In this section, we provide detailed proof of our insights in Section 4 of the main manuscript. The proof is based on Assumption~\ref{appen:assup:gp}, which describes the characteristic of the feature space.

\begin{assumption}
(Gaussian Mixture Feature Space): The feature mapping function $h$ maps the input ID data to a Gaussian mixture $v_1\mathcal{N}(\mu_1, \Sigma_1) + v_2\mathcal{N}(\mu_2, \Sigma_2)$. Specifically, when $y=1$, we have $h(x)\sim \mathcal{N}(\mu_1, \Sigma_1)$. And when $y=2$, we have $f(x)\sim \mathcal{N}(\mu_2, \Sigma_2)$.
\label{appen:assup:gp}
\end{assumption}

\subsection{Insight 4.4: Unreliability}

(Unreliability) suppose we have $(x_1, y_1)$ from class 1 and $D_2 = \{(x, y); ||h(x)-h(x_1)||<\epsilon, y=2 \}$ from class 2. Then, unreliability can occur around  $h(x_1)$.

\textit{Proof}:

The final output probability is a composition of the feature mapping function $h$ and the softmax classification function $g$. As we know, the softmax function is Lipschitz continuous with a Lipschitz constant $c$ bounded by 1.

Because of the definition of $D_2$, we can know that
\begin{align}
    ||g(h(x)) - g(h(x_1))|| \leq c||h(x)-h(x_1)||.
\end{align}
Suppose $g(h(x_1))=\{p_{11}, p_{12}\}$ and $g(h(x_2))=\{p_{21}, p_{22}\}$ for any $x_2 \in D_2$. Then we can know that
\begin{align}
    ||p_{11} - p_{21}|| \leq c\epsilon.
\end{align}
Without loss of generality, suppose $x_1$ is predicted correctly, i.e., we have $p_{11}>0.5$. Then, we can have 
\begin{align}
    p_{21} \geq p_{11} - c\epsilon > 0.5, \ \epsilon \rightarrow 0.
\end{align}
Then, with small $\epsilon$, the samples in $D_2$ will have $p_{21}>0.5$ and be incorrectly predicted to class 1. Thus, for any $\epsilon_0<\epsilon$, we can define $D_2(\epsilon_0)$ and have its expected accuracy as:
\begin{align}
    D_2(\epsilon_0) &= \{(x, y); ||h(x)-h(x_1)||<\epsilon_0, y=2 \},\\
    \mathbf{E}_{D_2(\epsilon_0)} & [1\{ \hat{y}=y \}] = \mathbf{E}_{D_2(\epsilon_0)} [1\{ 1=2 \}]  = 0
\end{align}
In this case, suppose the samples near $x_1$ are $D(\epsilon_0)$. And $D(\epsilon_0)$ can be divided into $D_1(\epsilon_0)$ and $D_2(\epsilon_0)$. $D_i=\{(x, y); ||h(x)-h(x_1)||<\epsilon_0, y=i \}$. The ratio of the size of $D_1$ and $D_2$ can be approximated by
\begin{align}
    \frac{|D_1|}{|D_2|} = \frac{w_1}{w_2} = \frac{\mathcal{N}(h(x_1) ; \mu_{1}, \Sigma_{1})}{\mathcal{N}(h(x_1) ; \mu_{2}, \Sigma_{2})},\ \text{where} \ \ w_1+w_2=1.
\end{align}
Then, we can calculate the unreliability level following the Definition 4.2:
\begin{align}
    &\mathbf{E}_{D(\epsilon)} [ \max_{c\in \{1,2\}} \mathcal{N}(h(x) ; \mu_{c}, \Sigma_{c})] - \mathbf{E}_{D(\epsilon)} [1\{ \hat{y}=y \}] \nonumber  \\
    &=w_1 \bigg{(}\mathbf{E}_{D_1(\epsilon)} [ \max_{c\in \{1,2\}} \mathcal{N}(h(x) ; \mu_{c}, \Sigma_{c})] - \mathbf{E}_{D_1(\epsilon)} [1\{ \hat{y}=y \}] \bigg{)}\nonumber  \\
    &\ \ \ \ + w_2 \bigg{(}\mathbf{E}_{D_2(\epsilon)} [ \max_{c\in \{1,2\}} \mathcal{N}(h(x) ; \mu_{c}, \Sigma_{c})] - \mathbf{E}_{D_2(\epsilon)} [1\{ \hat{y}=y \}] \bigg{)} \nonumber \\
    & = w_1 \bigg{(}\mathbf{E}_{D_1(\epsilon)} [ \max_{c\in \{1,2\}} \mathcal{N}(h(x) ; \mu_{c}, \Sigma_{c})] - 1 \bigg{)} +
    w_2 \bigg{(}\mathbf{E}_{D_2(\epsilon)} [ \max_{c\in \{1,2\}} \mathcal{N}(h(x) ; \mu_{c}, \Sigma_{c})] - 0 \bigg{)} \nonumber \\
    & \geq p_{11} - c\epsilon - w_1 \nonumber \\
\end{align}
Then, we can choose $x_1$ which is not very close to $\mu_1$ such that $w_1 < p_{11} - c\epsilon -\eta$. Then, we will have:
\begin{align}
    \mathbf{E}_{D(\epsilon)} [ \max_{c\in \{1,2\}} \mathcal{N}(h(x) ; \mu_{c}, \Sigma_{c})] - \mathbf{E}_{D(\epsilon)} [1\{ \hat{y}=y \}] > \geq p_{11} - c\epsilon - w_1 > \eta.
\end{align}
The proof is complete.

\subsection{Insight 4.6: Hard-ID Reliability}

(Hard-ID Reliability) Suppose we have the same $(x_1, y_1)$ and $D_2$ as in Insight 4.4. If the model is trained with our \sysname method and the extra dimension successfully stores the unknown information, the unreliability can be solved, i.e.,
\begin{equation}
    \mathbf{E}_{D(\epsilon)} [\max_{c\in \{1,2\}} \mathcal{N}(h(x) ; \mu_{c}, \Sigma_{c})] - \mathbf{E}_{D(\epsilon)} [1\{ \hat{y}=y \}] < \eta.    
\end{equation}

\textit{Proof}:

Following the notations in the proof of Insight 4.4, we can know that unreliability around $x_1$ has two parts. One is from $D_1$ and the other is from $D_2$. And the majority of the unreliability comes from $D_2$. Thus, in order to reduce the discrepancy between confidence and accuracy, we need to mainly reduce the part of $D_2$, which is:
\begin{align}
    \mathbf{E}_{D_2(\epsilon)} [ \max_{c\in \{1,2\}} \mathcal{N}(h(x) ; \mu_{c}, \Sigma_{c})] - \mathbf{E}_{D_2(\epsilon)} [1\{ \hat{y}=y \}] 
\end{align}
Suppose the two classes of data are balanced. Since $x_1$ is predicted correctly, we can know that $w_1 > w_2$ and $|D_1|>|D_2|$. 

In this case, if the model can find a better feature mapping $h$ that does not have such a discrepancy for any set of data $D$, then there is no safety problem. However, if finding such a suitable feature mapping $h$ is too difficult, which is usually the case, we can only change the softmax classification function $g$. In sparse training, the update route is cut off, generating spurious local optimization, leading to a more challenging exploration of the weight space and thus an inability to find a suitable feature mapping $h$.

If we choose the widely-used cross entropy loss $L$ which in our case will take the form as below:
\begin{align}
    L(D) &= L(D_1) + L(D_2) \nonumber\\
    & = - \sum_{i \in D_1} \log p_{i1} - \sum_{i \in D_2} \log p_{i2}\nonumber\\
    & = - \sum_{i \in D_1} \log p_{i1} - \sum_{i \in D_2} \log (1-p_{i1})
\end{align}
Since it is difficult to find a better mapping $h$, we assume $h$ is fixed in this case. If $\epsilon$ is small, then the gradient of samples in $D(\epsilon)$ will be similar to that of $x_1$. Suppose $g(h(x_1))=(p_1^*, p_2^*)$. Then, we can have:
\begin{align}
    L(D) &= - \sum_{i \in D_1} \log p_{i1} - \sum_{i \in D_2} \log (1-p_{i1})\\
    & \approx - |D_1|\log p_1^* - |D_2| \log (1 - p_{i1})
\end{align}
And the gradient of loss will be:
\begin{align}
    \nabla L(D) & \propto \frac{(|D_2| + |D_1|) * p_1^* - |D_1|}{p_1^*(1-p_1^*)} 
\end{align}
Thus, to reduce the loss, $p_1^*$ will stuck at $\frac{|D_1|}{|D_1|+|D_2|}$ and can not reduce the discrepancy, implying severe unreliability in $D$.

If we choose our \sysname, there will be three dimensions in the output probability and $g(h(x_1))=(p_1^*, p_2^*, p_3^*)$. And the new loss will take the form as below:
\begin{align}
    L(D) &= L(D_1) + L(D_2) \nonumber\\
    & = - \sum_{i \in D_1} \log p_{i1} - \sum_{i \in D_2} 
    \big{(} 1 + \frac{w}{1+w p_{i3}} \big{)} \log p_{i2}\nonumber\\
    & = - \sum_{i \in D_1} \log p_{i1} - \sum_{i \in D_2} 
    \big{(} 1 + \frac{w}{1+w p_{i3}} \big{)} \log (1-p_{i1})\nonumber\\
    & \approx -|D_1| \log p_1^* - |D_2| (1+\frac{w}{1+w p_{3}^*})\log (1-p_1^*)
\end{align}
Then, the gradient of the loss will be as below
\begin{align}
    \nabla L(D) & \propto \frac{(|D_1| + |D_2| + \frac{w}{1+wp_3^*}|D_2|) * p_1^* - |D_1|}{p_1^*(1-p_1^*)} 
\end{align}
We can see that in order to minimize the loss, we can update not only $p_{1}^*$ but also $p_{3}^*$, thus providing more degrees of freedom for the model to learn reliable predictions. Instead of increasing $p_{1}^*$ and being stuck in an unreliable state, the model can increase $p_{3}^*$ when it encounters hard-ID samples. Suppose we have nonzero $p_{3}^*$ after the parameter update. We can calculate the new unreliability level following the Definition 4.2:
\begin{align}
    &\mathbf{E}_{D(\epsilon)} [ \max_{c\in \{1,2\}} \mathcal{N}(h(x) ; \mu_{c}, \Sigma_{c})] - \mathbf{E}_{D(\epsilon)} [1\{ \hat{y}=y \}] \nonumber  \\
    &=w_1 \bigg{(}\mathbf{E}_{D_1(\epsilon)} [ \max_{c\in \{1,2\}} \mathcal{N}(h(x) ; \mu_{c}, \Sigma_{c})] - \mathbf{E}_{D_1(\epsilon)} [1\{ \hat{y}=y \}] \bigg{)} \nonumber  \\
    &\ \ \ \ + w_2 \bigg{(}\mathbf{E}_{D_2(\epsilon)} [ \max_{c\in \{1,2\}} \mathcal{N}(h(x) ; \mu_{c}, \Sigma_{c})] - \mathbf{E}_{D_2(\epsilon)} [1\{ \hat{y}=y \}] \bigg{)} \nonumber \\
    & = w_1 \bigg{(}\mathbf{E}_{D_1(\epsilon)} [ \max_{c\in \{1,2\}} \mathcal{N}(h(x) ; \mu_{c}, \Sigma_{c})] - 1 \bigg{)} +
    w_2 \bigg{(}\mathbf{E}_{D_2(\epsilon)} [ \max_{c\in \{1,2\}} \mathcal{N}(h(x) ; \mu_{c}, \Sigma_{c})] - 0 \bigg{)} \nonumber \\
    & \leq w_1( p_{11} + c\epsilon -p_{3}^*-1) + w_2(p_{11} + c\epsilon -p_{3}^*)\nonumber \\
    & = p_{11} + c\epsilon - p_{3}^* - w_1 
\end{align}
Since the extra dimension (i.e., $p_{3}^*$) successfully stores the unknown information, we can assume that $p_{3}^*$ is reasonably large, i.e., $p_{3}^* > p_{11} + c\epsilon - w_1 - \eta$. Then, the unreliability level can be reduced to a value less than $\eta$:
\begin{align}
    \mathbf{E}_{D(\epsilon)} [ \max_{c\in \{1,2\}} \mathcal{N}(h(x) ; \mu_{c}, \Sigma_{c})] - \mathbf{E}_{D(\epsilon)} [1\{ \hat{y}=y \}] \leq p_{11} + c\epsilon - p_{3}^* - w_1 < \eta
\end{align}
The proof is complete.

\subsection{Insight 4.8: OOD Reliability}

(OOD Reliability) Suppose we achieve Hard-ID Reliability in Insight 4.8 with our \sysname method, we can have lower confidence on OOD data, implying stronger OOD detection.

\textit{Proof}:

We know that hard-to-detect OOD data $D_3$ usually have features close to those of hard ID data $D_2$, which is similar to the scenario described in Insight 4.4. Otherwise, if the features of the hard-to-detect OOD data are very different from those of hard ID data, then the OOD data becomes easy to detect without the OOD unreliability.

Thus, these hard-to-detect ID samples can be viewed as pseudo-OOD data. And the $\{ h(x_2); x_2 \in D_2 \}$ are close to $\{ h(x_3); x_3 \in D_3 \}$.
Then we can follow the proof of Insignt 4.6 and show that:
\begin{align}
    \mathbf{E}_{D_3(\epsilon)} [ \max_{c\in \{1,2\}} \mathcal{N}(h(x) ; \mu_{c}, \Sigma_{c})] \text{ is reduced.}
\end{align}
In this way, we are able to have smaller confidence in the hard-to-detect OOD data $D_3$, which helps to enhance the OOD reliability. This describes how we can extract unknown information from hard ID data, explaining the outlier-exposure-free property of \texttt{MOON}.

%% file: sections/_d_limit.tex
\section{Appendix: Limitations of Our \sysname}

\subsection{More General Reliability}

Despite the enhanced OOD reliability achieved in sparse training, \sysname does not address this matter comprehensively from a broader reliability perspective, including robust generalization~\citep{silva2020opportunities, ozdenizci2021training, chen2022sparsity} and adaptation~\citep{emam2021active, alayrac2022flamingo, hu2022pushing}, which are equally vital aspects of reliability and deserve further investigation~\citep{tran2022plex}.

\subsection{Reliability in Large Generative Models}

Additionally, the effectiveness of \sysname remains unexplored in the context of widely employed large generative models~\citep{ouyang2022training, croitoru2023diffusion}. On the one hand, as model size expands, novel changes in model characteristics may arise~\citep{wei2022emergent}. On the other hand, given the prevalence of non-classification tasks within LLMs, the exploration of how to evaluate their reliability remains an uncharted domain.

\section{Broader Impact}

This work is likely to encourage other future work on more general reliability in sparse training to achieve efficiency and comprehensive reliability in real-world DNN deployments.

In addition, as large generative models develop, both computational efficiency and decision reliability are important aspects of their application. This work can increase the progress of efficient and reliable large generative models.

%% file: sections/_e_flops.tex
\section{FLOPs Analysis}

In this section, we add FLOPs analysis to \texttt{MOON}. We use 80\% sparse ResNet-50 on Image-Net-2012 as an example. The training FLOPs using RigL are about 7.4e17~\citep{evci2020rigging}. For our \texttt{MOON}, the increased FLOPs come from three components, i.e., loss modification, auto tuning, and voting scheme. 

(i) For loss modification and auto-tuning.
\begin{itemize}
    \item In the early stage where $w=0$, we have about 1.3e5 FLOPs in each iteration because of the computation of $\beta$.
    \item After the early stage, from the calculation of $w$ and the new loss, we have about 2.6e5 FLOPs in each iteration.
    \item Thus, the loss modification and auto tuning introduce an additional 2.3e15 FLOPs.
\end{itemize}

(ii) For voting scheme,

We start doing weight averaging in 80\% of the training epoch and have about 1.8e9 FLOPs from the weight averaging. In each epoch after the 80\% training epoch, we also have about 2.4e15 FLOPs from the batchNorm buffer updates. Thus, the voting scheme introduces an additional 4.8e16 FLOPs.

Overall, our \sysname introduces an additional 5.0e16 FLOPs, which is about 6.8\% of the original RigL's FLOPs. The main cause of the increase in FLOPs is the update of the batchNorm buffer. To further reduce the increase of FLOPs, we can reduce the update frequency of the batchNorm buffer.

We also conduct experiments to compare the training time using our MOON and baseline methods. We train 80\%, 90\%, 95\%, and 99\% sparse ResNet-18 on both CIFAR-10 and CIFAR-100 using MOON + RigL and RigL. We denote the training time of our benchmark method, RigL, as 1 to help understand exactly what overhead our MOON adds to the training process. The results are summarized in the Table~\ref{tb: cost-c10}-\ref{tb: cost-c100}. We can see that MOON adds less than 5\% overhead.

\linespread{1.2}
\setlength\tabcolsep{5.5pt}
\begin{table*}[ht]
\caption{Training time comparison using our \sysname + RigL and RigL with 80\%, 90\%, 95\%, and 99\% sparse ResNet-18 on both CIFAR-10.}
\label{tb: cost-c10}
\begin{center}
\begin{scriptsize}
\begin{sc}
\begin{tabular}{ccccc}
\toprule \toprule
& CIFAR-10 (80\%) & CIFAR-10 (90\%) & CIFAR-10 (95\%) & CIFAR-10 (99\%) \\
\hline 
RigL & 1 & 1 & 1 & 1    \\
\sysname + RigL & 1.045 & 1.042 & 1.039 & 1.046\\
\hline
\bottomrule
\end{tabular}
\end{sc}
\end{scriptsize}
\end{center}
\end{table*}

\linespread{1.2}
\setlength\tabcolsep{5.5pt}
\begin{table*}[ht]
\caption{Training time comparison using our \sysname + RigL and RigL with 80\%, 90\%, 95\%, and 99\% sparse ResNet-18 on both CIFAR-100.}
\label{tb: cost-c100}
\begin{center}
\begin{scriptsize}
\begin{sc}
\begin{tabular}{ccccc}
\toprule \toprule
& CIFAR-10 (80\%) & CIFAR-10 (90\%) & CIFAR-10 (95\%) & CIFAR-10 (99\%) \\
\hline 
RigL & 1 & 1 & 1 & 1    \\
\sysname + RigL & 1.049 & 1.046 & 1.048 & 1.043\\
\hline
\bottomrule
\end{tabular}
\end{sc}
\end{scriptsize}
\end{center}
\end{table*}

\section{Hyperparameter Tuning}

The hyperparameters of the baseline method follow their original settings in~\citep{evci2020rigging, yang2022openood}. The hyperparameters of our \sysname are described as follows.

(i) For the learning rate, following~\citep{yang2022openood}, we use the cosine annealing scheduler, where the maximum learning rate is set to 1.2 and the minimum learning rate is set to 0.08.

(ii) For the fraction of weights removed and added (pruning ratio) in each sparse pattern update, we follow the original settings in~\citep{evci2020rigging, sundar2021reproducibility}.
\begin{itemize}
    \item The initial fraction is 0.3.
    \item This fraction decays in the mask update step by cosine annealing.
    \item The score is set to zero after 70\% of the training cycles.
\end{itemize}

(iii) For the weight distribution, we follow the original settings in~\citep{evci2020rigging, sundar2021reproducibility} and initialize the mask by Erdos-Renyi-Kernel (ERK).

(iv) For $w_f$ and $r$
\begin{itemize}
    \item For MNIST, we set $w_f$ and $r$ as 0.01 and 64, respectively.
    \item For CIFAR-10, we set $w_f$ and $r$ as 1 and 64, respectively.
    \item For CIFAR-100, we set $w_f$ and $r$ as 2 and 64, respectively.
    \item For ImageNet-2012, we set $w_f$ and $r$ as 0.001 and 64, respectively.
\end{itemize}

\section{Feature Visualization}

One of the assumptions on which our theory is based is that the feature distribution of a DNN feature extractor is a Gaussian mixture. Additional experiments are added to support this assumption. Specifically, we train ResNet-18 on CIFAR-10 with 80\%, 90\%, 95\% and 99\% sparsity and collect features before the last layer. Then, to visualize the feature distribution, we use the PCA method to reduce the dimensionality and plot the histogram of the first component.

\begin{figure*}[!t]
\begin{center}
\subfigure[Class 0]{\includegraphics[scale=0.33]{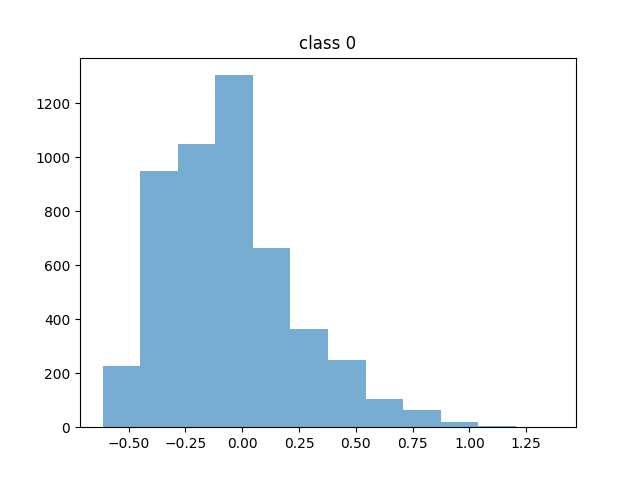}}
\subfigure[Class 1]{\includegraphics[scale=0.33]{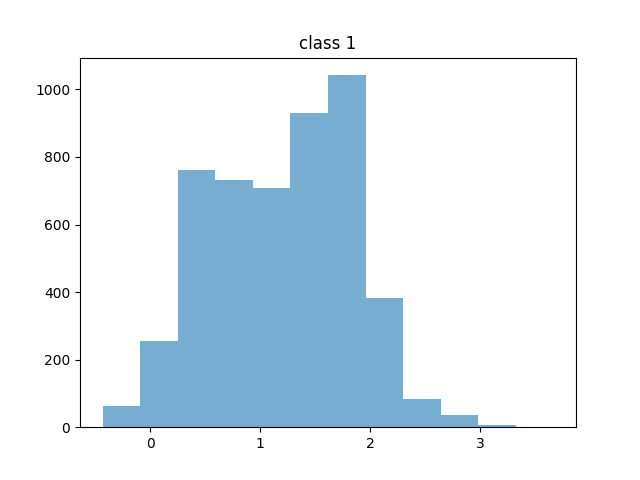}}
\subfigure[Class 2]{\includegraphics[scale=0.33]{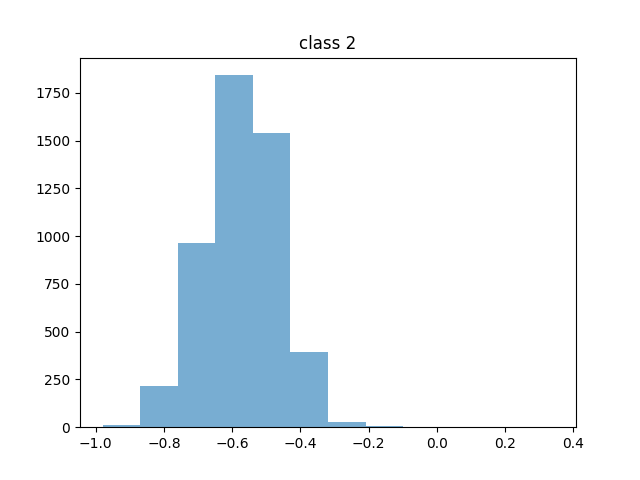}}
\subfigure[Class 3]{\includegraphics[scale=0.33]{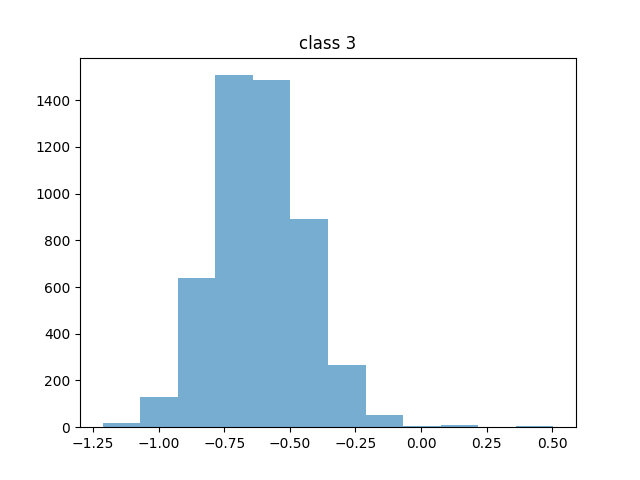}}
\subfigure[Class 4]{\includegraphics[scale=0.33]{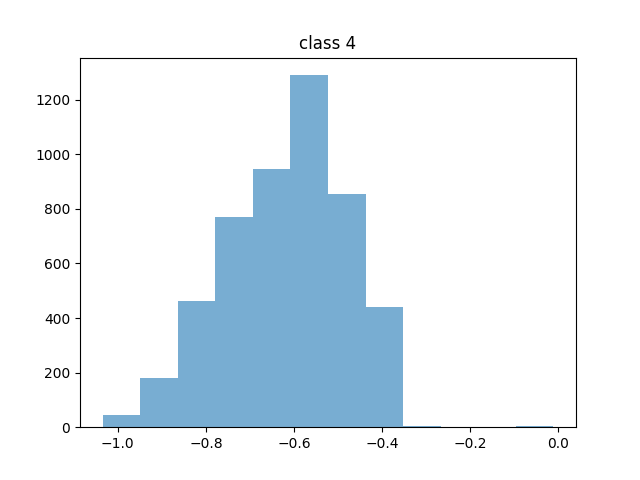}}
\subfigure[Class 5]{\includegraphics[scale=0.33]{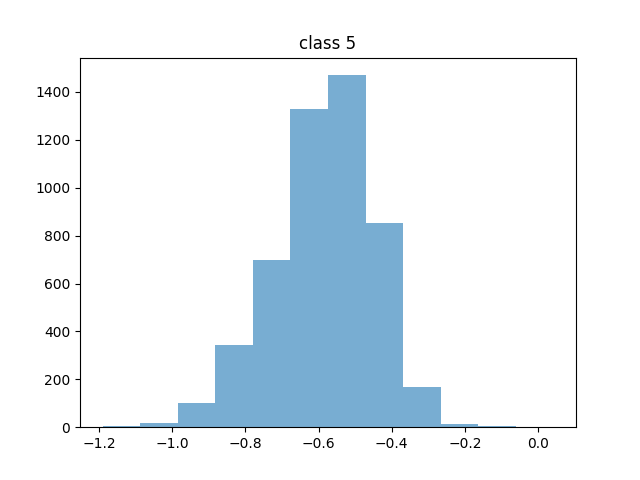}}
\subfigure[Class 6]{\includegraphics[scale=0.33]{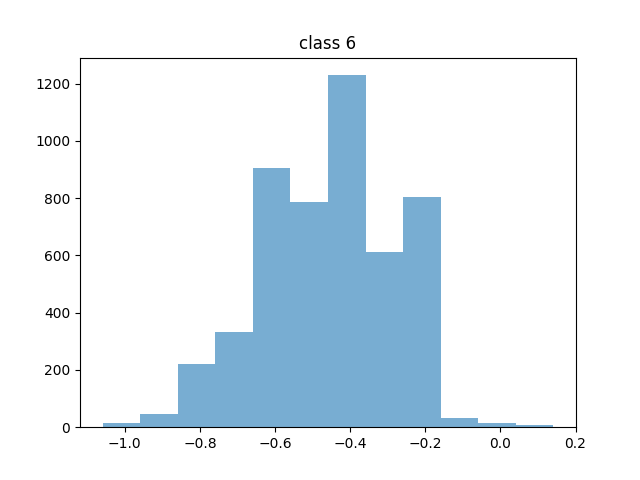}}
\subfigure[Class 7]{\includegraphics[scale=0.33]{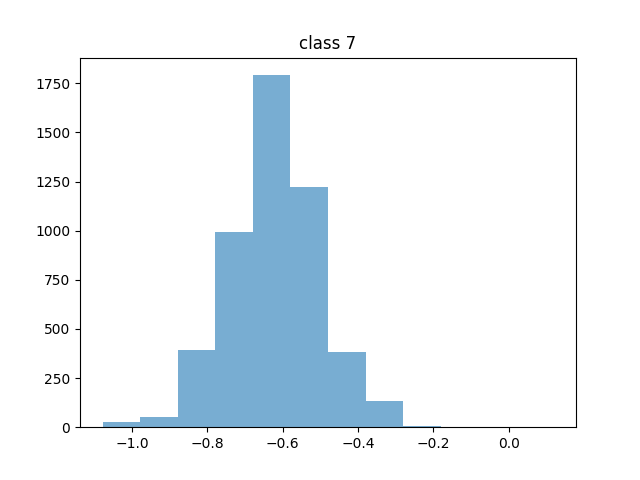}}
\subfigure[Class 8]{\includegraphics[scale=0.33]{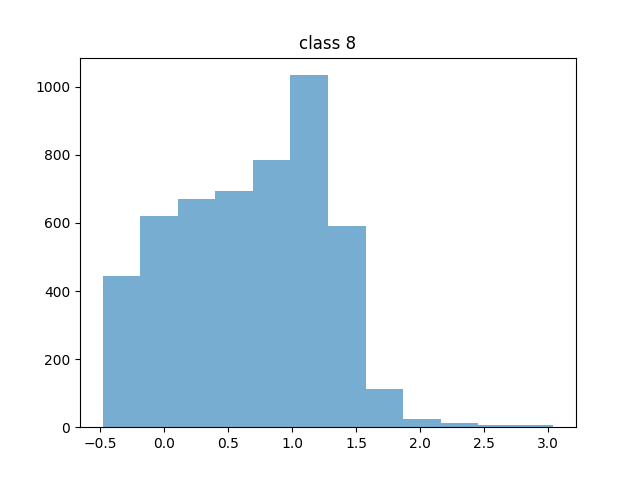}}
\subfigure[Class 9]{\includegraphics[scale=0.33]{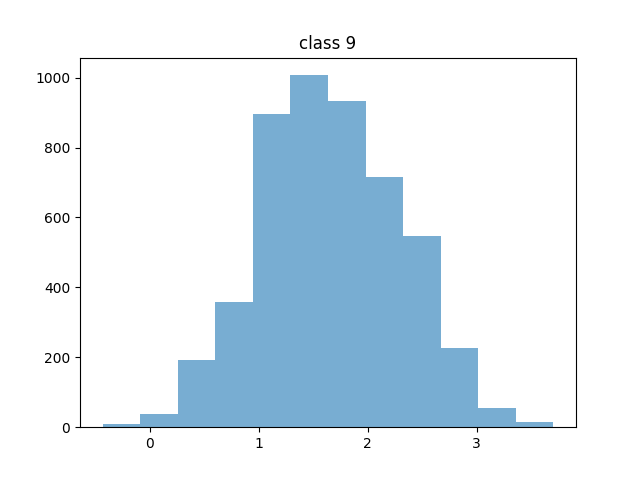}}
\caption{Feature distribution visualization on CIFAR-10 for 80\% sparse ResNet-18. The feature distribution shows a high density in the center and then decreases towards the two sides across all classes.}
\label{fig: fea-distri}
\end{center}
\end{figure*}

\begin{figure*}[!t]
\begin{center}
\subfigure[Class 0]{\includegraphics[scale=0.33]{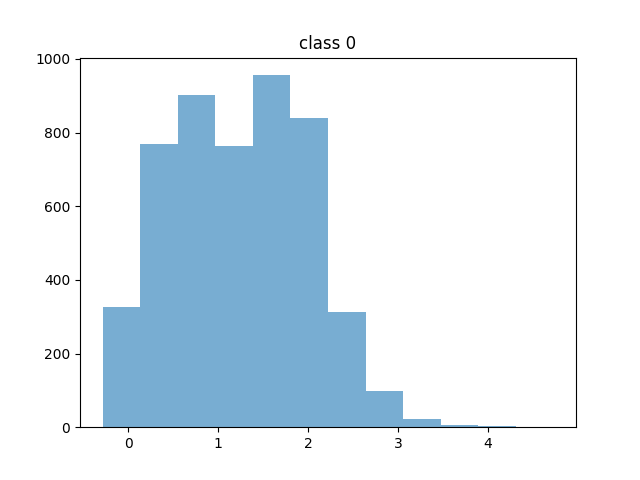}}
\subfigure[Class 1]{\includegraphics[scale=0.33]{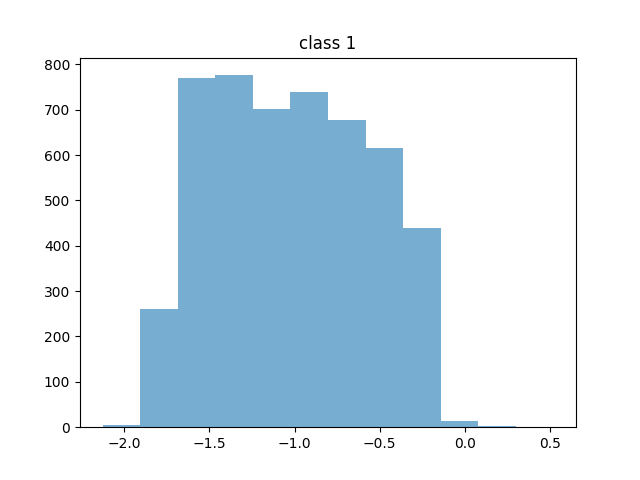}}
\subfigure[Class 2]{\includegraphics[scale=0.33]{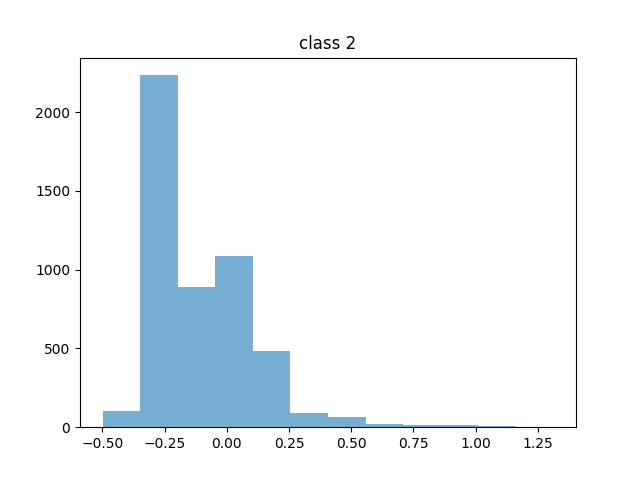}}
\subfigure[Class 3]{\includegraphics[scale=0.33]{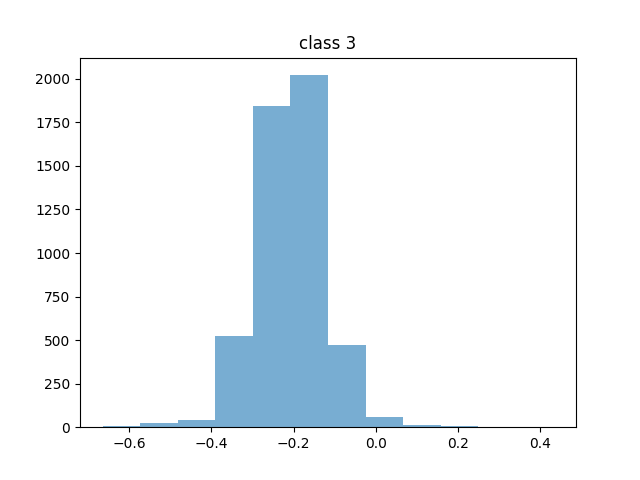}}
\subfigure[Class 4]{\includegraphics[scale=0.33]{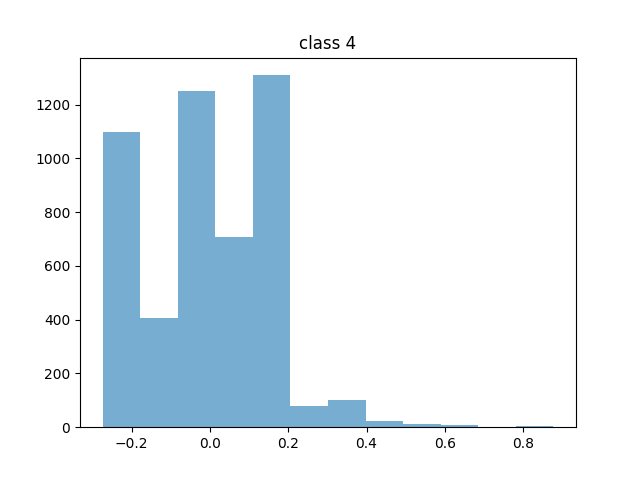}}
\subfigure[Class 5]{\includegraphics[scale=0.33]{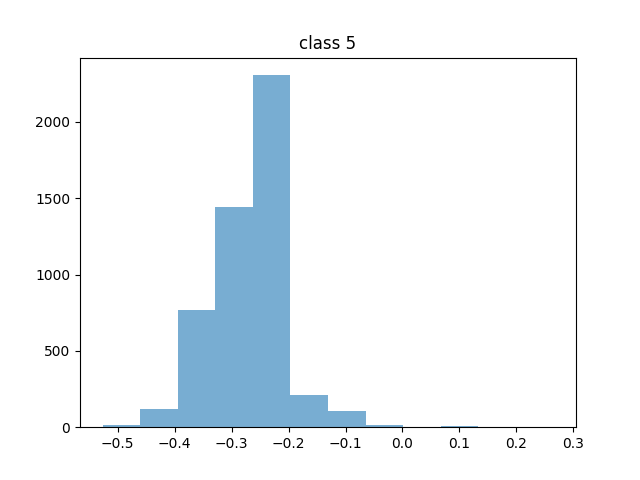}}
\subfigure[Class 6]{\includegraphics[scale=0.33]{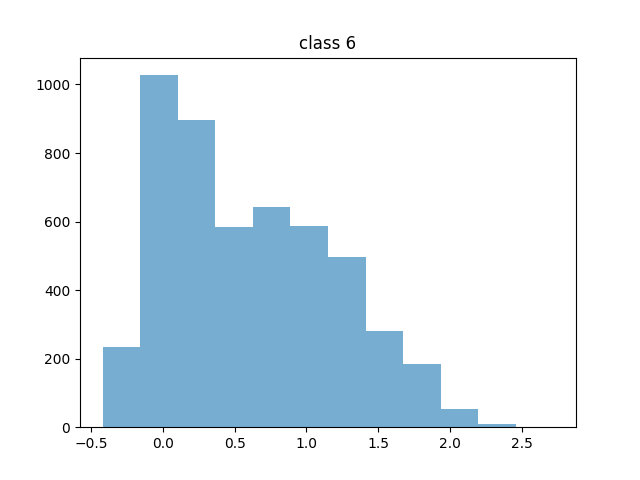}}
\subfigure[Class 7]{\includegraphics[scale=0.33]{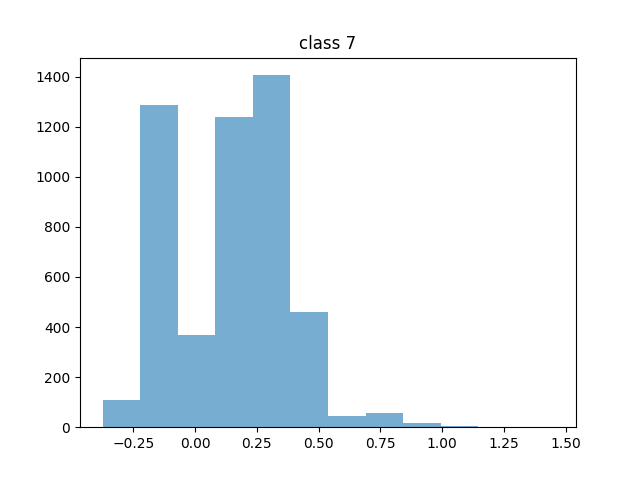}}
\subfigure[Class 8]{\includegraphics[scale=0.33]{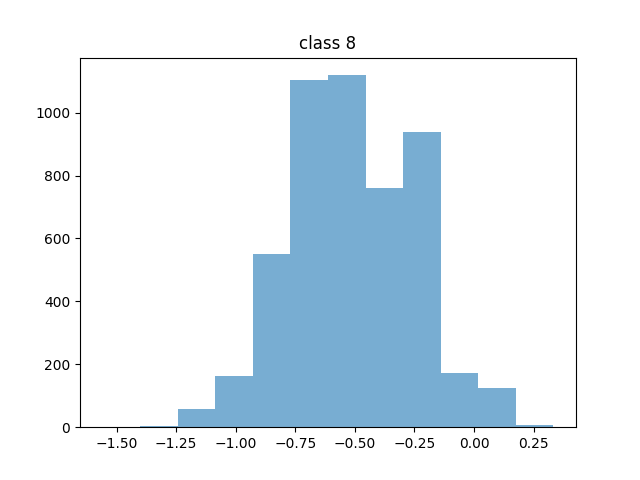}}
\subfigure[Class 9]{\includegraphics[scale=0.33]{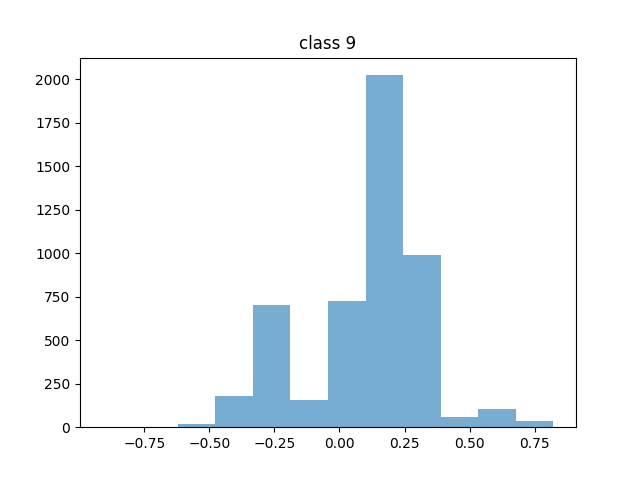}}
\caption{Feature distribution visualization on CIFAR-10 for 90\% sparse ResNet-18. The feature distribution shows a high density in the center and then decreases towards the two sides across all classes.}
\label{fig: fea-distri1}
\end{center}
\end{figure*}

\begin{figure*}[!t]
\begin{center}
\subfigure[Class 0]{\includegraphics[scale=0.33]{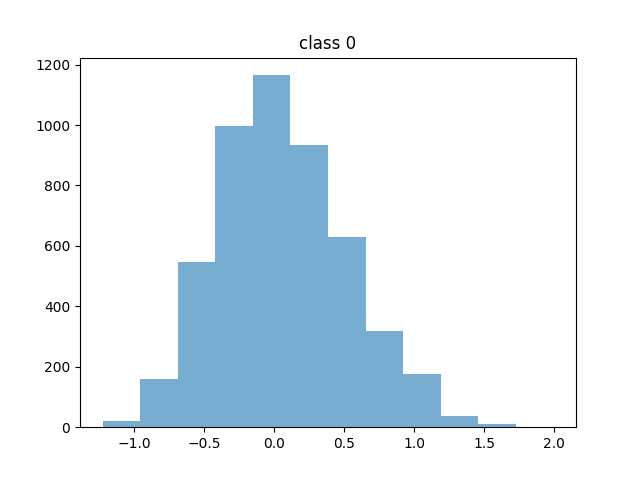}}
\subfigure[Class 1]{\includegraphics[scale=0.33]{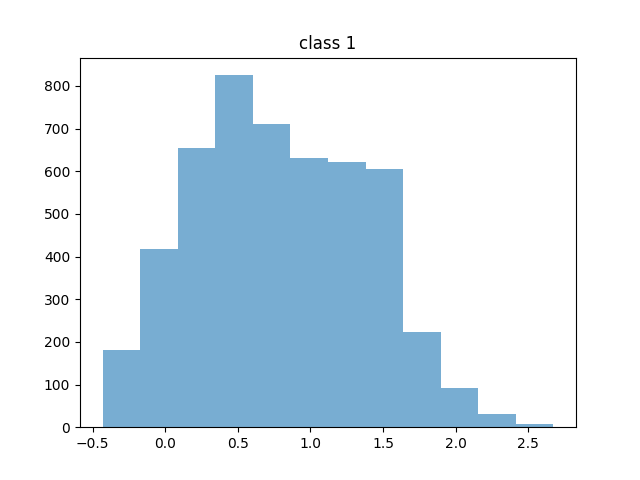}}
\subfigure[Class 2]{\includegraphics[scale=0.33]{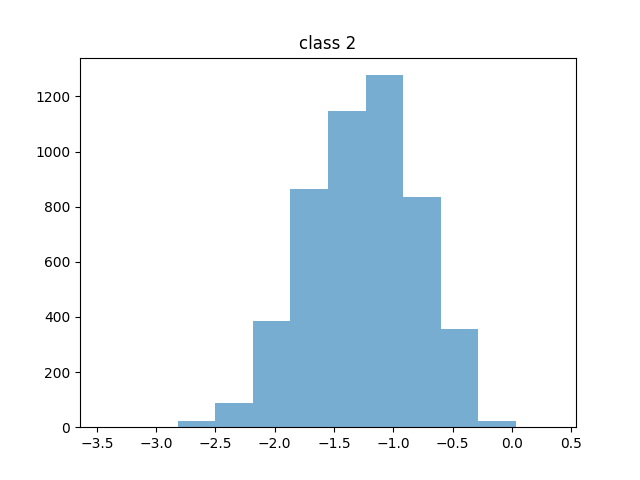}}
\subfigure[Class 3]{\includegraphics[scale=0.33]{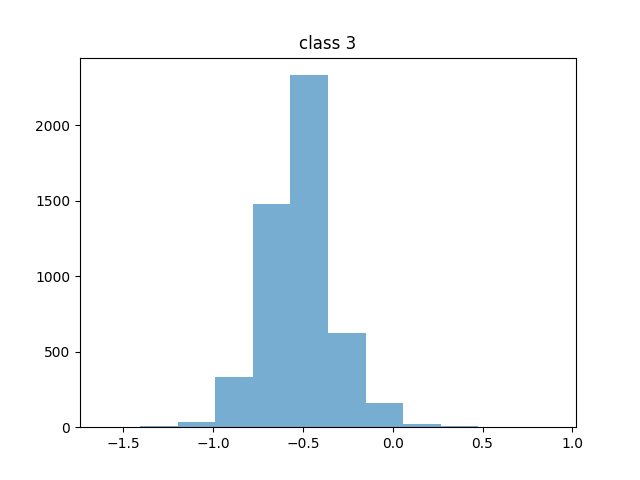}}
\subfigure[Class 4]{\includegraphics[scale=0.33]{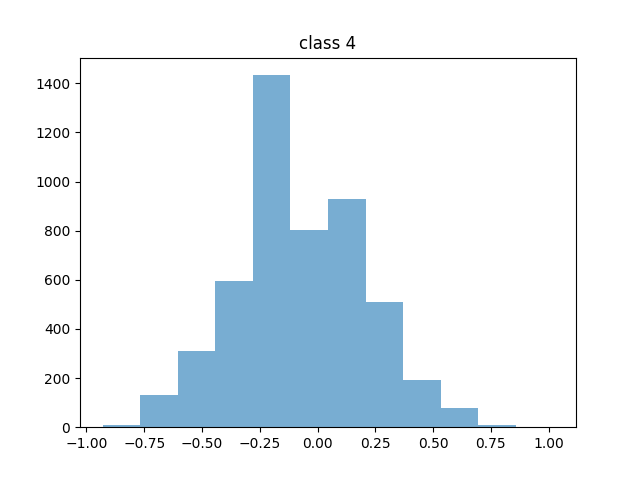}}
\subfigure[Class 5]{\includegraphics[scale=0.33]{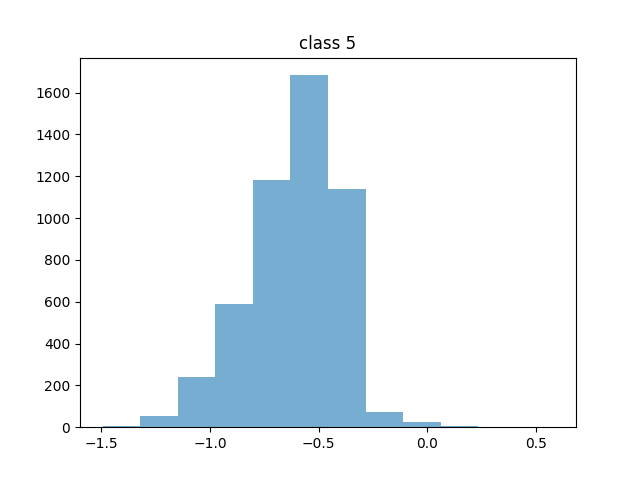}}
\subfigure[Class 6]{\includegraphics[scale=0.33]{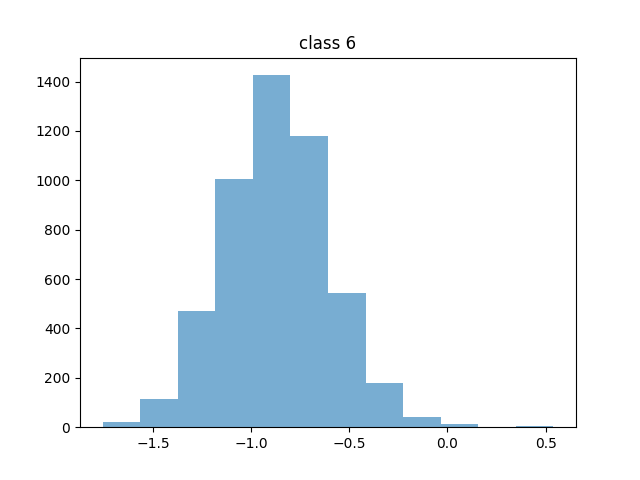}}
\subfigure[Class 7]{\includegraphics[scale=0.33]{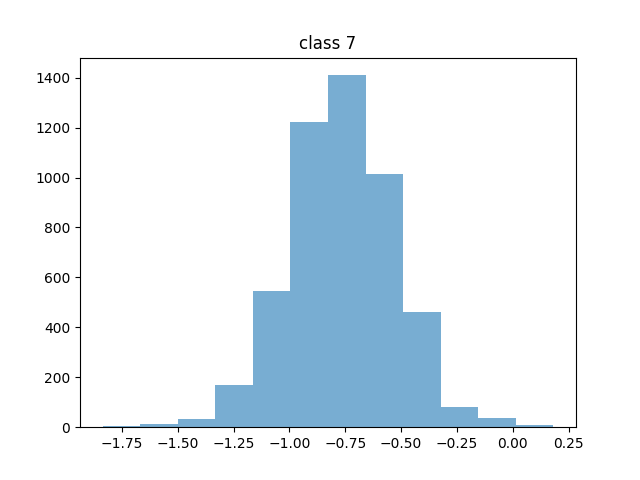}}
\subfigure[Class 8]{\includegraphics[scale=0.33]{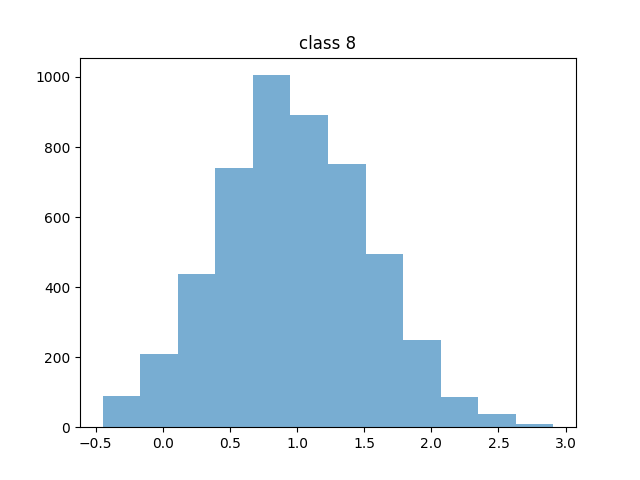}}
\subfigure[Class 9]{\includegraphics[scale=0.33]{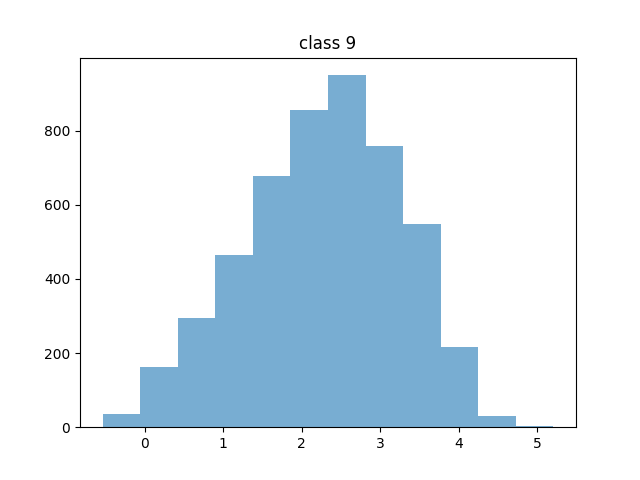}}
\caption{Feature distribution visualization on CIFAR-10 for 95\% sparse ResNet-18. The feature distribution shows a high density in the center and then decreases towards the two sides across all classes.}
\label{fig: fea-distri2}
\end{center}
\end{figure*}

\begin{figure*}[!t]
\begin{center}
\subfigure[Class 0]{\includegraphics[scale=0.33]{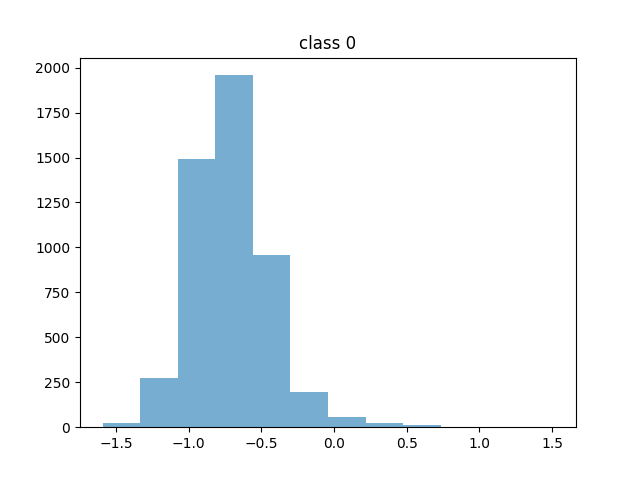}}
\subfigure[Class 1]{\includegraphics[scale=0.33]{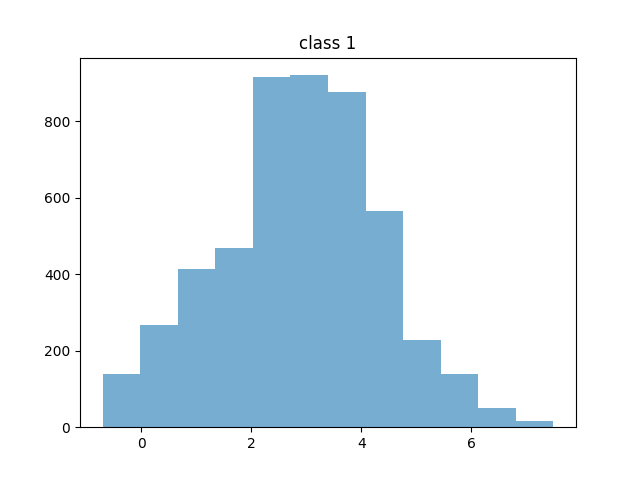}}
\subfigure[Class 2]{\includegraphics[scale=0.33]{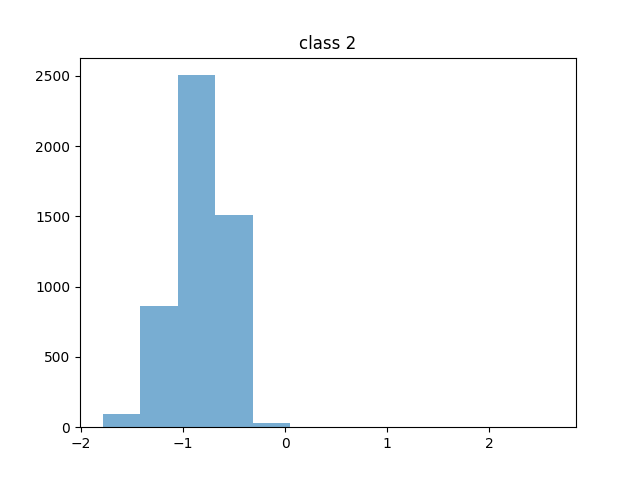}}
\subfigure[Class 3]{\includegraphics[scale=0.33]{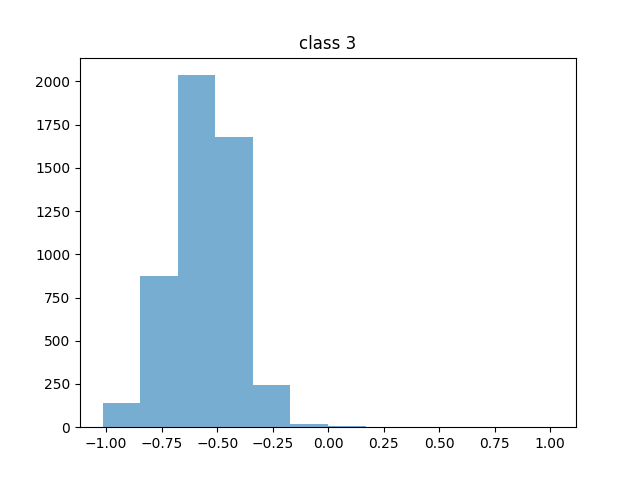}}
\subfigure[Class 4]{\includegraphics[scale=0.33]{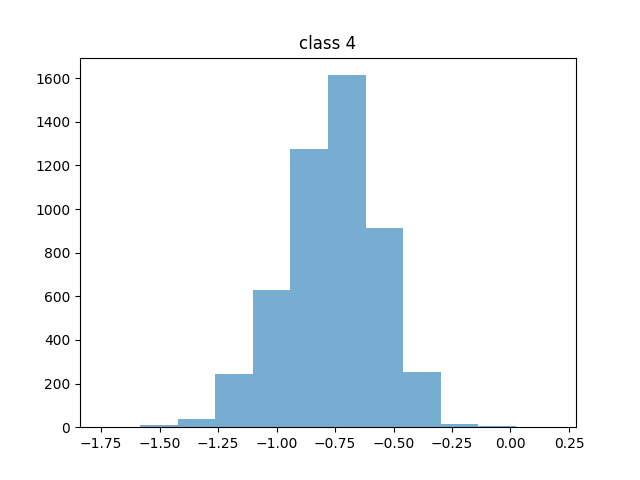}}
\subfigure[Class 5]{\includegraphics[scale=0.33]{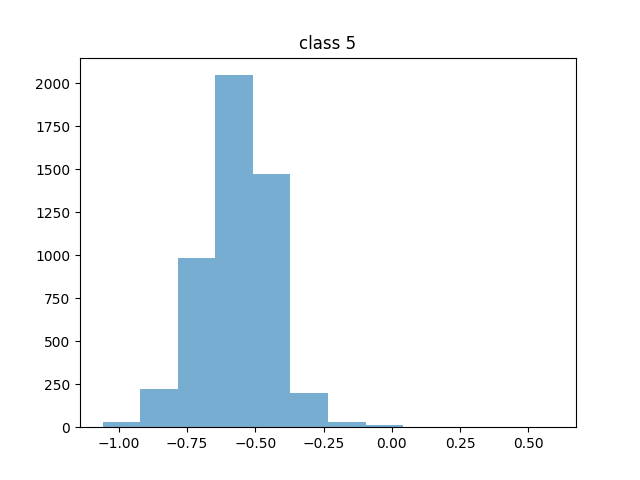}}
\subfigure[Class 6]{\includegraphics[scale=0.33]{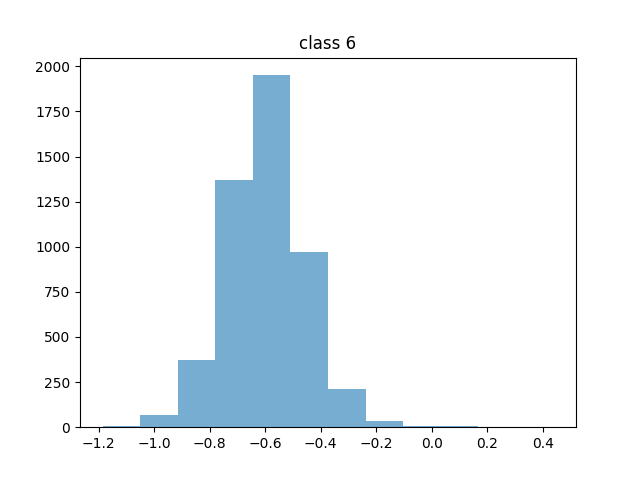}}
\subfigure[Class 7]{\includegraphics[scale=0.33]{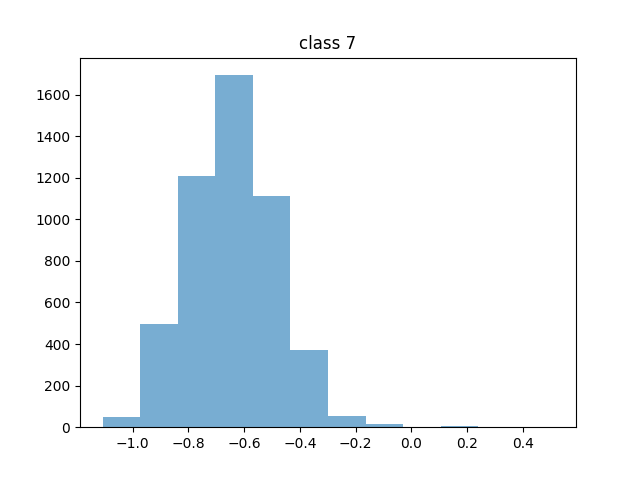}}
\subfigure[Class 8]{\includegraphics[scale=0.33]{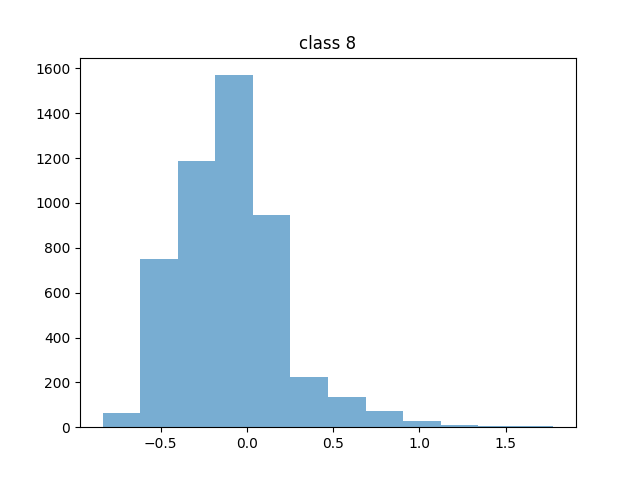}}
\subfigure[Class 9]{\includegraphics[scale=0.33]{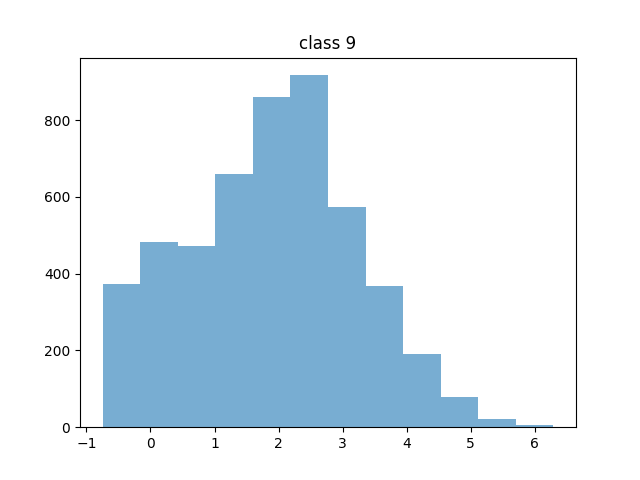}}
\caption{Feature distribution visualization on CIFAR-10 for 99\% sparse ResNet-18. The feature distribution shows a high density in the center and then decreases towards the two sides across all classes.}
\label{fig: fea-distri3}
\end{center}
\end{figure*}

As shown in Figure~\ref{fig: fea-distri}-\ref{fig: fea-distri3}, the feature distribution shows a high density in the center and then decreases towards the two sides across all classes and sparsity levels. This characterization is consistent with a Gaussian distribution. Therefore, the overall distribution for all classes can be approximated by a Gaussian mixture distribution, supporting our assumption.

